%% file: main.tex
\documentclass[10pt]{article} % For LaTeX2e
\usepackage[accepted]{tmlr}

\usepackage{graphicx} % Required for inserting images
\usepackage{bm}
\usepackage{xspace}
\usepackage{amssymb}
\usepackage{mathtools}
\usepackage{amsthm}
\usepackage{amsmath}
\usepackage{natbib}

\usepackage[utf8]{inputenc} % allow utf-8 input
\usepackage[T1]{fontenc}    % use 8-bit T1 fonts
\usepackage{hyperref}       % hyperlinks
\usepackage{url}            % simple URL typesetting
\usepackage{booktabs}       % professional-quality tables
\usepackage{amsfonts}       % blackboard math symbols
\usepackage{nicefrac}       % compact symbols for 1/2, etc.
\usepackage{microtype}      % microtypography
\usepackage{xcolor}         % colors
\usepackage{tikz}
\usetikzlibrary{arrows, positioning}

\usepackage{enumitem}
\usepackage{dsfont}

\usepackage{listings}
\usepackage{algorithm}

\let\classAND\AND
\let\AND\relax
\usepackage{algorithmic}
\let\AND\classAND
\AtBeginEnvironment{algorithmic}{\let\AND\algoAND}

\usepackage{bold-extra}

\theoremstyle{plain}

\hypersetup{
    colorlinks,
    linkcolor={red!50!black},
    citecolor={blue!50!black},
    urlcolor={blue!80!black}
}

\title{
A unifying framework for generalised Bayesian online learning in non-stationary environments
}
\author{\name Gerardo Duran-Martin
    \email g.duran@me.com\\
    \addr School of Mathematical Sciences\\
    Queen Mary University of London, UK
    \AND
        Leandro Sánchez-Betancourt
    \email sanchezbetan@maths.ox.ac.uk\\
    \addr Mathematical Institute \& \\ Oxford-Man Institute of Quantitative Finance \\ University of Oxford, UK
    \AND
    \name Alexander Y. Shestopaloff
    \email a.shestopaloff@qmul.ac.uk\\
    \addr School of Mathematical Sciences\\
    Queen Mary University of London, UK
    \& \\ Department of Mathematics and Statistics \\ Memorial University of Newfoundland, Canada
    \AND
    \name Kevin Murphy 
    \email kpmurphy@google.com\\
    \addr Google DeepMind\\
    Moutain View, CA, USA 
}

\def\month{03}  % Insert correct month for camera-ready version
\def\year{2025} % Insert correct year for camera-ready version
\def\openreview{\url{https://openreview.net/forum?id=osesw2V10u}} % Insert correct link to OpenReview for camera-ready version

\input{macros.tex}

\begin{document}

\maketitle

\input{abstract}

\input{sections/introduction}
\input{sections/method}

\input{sections/experiments}

\input{sections/conclusions}

\bibliography{refs}
\bibliographystyle{tmlr}

\appendix
\input{sections/appendix/weights-caux}

\input{sections/appendix/algorithms}
\input{sections/appendix/full-prediction-in-bayes}

\end{document}

%% file: macros.tex
\newcommand{\eat}[1]{}

\newcommand{\abonelib}{\{redacted\}}

\newcommand{\namemethod}[1]{\texorpdfstring{\texttt{\color{black}{#1}}\xspace}{}}
\newcommand{\newmethod}[1]{\texorpdfstring{\texttt{\color{black}{#1}}\xspace}{}}

\newcommand{\cModelraw}{(M.1)\xspace}
\newcommand{\cAuxraw}{(M.2)\xspace}
\newcommand{\cPriorraw}{(M.3)\xspace}
\newcommand{\cPosteriorraw}{(A.1)\xspace} 
\newcommand{\cWeightraw}{(A.2)\xspace}

\newcommand{\cModel}{(M.1: model)\xspace}
\newcommand{\cAux}{(M.2: auxvar)\xspace}
\newcommand{\cPrior}{(M.3: prior)\xspace}
\newcommand{\cPosterior}{(A.1: posterior)\xspace} 
\newcommand{\cWeight}{(A.2: weighting)\xspace}

\newcommand{\cAuxname}{M.2: auxvar\xspace}
\newcommand{\cPriorname}{M.3: prior\xspace}
\newcommand{\cPosteriorname}{A.1: posterior\xspace} 
\newcommand{\cWeightname}{A.2: weight\xspace}

\newcommand{\RLSPR}[1][]{\newmethod{RL[1]-OUPR*#1}}
\newcommand{\RLPRWoLF}[1][inf]{\newmethod{WoLF+RL[#1]-PR*}}
\newcommand{\RLPRKF}[1][inf]{\namemethod{LG+RL[#1]-PR}}
\newcommand{\staticKF}{\namemethod{LG+C-Static}}

\newcommand{\CACI}[1][]{\namemethod{C-ACI#1}} % constant covariance inflation
\newcommand{\RLPR}[1][1]{\namemethod{RL[#1]-PR}}
\newcommand{\RLPRK}{\namemethod{RL[K]-PR}}
\newcommand{\CPPD}[1][]{\namemethod{CPP-OU#1}}

\renewcommand{\d}{\mathrm{d}}
\DeclareMathOperator*{\argmax}{arg\,max}
\DeclareMathOperator*{\argmin}{arg\,min}
\newcommand{\cond}{\,\vert\,}

\newcommand{\dimstate}{m}
\newcommand{\dimobs}{d}
\newcommand{\dimin}{q}

\newcommand{\auxv}{\psi}

\newcommand{\real}{\mathbb{R}}

\newcommand{\myvec}[1]{\mathbf{#1}}
\newcommand{\myvecsym}[1]{\boldsymbol{#1}} 

\newcommand{\vzero}{\myvecsym{0}}

\newcommand{\valpha}{\myvecsym{\alpha}}

\newcommand{\vepsilon}{\myvecsym{\epsilon}}
\newcommand{\vzeta}{\myvecsym{\zeta}}

\newcommand{\vmu}{\myvecsym{\mu}}

\newcommand{\vpsi}{\myvecsym{\psi}}
\newcommand{\vPsi}{\myvecsym{\Psi}}

\newcommand{\vtheta}{\myvecsym{\theta}}

\newcommand{\vSigma}{\myvecsym{\Sigma}}

\newcommand{\vb}{\myvec{b}}

\newcommand{\vh}{\myvec{h}}

\newcommand{\vw}{\myvec{w}}

\newcommand{\vx}{\myvec{x}}

\newcommand{\vy}{\myvec{y}}

\newcommand{\vz}{\myvec{z}}
\def\vtheta{{\bm{\theta}}}

\def\vb{{\bm{b}}}

\def\vh{{\bm{h}}}

\def\vw{{\bm{w}}}
\def\vx{{\bm{x}}}
\def\vy{{\bm{y}}}
\def\vz{{\bm{z}}}

\newcommand{\vF}{\myvec{F}}

\newcommand{\vH}{\myvec{H}}
\newcommand{\vI}{\myvec{I}}

\newcommand{\vK}{\myvec{K}}

\newcommand{\vQ}{\myvec{Q}}
\newcommand{\vR}{\myvec{R}}
\newcommand{\vS}{\myvec{S}}

\newcommand{\data}{\mathcal{D}}

%% file: abstract.tex
\begin{abstract}
We propose a unifying framework
for methods
that perform probabilistic online learning in non-stationary environments. 
We call the framework BONE, which stands for generalised (B)ayesian (O)nline learning in (N)on-stationary (E)nvironments. 
BONE provides a common structure to tackle a variety of problems,
including online continual learning, prequential forecasting, 
and contextual bandits.
The framework requires specifying
 three modelling choices:
(i) a model for measurements (e.g., a neural network),
(ii) an auxiliary process to model non-stationarity (e.g., the time since the last changepoint), and
(iii) a conditional prior over model parameters  (e.g., a multivariate Gaussian).
The framework also requires two algorithmic choices, which we use to carry out approximate inference under this framework:
(i)
an algorithm to estimate beliefs (posterior distribution) about the model parameters given the auxiliary variable,
and
(ii) an algorithm to estimate beliefs about the auxiliary variable.
We show how the modularity of our framework allows for many existing
methods to
be reinterpreted as instances of BONE, and it allows us 
to propose new methods.
We compare experimentally existing methods with our proposed new method on several datasets,
providing insights into the situations that make each method more suitable for a specific task.
We provide a Jax open source library to facilitate the adoption of this framework.
\end{abstract}

%% file: sections/introduction.tex
\section{Introduction}

In this paper, we unify adaptive probabilistic methods
that learn to make predictions
about the next output  $\vy_{t+1}$ 
based on the next input $\vx_{t+1}$
and a sequence of past inputs
and outputs, $(\vx_{1:t}, \vy_{1:t})$,
where $t$ indexes 
time.
This is often called prequential forecasting \citep{gama2008streamlearn},
online learning \citep[Chapter 1]{zhang2023ltbook},
or sequential online prediction \citep{liu2023bdemm}.

Many probabilistic prediction methods assume that  the data 
generating process (DGP)
$p(\vy_{t+1} \cond \vx_{t+1},\vx_{1:t},\vy_{1:t})$
is static through time. 
%,or, more generally, that the model to make predictions is well-specififed.
However, real-world data often comes from non-stationary distributions, where the
underlying data distribution changes,
either gradually (e.g., rising mean temperature)
or abruptly (e.g., price shocks after a major
news event).
In this paper we propose a unified framework for tackling  (one-step-ahead) forecasts in such (potentially) non-stationary  environments.
Building on recent advances in generalised (pseudo) Bayesian methods \citep{bissiri2016generalbayes,knoblauch2022gvi,khan2023bayesian},
we show that our framework unifies classical Bayesian approaches and naturally extends to methods derived from Bayes’ rule.
We demonstrate that all these methods are related to a three-layered hierarchical state-space model,
where the top layer captures changes in the data, the second layer governs the evolution of model parameters,
and the third layer represents the observations.

Our framework
accommodates and extends many existing lines of work,
including online continual learning \citep{dohare2024continualbackprop},
prequential (one-step-ahead) forecasting \citep{liu2023bdemm}, 
test-time adaptation \citep{schirmer2024ssmtta},
neural contextual bandits \citep{riquelme2018banditshowdown},
filtering \citep{basseville1993onlinedetection}, and
changepoint detection or segmentation \citep{gupta2024changepointsurvey}.
For example, methods that were originally developed to tackle changepoint detection
can be reinterpreted and applied within our framework to address a wider range of sequential online learning problems.
This flexibility allows us to gather tools that were
 designed for specific tasks---often studied in isolation within distinct subfields---into
a unifying framework for handling non-stationarity across various application domains.
% deep reinforcement learning \cite{asadi2024rlreset}.

The framework we propose in this paper,
which we call  BONE---which stands for (B)ayesian (O)nline learning in (N)on-stationary (E)nvironments---is
based on a form of generalised Bayesian inference in a hierarchical model,
% \footnote{
% It is worth clarifying what we mean by ``a form of generalised Bayesian inference''.
% We focus on adaptive rules, which are defined as
% ``accumulating experience about the properties of the environment and leveraging this experience to improve the model's performance'' \citep{peterka1981bayesian}.
% While the adaptive rules that we consider in this work align with Bayes' rule,
% this alignment does not imply that we adopt the full set of assumptions required to be strictly Bayesian.
% By strictly Bayesian, we mean a framework in which the properties of the data-generating process are fully known, well specified, and described within a formal mathematical model with uncertainty quantified about all unknowns \citep{faden1976postbayes,peterka1981bayesian,knoblauch2022gvi}.
% Thus, our approach uses Bayes' rule only to provide a rational basis for sequential decision-making while preserving adaptability.
% }
and is composed of three modelling choices and two algorithmic choices.
The modelling choices are:
% The three ingredients of our framework are: 
\cModelraw a model for the measurements,
\cAuxraw an auxiliary process to model non-stationarity, and
\cPriorraw a prior over model parameters conditioned on the auxiliary process and the past data.
The  algorithmic choices are:
\cPosteriorraw an algorithm to compute an approximate posterior over the
model parameters given the auxiliary variables,
and 
\cWeightraw an algorithm to compute an approximate posterior---or more generally, a set of weights---over the auxiliary variables.
We show how these different axes of variation span a wide variety of existing and new methods.\footnote{
BONE is designed for sequential prediction with time-indexed data, thus, 
it does not aim to find a ``global'' fit over all possible data points.
Instead, BONE focuses on adaptive strategies that forecast the next datapoint (one at a time).
This is in contrast to global optimisation approaches, such as \cite{luo2024-globalbayesopt},
which focus on finding the best global model for all datapoints.
There are a number of probabilistic methods that handle non-stationarity and are not included in BONE. For example, 
\cite{scalzo2021nonstationary} and \cite{vilmarest2021viking}
 do not  decouple the modelling of non-stationarity and the modelling over model parameters,
 thus, they cannot be written within our formulation.
}
% (see Table \ref{tab:related-bone-methods}).
To illustrate this, Figure \ref{fig:diagram-bone-examples} shows how various methods can be written within the BONE framework.
Here, we categorised methods by how the auxiliary variable is used to model non-stationarity and
how it influences the prior over model parameters before observing a new data point.
% We then perform an experimental comparison on a variety of tasks, including
% prequential forecasting, classification, bandits, and unsupervised time-series segmentation

Examples demonstrating the use of an easy-to-use library that implements these methods, written in Jax \citep{jax2018github},
are available at
\url{https://github.com/gerdm/BONE}.

To summarise,
our contributions are threefold:
(1) we provide an extensive literature review on methods that tackle non-stationarity, and show
that they can all be written as instances
of our unified  BONE framework;
(2) we use the BONE framework to develop a new method; and
(3) we perform an experimental comparison of many existing  methods and the new method on  environments with both abrupt changes and gradually-changing distributions.

\begin{figure}[htb]
    \centering
    \input{diagrams/bone-diagram}
    \caption{
        Overview of BONE methods grouped by the qualitative nature of the auxiliary variable and the conditional prior.
        See Table \ref{tab:related-bone-methods} for a detailed breakdown of these methods.
    }
    \label{fig:diagram-bone-examples}
\end{figure}

%% file: diagrams/bone-diagram.tex
% Original styles
\tikzstyle{start}    = [rectangle, rounded corners, minimum width=2.5cm, minimum height=1cm, text centered, draw=black, fill=gray!0]
\tikzstyle{decision} = [rectangle, rounded corners, minimum width=2cm, minimum height=1cm, text centered, draw=black, fill=gray!0]
\tikzstyle{process}  = [rectangle, rounded corners, minimum width=2.5cm, minimum height=1cm, text centered, draw=black, fill=gray!0]
\tikzstyle{method}   = [rectangle, rounded corners, minimum width=3cm, minimum height=1cm, text centered, draw=black, fill=gray!0]
\tikzstyle{future}   = [rectangle, rounded corners, minimum width=1cm, minimum height=1cm, text centered, draw=black, dashed, fill=red!0]
\tikzstyle{arrow}    = [thick,->,>=stealth]

\begin{tikzpicture}[node distance=1.5cm and 2.5cm] % Increased horizontal spacing
    \scriptsize
    % Level 2: Initial nodes (centered)
    \node (discrete) [decision, align=center] {Discrete \\auxiliary variable};
    \node (continuous) [decision, below=of discrete, align=center] {Continuous \\auxiliary variable};
    \node (noaux) [decision, below=of continuous, align=center] {No \\auxiliary variable};
    
    % Level 3: Process nodes for Discrete (aligned with spacing)
    \node (lookback) [process, right=of discrete, yshift=3.0cm, align=center] {abrupt\\with reset};
    \node (mixture) [process, right=of discrete, yshift=1.5cm, align=center] {mixture\\of experts};
    \node (pastK) [process, right=of discrete, yshift=0.0cm, align=center] {choose from\\replay buffers};
    \node (abruptdrift) [process, right=of discrete, yshift=-1.5cm, align=center] {abrupt\\and gradual};
    
    % Level 3: Process nodes for Continuous (aligned with spacing)
    \node (forgetting) [process, right=of continuous, align=center, yshift=-2.0cm] {degree\\of forgetting};
    \node (ppastK) [process, right=of continuous, yshift=-0.5cm, align=center] {weighted from\\replay buffers};

    % Level 3: Process nodes for fixed (aligned with spacing)
    \node (ratechange) [process, right=of noaux, yshift=-1.0cm] {constant change};
    \node (knowndyn) [process, right=of noaux, yshift=-2.5cm] {known dynamics};
    % \node (ratechange) [process, right=of noaux, yshift=-2.0cm] {constant change};
    
    % Level 4: Method nodes (adjusted horizontal placement)
    % Discrete branch
    \node (boCD) [method, right=of lookback, align=left, xshift=-0.5cm] 
    {\cite{adams2007bocd}\\\cite{fearnhead2007line}\\\cite{wilson2010-bocd-hazard-rate}
    \\\cite{knoblauch2018doublyrobust-bocd}};
    \node (mixtureMethods) [method, right=of mixture, xshift=-0.5cm, align=left] 
      {\cite{chang1978switchingkf}\\\cite{chaer1997mixturekf}\\\cite{liu2023bdemm}\\\cite{abeles2024adaptive}};
    \node (pastKMethods) [method, right=of pastK, xshift=-0.5cm, align=left] 
      {\cite{nguyen2017vcl}\\\cite{li2021onlinelearning}};
    \node (abruptdriftmethods) [method, right=of abruptdrift, xshift=-0.5cm, align=left] 
      {\\\cite{fearnhead2011adaptivecp}\\New method in  \eqref{eq:SPR-equation-gt}};
    
    % Continuous branch
    \node (forgettingMethods) [method, right=of forgetting, xshift=-0.5cm, align=left] 
      {\cite{kurle2019continual}\\\cite{titsias2023kalman}\\\cite{Galashov2024}};
    \node (probak) [method, right=of ppastK, xshift=-0.5cm, align=left] 
      {\cite{nassar2022bam}};

    % Fixed branch
    \node (rateChangeMethods) [method, right=of ratechange, xshift=-0.5cm, align=left]  {\cite{chang2023lofi}};
    \node (knowndynMethods) [method, right=of knowndyn, xshift=-0.5cm, align=left] {\cite{kalman1960filter}};
    
    % Arrows
    \draw [arrow] (discrete.east) -- (lookback.west);
    \draw [arrow] (discrete.east) -- (mixture.west);
    \draw [arrow] (discrete.east) -- (pastK.west);
    \draw [arrow] (discrete.east) -- (abruptdrift.west);
    
    \draw [arrow] (continuous.east) -- (forgetting.west);
    \draw [arrow] (continuous.east) -- (ppastK.west);
    
    \draw [arrow] (noaux.east) -- (ratechange.west);
    \draw [arrow] (noaux.east) -- (knowndyn.west);
    
    \draw [arrow] (lookback.east) -- (boCD.west);
    \draw [arrow] (mixture.east) -- (mixtureMethods.west);
    \draw [arrow] (pastK.east) -- (pastKMethods.west);
    \draw [arrow] (abruptdrift.east) -- (abruptdriftmethods.west);
    \draw [arrow] (forgetting.east) -- (forgettingMethods.west);
    \draw [arrow] (ppastK.east) -- (probak.west);

    \draw [arrow] (ratechange.east) -- (rateChangeMethods.west);
    \draw [arrow] (knowndyn.east) -- (knowndynMethods.west);
\end{tikzpicture}

%% file: sections/method.tex
\section{The framework}
Consider a sequence of  measurements $\vy_{1:T} = (\vy_1, \ldots, \vy_{T})$ with $ \vy_i \in \real^\dimobs$,
and (unmodelled) features $\vx_{1:T} = (\vx_1, \ldots, \vx_T)$ with $ \vx_i \in\real^{\dimin}$.
Let $\data_t = (\vx_t, \vy_t)$ be a datapoint and $\data_{1:T} = (\data_1, \ldots, \data_T)$ the dataset at time $T$.
Consider a conditional probabilistic model for $\vy_t$, conditioned on $\vx_t$, and parametrised by $\vtheta_t\in\real^\dimstate$, given by
$p(\vy_t \cond \vtheta_t, \vx_t)$, and sometimes called the likelihood.

Assume that the conditional mean for $\vy_t$
is encoded in a parametric model 
$h(\vtheta_{t},\, \vx_{t})$, e.g., a neural network.
A popular modelling choice for regression problems is to assume that  
$p(\vy_t \cond \vtheta_t, \vx_t) = {\cal N}(\vy_{t} \cond h(\vtheta_t,\, \vx_t), \vR_t)$ for known covariance $\vR_t$.
Next, assume there exists an auxiliary variable $\auxv_t \in \vPsi_t$ that
evolves following the dynamics $p(\auxv_t \cond \auxv_{t-1})$ and encodes information about the non-stationarity of the sequence at time $t$.
Here, $\vPsi_t$ is the set of possible values of the auxiliary variable $\auxv_t$.
The purpose of this variable is, for instance,
to determine which past datapoints $\vy_{1:t-1}$ most closely align with the most recent measurement $\vy_t$.
We describe the auxiliary variable in detail in Section \ref{sec:auxiliary-variable}.
Finally, the model parameters $\vtheta_t$ evolve following the dynamics $p(\vtheta_t \cond \vtheta_{t-1}, \auxv_t)$.
This represents how much parameters change, given the state of the auxiliary variable.

Figure \ref{fig:diagram-bone-ssm}
shows the probabilistic graphical model that motivates our formulation; this resembles the one in \cite{Doucet2000}
with an additional optional dependence between the auxiliary variable and the measurements; in what follows we omit this dependence for brevity.
\begin{figure}[htb]
    \centering
    \input{diagrams/bone-ssm}
    \caption{
        Two-levelled hierarchical state-space model (SSM) with known dynamics, motivating our BONE framework
        Solid arrows indicate required dependencies, while dashed arrows represent optional dependencies.
        Rectangles denote exogenous variables, and circles represent random variables. Observed elements are shaded in gray.
        The left shift in $\vx_t$ represents that features are observed before observing $\vy_t$.
    }
    \label{fig:diagram-bone-ssm}
\end{figure}

For an experiment of length $T\in\mathbb{N}$, the joint conditional density over the model parameters,
induced by the graphical model shown in Figure \ref{fig:diagram-bone-ssm}, is given by
\begin{equation}\label{eq:bone-log-joint}
    p(\vy_{1:T}, \vtheta_{0:T}, \auxv_{0:T}  \cond \vx_{1:T})
    = p(\vtheta_0)\,p(\auxv_0)\,\prod_{t=1}^T
    p(\vy_t \cond \vtheta_t, \vx_t)\, p(\vtheta_t \cond \vtheta_{t-1}, \auxv_t)\,p(\auxv_t \cond \auxv_{t-1}).
\end{equation}

In this paper,
we are interested in methods
that efficiently compute the so-called \textit{expected} posterior predictive
$\hat{\vy}_{t+1} := \mathbb{E}_{p(\vtheta_t, \auxv_t \cond \data_{1:t})}[h(\vtheta_t,\vx_{t+1})]$ in an online and recursive manner.
In our setting, one observes $\vx_{t+1}$ just before observing $\vy_{t+1}$;
thus, to make a prediction about $\vy_{t+1}$, we have $\vx_{t+1}$ and $\data_{1:t}$ at our disposal.\footnote{
The  input features $\vx_{t+1}$ and
output measurements $\vy_{t+1}$ can correspond to different time steps.
For example, $\vx_{t+1}$ can be the state of the stock market at a fixed date and
$\vy_{t+1}$ is the return on a stock some days into the future.
}
For the case of a  discrete auxiliary variable $\auxv_t \in \Psi_t$,
the form of the expected posterior predictive for $\vy_{t+1}$, induced by \eqref{eq:bone-log-joint}, is
\begin{equation}\label{eq:bone-posterior-predictive}
\begin{aligned}
    \hat{\vy}_{t+1} &= \mathbb{E}_{p(\vtheta_t, \auxv_t \cond \data_{1:t})}[h(\vtheta_t,\vx_{t+1})]\\
    &= \sum_{\auxv_t \in \Psi_t}\int h(\vtheta_t, \vx_{t+1})\,p(\vtheta_t, \auxv_t \cond \data_{1:t}) \d\vtheta_t\\
    &= \sum_{\auxv_t \in \Psi_t}p(\auxv_t \cond \data_{1:t})
    \int h(\vtheta_t, \vx_{t+1}) p(\vtheta_t \cond \auxv_t, \data_{1:t})\d\vtheta_t,
\end{aligned}
\end{equation}

where
\begin{align}
    p(\vtheta_t \cond \auxv_t, \data_{1:t})
    &\propto p(\vy_t \cond \vtheta_t, \vx_t)\,p(\vtheta_t \cond \auxv_t, \data_{1:t-1}),\label{eq:bone-extra-i} \\
    p(\vtheta_t \cond \auxv_t, \data_{1:t-1})
   &=\int p(\vtheta_t \cond \vtheta_{t-1}, \auxv_t)\,p(\vtheta_{t-1} \cond\data_{1:t-1})\,\d\vtheta_{t-1}, \label{eq:bone-extra-ii}\\
    p(\auxv_t \cond \data_{1:t})
    &=  p(\vy_t \cond \vx_t, \auxv_t, \data_{1:t-1})\,
    \sum_{\auxv_{t-1}\in\vPsi_{t-1}}
    p(\auxv_{t-1} \cond \data_{1:t-1})\,
    p(\auxv_t \cond \auxv_{t-1}, \data_{1:t-1}) \label{eq:bone-extra-iii}.
\end{align}

From \eqref{eq:bone-posterior-predictive}, \eqref{eq:bone-extra-i}, \eqref{eq:bone-extra-ii}, and \eqref{eq:bone-extra-iii}
we argue that there are three key modelling choices and two algorithmic choices.  
Specifically, the three key modelling choices are:
\cModelraw the conditional mean $h(\vtheta, \vx)$ together with the likelihood $p(\vy \cond \vtheta, \vx)$;
\cAuxraw the auxiliary variable $\auxv_t$; and
\cPriorraw the conditional prior $p(\vtheta_t \cond \auxv_t, \data_{1:t-1})$.
Additionally, the two algorithmic choices are:
\cPosteriorraw the algorithm to compute (or approximate)  the conditional posterior over model parameters $p(\vtheta_t \cond \auxv_t, \data_{1:t})$, and
\cWeightraw the algorithm that computes (or approximates) the posterior over weights $p(\auxv_t, \cond \data_{1:t})$.

The BONE framework generalises these choices,
allowing for greater flexibility while maintaining the motivating probabilistic structure.
Instead of the likelihood model $p(\vy_t \cond \vtheta_t, \vx_t)$ with conditional mean $h(\vtheta_t, \vx_t)$,
we consider a general function $\exp(-\ell(\vy_t;\, \vtheta_t, \vx_t))$,
where $\ell(\vy_t; \,\vtheta_t,  \vx_t)$ could be either a loss function or a log-likelihood.
Next, instead of the conditional prior $p(\vtheta_t \cond \auxv_t, \data_{1:t-1})$,
we introduce a more general modelling function, $\pi(\vtheta_t; \auxv_t, \data_{1:t-1})$
that governs the prior over model parameters.\footnote{
This function adopts an ad hoc approach to parameter evolution instead of explicitly solving the integration step \eqref{eq:bone-extra-ii}.
In Appendix \ref{app:full-bayes-prediction},
we detail how the inference is carried out when the dynamics are known. More precisely, we conduct a full-Bayesian treatment by calculating the posterior of model parameters at $t+1$ and using this for inference.
}
Similarly,
instead of the posterior density $p(\vtheta_t \cond \auxv_t, \data_{1:t})$,
we employ the  function $q(\vtheta_t;\, \auxv_t, \data_{1:t})$; e.g., an approximation of the posterior,
or a generalised posterior \citep{bissiri2016generalbayes}.
Finally, instead of the posterior over weights $p(\auxv_t \cond \data_{1:t})$, we 
consider a weighting function $\nu(\auxv_t;\, \data_{1:t})$,
which can be the Bayesian posterior or an ad-hoc time-dependent weighting function.

This generalisation is important because it unifies a wide range of existing methods under a common framework.
Many well-known approaches in the literature can be written as elements of BONE by appropriately selecting the model for measurements, the conditional prior, and the posterior approximations.
BONE highlights connections between different methods,
it also enables systematic comparisons under a common umbrella, 
and it allows us to develop novel algorithms.
Table \ref{tab:modelling-choices-BONE} explicitly contrasts the choices in BONE with those in the
classical Bayesian formalism.
\begin{table}[htb]
    \centering
    \begin{tabular}{l|l|l}
         component & BONE & Bayes \\
         \hline
         \cModel & $h(\vtheta_t, \vx_t)$\,\,\&\,\,$\exp(-\ell(\vy_t; \vtheta_t, \vx_t))$ &  $h(\vtheta_t, \vx_t)$\,\,\&\,\,$p(\vy_t \cond \vtheta_t, \vx_t)$ \\
         \cAux & $\auxv_t$ & $\auxv_t$ \\
         \cPrior & $\pi_t(\vtheta_t;\, \auxv_t) := \pi(\vtheta_t;\, \auxv_t, \data_{1:t-1})$ & $p(\vtheta_t \cond \auxv_t, \data_{1:t-1})$\\
         \cPosterior & $q_t(\vtheta_t; \auxv_t) := q(\vtheta_t;\, \auxv_t, \data_{1:t})$ & $p(\vtheta_t \cond \auxv_t, \data_{1:t})$\\
         \cWeight & $\nu_t(\auxv_t) := \nu(\auxv_t;\, \data_{1:t})$ & $p(\auxv_t \cond \data_{1:t})$\\
    \end{tabular}
    \caption{
        Components of the BONE framework.
    }
    \label{tab:modelling-choices-BONE}
\end{table}

With these modifications, the expected posterior predictive under BONE is
\begin{equation}\label{eq:bone-prediction}
\begin{aligned}
    \hat{\vy}_t
    &:= \sum_{\auxv_t \in \Psi_t}
    \underbrace{\nu(\auxv_t \cond \data_{1:t})}_{\text{\cWeight}}\,
    \int
    \underbrace{h(\vtheta_t, \vx_{t+1})}_{\text{\cModel}}\,
    \underbrace{q(\vtheta_t;\, \auxv_t, \data_{1:t})}_{\text{\cPosterior}}
    \d\vtheta_t,
\end{aligned}
\end{equation}
where
\begin{equation}\label{eq:conditional-posterior}
   q(\vtheta_t;\,\auxv_t, \data_{1:t})
    \propto \underbrace{\pi(\vtheta_t;\, \auxv_t, \data_{1:t-1})}_\text{\cPrior}\,
    \underbrace{\exp(-\ell(\vy_t; \vtheta_t, \vx_t))}_\text{\cModel}
\end{equation}
takes the form of a generalised posterior \citep{bissiri2016generalbayes}.
In classical Bayesian setting, the loss function takes the form of the negative log-likelihood, i.e.,
\begin{equation}\label{eq:loss-function-neg-logl}
    \ell(\vy_t; \vtheta_t, \vx_t) = -\log p(\vy_t \cond \vtheta_t, \vx_t).
\end{equation}
Unless stated otherwise, we work with the negative log-likelihood in \eqref{eq:loss-function-neg-logl}
found in the classical Bayesian setting.

A prediction for $\vy_{t+1}$ given $\data_{1:t}$, $\vx_{t+1}$, and $\auxv_t$ is
\begin{equation}\label{eq:forecast-bone}
    \hat{\vy}_{t+1}^{(\auxv_t)}
    = \mathbb{E}_{q_t}[h(\vtheta_t;\, \vx_{t+1}) \cond \auxv_t]
    := \int h(\vtheta_t;\, \vx_{t+1})\,q(\vtheta_t;\,\auxv_t, \data_{1:t})\d \vtheta_t\,.
\end{equation}  
Here we use the shorthand notation $q_t = q(\vtheta_t;\,\auxv_t, \data_{1:t})$ and
$\mathbb{E}_{q_t}[\cdot \cond \auxv_t]$ to highlight dependence on $\auxv_t$.  

Algorithm \ref{algo:bone-step} provides pseudocode for the prediction and update steps in the BONE framework.
Notably, these components can be broadly divided into two categories: modelling and algorithmic.
The modelling components determine the inductive biases in the model, and correspond to $h$, $\ell$, $\auxv_t$, and $\pi$.
The algorithmic components dictate how operations are carried out to produce a final prediction --- this corresponds to $q_t$ and $\nu_t$.
\begin{algorithm}[H]
\small
\begin{algorithmic}[1]
    \REQUIRE $\data_{1:t}$  // past data
      \REQUIRE $\vx_{t+1}$ // optional inputs
    \REQUIRE $h(\vtheta, \vx_{t})$ // Choice of \cModel
   \REQUIRE $\vPsi_t$ // Choice of \cAux
    \FOR{$\auxv_t\in\vPsi_t$}
        \STATE $\pi_t(\vtheta_t;\, \auxv_t) \gets \pi(\vtheta_t ;\, \auxv_t, \data_{1:t-1})$ // choice of \cPrior
        \STATE $q_t(\vtheta_t;\,\auxv_t) \gets q(\vtheta_t ;\, \auxv_t, \data_{1:t}) \propto \pi_{t}(\vtheta_t\;\auxv_t)\,\exp(-\ell(\vy_t; \vtheta_t, \vx_t))$// choice of \cPosterior
        \STATE $\nu_t(\auxv_t) \gets \nu(\auxv_t ;\, \data_{1:t})$  // choice of \cWeight
        \STATE $\hat{\vy}_{t+1}^{(\auxv_t)} \gets
        \mathbb{E}_{q_t}[h(\vtheta_t, \vx_{t+1}) ;\, \auxv_t]$ // conditional prequential prediction
    \ENDFOR
    \STATE $\hat{\vy}_{t+1} \gets \sum_{\auxv_t}\nu_t(\auxv_t)\,\hat{\vy}_{t+1}^{(\auxv_t)}$ // weighted prequential prediction
\end{algorithmic}
\caption{
    Generic predict and update step of BONE with discrete $\auxv_t$ at time $t$.
}
\label{algo:bone-step}
\end{algorithm}

\subsection{Example tasks that can be solved with BONE}
\label{sec:tasks}

Before going into more detail about BONE,
we give some concrete examples of tasks in which the BONE framework can be applied.
We group these examples into unsupervised tasks and supervised tasks.

\subsubsection{Unsupervised tasks}

Unsupervised tasks 
%correspond to settings where  $\data_t=\vy_t$,
%but there are no input covariates $\vx_t$.
%Thus we need to solve unconditional density modeling
involve estimating unobservable quantities of interest from the data $\data_{1:t}$.
Below, we present three common tasks in this category.

\paragraph{\underline{\texttt{Segmentation}}} 

Segmentation involves partitioning the data stream into contiguous subsequences or ``blocks'' $\{\data_{1:t_1},
\data_{t_1+1:t_2}, \ldots\}$, where  the DGP
for each block
is governed by a sequence of unknown functions \citep{barry1992ppm}.
The goal is to determine the points in time when a new block begins,
known as changepoints.
This is useful in many applications, such as finance, where detecting changes in market trends is critical.
In this setting, non-stationarity is assumed to be abrupt and occurring at unknown points in time.
We study an example in Section \ref{experiment:segmentation}.
For a survey of segmentation methods, see e.g,
\cite{aminikhanghahi2017changepointsurvey,gupta2024changepointsurvey}.

\paragraph{\underline{\texttt{Filtering using state-space models (SSM})}}
Filtering estimates an underlying latent state $\vtheta_t$
that evolves over time (often representing a meaningful concept). 
The posterior estimate of $\vtheta_t$
is computed by applying Bayesian
inference to the corresponding
state space model (SSM),
which determines the choice of \cModel,
and how the state changes over time,
through the choice of \cPrior.
Examples include estimating the state of the atmosphere \citep{evensen1994enkf},
tracking the position of a moving object \citep{battin1982apollo},
or recovering a signal from a noisy system \citep{basseville1993onlinedetection}.
In this setting, non-stationarity is usually
assumed to be continuous and occurring at possible time-varying rates.
For a survey of filtering methods, see e.g., \cite{chen2003bayesianfiltersurvey}.

\paragraph{\underline{\texttt{Segmentation using Switching state-space models (SSSM)}}}
In this task, the modeller extends the standard SSM with a set of discrete latent variables $\psi_t \in \{1,\ldots,K\}$,
which may change value at each time step according to a state transition matrix.
The parameters of the rest of the DGP
depend on the discrete state $\psi_t$.
The objective is to infer the sequence of underlying discrete states that best ``explains'' the observed data
\citep{ostendorf1996hmm,ghahramani2000vsssm,beal2001infinite,fox2007sticky,van2008beam,linderman2017rslds}.
In this context, non-stationarity arises from the switching behaviour of the underlying discrete process.

\subsubsection{Supervised tasks}

Supervised tasks involve predicting a measurable outcome $\vy_t$. 
Unlike unsupervised tasks,
this  allows the performance of the model to be assessed objectively, since we can compare the prediction
to the actual observation.
%Supervised tasks have been the main focus of the machine learning community.
We present three common tasks in this category below.

\paragraph{\underline{\texttt{Prequential forecasting}}}  
Prequential (or one-step-ahead) forecasting \citep{gama2008streamlearn} seeks to predict the value $\vy_{t+1}$ given $\data_{1:t}$ and $\vx_{t+1}$. 
This is distinct from  time-series forecasting, which typically does not consider exogenous variables $\vx_t$,
and thus can forecast (or  ``roll out'')
many steps
into the future.
We study an example in Section \ref{experiment:prequential}.
For a survey on prequential forecasting under non-stationarity, see e.g., \cite{lu2018preqnonstationarysurvey}.

\paragraph{\underline{\texttt{Online continual learning (OCL)}}}  
OCL is 
a broad term used for learning regression or classification models online, typically with neural networks. These methods usually assume that the underlying data generating mechanism could shift.
The objective of OCL methods 
is to train a  model that performs consistently across both past and future data,
rather than just focusing on future forecasting \citep{cai2021ocl}.
The changepoints (corresponding to different ``tasks'') may or may not be known.
This setting addresses the stability-plasticity dilemma,
focusing on retaining previously learned knowledge while adapting to new tasks.
%Given that optimal OCL is an NP-hard problem \citep{knoblauch2020clnphard},
%recent research has focused  on prequential-like forecasting, where the model is continuously adapted to new information.
We study an example of OCL for classification, when the task boundaries are not known, in Section \ref{sec:periodic-drifts}.
For a survey on recent methods for OCL, see e.g., \cite{gunasekara2023surveyocl}.

% Given a sequence of tasks indexed $\{{\cal T}_k\}_{k=1}^K$, with ${\cal T}_k$ a set of indices belonging to the $k$-th task.
% One metric in OCL is to evaluate.
% \begin{equation}
%     M[{\cal T}_k] = \sum_{t \in {\cal T}_k} {\cal L}(\vy_{t}, \hat{\vy}_{t})
% \end{equation}
% \gdm{rewrite}

\paragraph{\underline{\texttt{(Non-stationary) contextual bandits}}}  
In contextual bandit problems, the agent is presented with features $\vx_{t+1}$,
and must choose an action (arm) that yields the highest expected reward \citep{li2010contextualbandits}. We let $\vy_{t+1} \in \mathbb{R}^A$ 
where $A > 2$ is the number of possible actions; this is a vector where the $a$-th entry contains the reward one would have obtained had one chosen arm $a$.
Let $\vy_{t}^{(a)}$ be the observed reward at time $t$ after choosing arm $a$, i.e., the $a$-th entry of $\vy_t$.
A popular approach for choosing the optimal action (while tackling the exploration-exploitation tradeoff) at each step
is  Thomson sampling (TS) \citep{thompson1933sampling}, which in our setting works as follows:
first, sample a parameter
vector from the posterior,
$\tilde\vtheta_{t}$ from $q(\vtheta_{t};\,\auxv_t, \data_{1:t})$; then, 
greedily choose the best arm
(the one with the highest expected payoff),
$a_{t+1} = \argmax_{a} \hat{\vy}_{t+1}^{(a)}$,
 where $\hat\vy_{t+1} = h(\tilde\vtheta_{t}; \vx_{t+1})$;
 and $\hat{\vy}_{t+1}^{(a)}$ is the $a$-th entry of $\hat{\vy}_{t+1}$;
finally,  receive a reward $\vy_{t+1}^{(a_{t+1})}$.
The goal is to select a sequence of arms $\{a_1, \ldots, a_T\}$ that maximises the cumulative reward $\sum_{t=1}^T \vy_{t}^{(a_t)}$.
% If the best possible arm at time $t$ is known, the agent's performance is evaluated using regret:
% \begin{equation}
%     R(T) = \sum_{t=1}^T \max_a \vy_{t}^{(a)} - \sum_{t=1}^T \vy_{t}^{(a_{t})}.
% \end{equation}
% Regret quantifies the difference between the optimal reward and the reward actually accumulated, reflecting the efficiency of the agent’s decisions.
When the mapping function $h$ is a neural network, this model is called a neural bandit.
 TS for neural bandits has been studied in many papers, see e.g., \cite{duran2022neuralbandits}
 and references therein.
Non-stationary bandits 
%in a Bayesian setting
have been studied in 
\cite{mellor2013changepointthompsonsampling,cartea2023bandits,alami2023banditnonstationary, liu2023nonstationarybandits}.
We study an example in Section \ref{experiment:bandits}.

\subsection{Details of BONE}
\label{sec:details}

In the following subsections,
we describe each component of the BONE framework in detail,
provide illustrative examples,
and reference relevant literature for further reading.

\subsection{The  measurement model \cModelraw}
\label{sec:the-task}

Recall that $h(\vtheta, \vx)$ is a parametric model
that encodes the conditional mean for $\vy$, given $\vtheta$ and $\vx$.
For linear measurement models, $h(\vtheta,  \vx)$  is given by:
\begin{equation}
    h(\vtheta, \vx) =
    \begin{cases}
        \vtheta^\intercal \vx & \text{(regression)}, y \in \real \\
        \sigma(\vtheta^\intercal \vx) & \text{(binary classification)}, y \in \{0,1\} \\
        {\rm Softmax}(\vtheta^\intercal \vx) & \text{(multi-class classification)},
        \vy \in \{0,1\}^C
    \end{cases}
\end{equation}
where $\sigma(z) = (1 + \exp(-z))^{-1}$ is the sigmoid function, $C\in\mathbb{N}$ is the number of classes,
${\rm Softmax}(\vz)_k = {\exp(\vz_k)}/{\sum_i \exp(\vz_i)}$ represents the softmax function with
$\vz\in\real^\dimobs$
and $\vz_i$  the $i$-th element of $\vz$.
In the machine learning literature,
the vector $\vz$ is called the logits of
the classifier. 
For non-linear measurement models, such as neural networks, $h(\vtheta, \vx)$ represents the output of the network parameterised by $\vtheta$.
%For example, in a deep neural network used for image classification, $\vtheta$ represents the weights and biases that transform input features $\vx$ into predicted outputs.
The best choice of $h$ will depend on the nature of the data,
as well as the nature of the task,
in particular, whether it is supervised or unsupervised.
We give some examples in 
Section \ref{section:experiments}.

\subsection{The auxiliary variable \cAuxraw }
\label{sec:auxiliary-variable}

The choice of auxiliary variable $\auxv_t$
is crucial to identify changes in the data-generating process,
allowing our framework to track non-stationarity.
Below, we give a list of the common auxiliary variables used in the literature.

% This variable can reflect computations of parameters given a subset of the data $\data_{1:t}$,
% track different \textit{model configurations},
% or weight previous parameters to facilitate faster transfer of information. Below, we explore a number of examples of this component.

\noindent \underline{\texttt{RL}} (runlength):  
    $\auxv_{t} = r_t \in \{0, \ldots, t\}$ is a scalar representing a \textit{lookback window}, defined as the number of steps since the last regime change.  
    The value $r_t = 0$ indicates the start of a new regime at time $t$,
    while $r_t \geq 1$ denotes the continuation of a regime with a lookback window of length $r_t$.  
    This choice of auxiliary variable is common in the changepoint detection literature.
    See e.g., \cite{adams2007bocd,
    knoblauch2018doublyrobust-bocd, alami2020restartedbocd, agudelo2020bocdprediction, altamirano2023robust, alami2023banditnonstationary}.
    This auxiliary variable is useful for non-stationary data with non-repeating temporal segments,
    provided we know the intensity with which new segments appear.

\noindent \underline{\texttt{RLCC}} (runlength and changepoint count):  
    $\auxv_{t} = (r_t, c_t) \in \{0, \ldots, t\} \times \{0, \ldots, t\}$ is a vector that represents both the runlength and the total number of changepoints,
    as proposed in \cite{wilson2010-bocd-hazard-rate}.
    When $r_t = t$, this implies $c_t = 0$, meaning no changepoints have occurred.  
    Conversely, $r_t = 0$ indicates the start of a new regime and implies $c_t \in \{1, \ldots, t\}$, accounting for at least one changepoint.  
    For a given $r_t\geq 0$, the changepoint count $c_t$ belongs to the range $\{1, \ldots, t - r_t\}$.  
    %(Tracking the count can be useful for performing online estimation of the rate at which changepoints occur.)
    As with \texttt{RL}, this auxiliary variable assumes consecutive time blocks, but additionally allows us to estimate the likelihood of entering a new regime by tracking the number of changepoints seen so far. 
    This auxiliary variable  is useful for non-stationary data with non-repeating temporal segments when the intensity with which new segments appear is unknown.

\noindent \underline{\texttt{CPT}} (changepoint timestep):
    $\auxv_t = \vzeta_t$, with $\vzeta_t = \{\zeta_{1,t}, \ldots, \zeta_{\ell,t}\}$,
    is a set of size $\ell \in \{0, \ldots t\}$ containing the $\ell$ times at which there was a changepoint,
    with the convention that  $0 \leq \zeta_{1,t} < \zeta_{2,t} < \ldots < \zeta_{\ell,t} \leq t$.
    This choice of auxiliary variable was introduced in \cite{fearnhead2007line} and has been studied in
    \cite{fearnhead2011adaptivecp, fearnhead2019robustchangepoint}.
    Under mild assumptions, it can be shown that \texttt{CPT} is equivalent to \texttt{RL},
    see e.g., \cite{knoblauch2018varbocd}.
    This auxiliary variable  is useful for non-stationary data with non-repeating temporal segments
    when the probability of a new segment appearing is unknown and knowledge of the changepoint location is required.

\noindent \underline{\texttt{CPL}} (changepoint location):  
    $\auxv_t=s_{1:t}\in\{0,1\}^t$ is a binary vector. In one interpretation,
    $s_i=1$ indicates the occurrence of a changepoint at time $i$,
    as in \cite{li2021onlinelearning},
    while in another, it means that $\data_t$ belongs to the current regime, as in \cite{nassar2022bam}.
    This auxiliary variable  is useful for non-stationary data with repeating temporal segments.
    It is useful when the segments are formed of non-consecutive datapoints.

\noindent \underline{\texttt{CPV}} (changepoint probability vector):  
    $\auxv_{t}=v_{1:t}\in(0,1)^t$ is a $t$-dimensional random vector representing the probability of each element in the history belonging to the current regime.  
    This generalises \texttt{CPL} and was introduced in \cite{nassar2022bam} for online continual learning, allowing for a more fine-grained representation of changepoints over time.
    This auxiliary variable is useful for  non-stationary data with repeating temporal segments.
    Unlike \texttt{CPL},
    it takes a vector of weights in $(0,1)$
    which allows for higher flexibilty when compared to \texttt{CPL}.

\noindent \underline{\texttt{CPP}} (changepoint probability):  
    $\auxv_t = \upsilon_t\in(0,1)$ represents the probability of a changepoint.  
    This is a special case of \texttt{CPV} that tracks only the most recent changepoint probability;
    this choice
    was used in \cite{titsias2023kalman} for online continual learning.

\noindent \underline{\texttt{ME}} (mixture of experts):  
    $\auxv_{t} =\alpha_t \in \{1, \ldots, K\}$ represents one of  $K$ experts.
    Each expert corresponds to either a choice of model
    or one of $K$ possible hyperparameters.  
    This approach has been applied to filtering \citep{chaer1997mixturekf} and
    prequential forecasting \citep{liu2023bdemm, abeles2024adaptive}.
    This auxiliary variable facilitates the weighting of predictions made by models
    when one has a fixed number of competing models.

\noindent \underline{\texttt{C}}:  
    $\auxv_t = c$ represents a constant auxiliary variable, where $c$ is just a placeholder or dummy value.
    This is equivalent to not having an auxiliary variable,
    or alternatively, to having a single expert that encodes all available information.

\paragraph{Space-time complexity}
There is a tradeoff between the complexity that $\auxv$ is able to encode and the computation power needed to perform updates. Loosely speaking, this can be seen in the cardinality of the set of possible values of $\auxv$ through time.
Let  $\vPsi_t$ be the space of possible values for $\auxv_t$.
Depending on the choice of $\auxv_t$,
the cardinality of $\vPsi_t$  either stay constant or increase over time, i.e., 
 $\vPsi_{t-1} \subseteq \vPsi_t$ for all $t=1,\ldots,T$.
For instance,
the possible values for \texttt{RL} increase by one at each timestep;
the possible values of $\texttt{CPL}$ double at each  timestep; finally,
the possible values for $\texttt{ME}$ do not increase. 
Table \ref{tab:auxv-time-complexity}
shows the space of values and cardinality that $\vPsi_t$ takes as a function of the choice of auxiliary variable.
% We give some concrete examples in \cref{section:experiments}.

\begin{table}[htb]
    \footnotesize
    \centering
    \begin{tabular}{c|cccccccc}
    name & \texttt{C} & \texttt{CPT} & \texttt{CPP} & \texttt{CPL} & \texttt{CPV} & \texttt{ME} & \texttt{RL} & \texttt{RLCC} \\
    %space-time ($\vPsi_t$) 
    values & $\{c\}$ & $2^{\{0, 1, \ldots, t\}}$ & $[0,1]$ & $\{0,1\}^t$ & $(0,1)^t$ & $\{1, \ldots, K\}$ & $\{0, 1, \ldots, t\}$ & $\{\{0, t\}, \ldots, \{t, 0\}\}$\\
    cardinality & $1$ & $2^t$ & $\infty$ & $2^t$ & $\inf$ & $K$ & $t$ & $2+ t(t+1)/2$
    \end{tabular}
    \caption{
    %Space-time complexity of auxiliary random variables $\auxv_t$.
    Design space for the auxiliary random variables $\auxv_t$.
    Here,
    $T$ denotes the total number of timesteps and
    $K$ denotes a fixed number of candidates.
    }
    \label{tab:auxv-time-complexity}
\end{table}

\subsection{Conditional prior \cPriorraw}
\label{sec:conditional-prior}

This component defines the prior predictive distribution over model parameters
conditioned on the choice of \cAux $\auxv_t$ and the dataset $\data_{1:t-1}$.
In some cases, explicit access to past data is not needed.
% For example, we may summarize it in terms of
% a mean and covariance over the parameters (corresponding to hyper-parameters of the prior), or a choice from a finite set of possible past candidate values.

%\subsubsection{Gaussian approximations}
For example, a common assumption  is to have a Gaussian conditional prior over model parameters.
In this case, we assume that, given data $\data_{1:t-1}$ and the auxiliary variable $\auxv_t$,
the conditional prior takes the form
\begin{equation}\label{eq:conditional-prior-gaussian}
    \pi(\vtheta_t;\, \auxv_t,\, \data_{1:t-1}) =
    {\cal N}\big(\vtheta_t\cond g_{t}(\auxv_t, \data_{1:t-1}), G_{t}(\auxv_t, \data_{1:t-1})\big),
\end{equation}
with
$g_{t}:\vPsi_t\times \real^{(\dimstate+\dimobs)(t-1)} \to \real^\dimstate$ a function that returns the mean vector of model parameters,
$\mathbb{E}[\vtheta_t \cond \auxv_t, \data_{1:t-1}]$,
and
$G_t:\vPsi_t\times \real^{(\dimstate+\dimobs)(t-1)}  \to \real^{\dimstate\times\dimstate}$ 
a function that returns a $\dimstate$-dimensional covariance matrix,
${\rm Cov}[\vtheta_t \cond \auxv_t, \data_{1:t-1}]$.
In what follows, we let $(\vmu_0, \vSigma_0)$ be the pre-defined initial prior mean and covariance.
Furthermore, we denote $(\vmu_{t-1}, \vSigma_{t-1})$ be the posterior mean and covariance found at time $t-1$,
which is used as a prior at time $t$.

Below, we provide a non-exhaustive list of possible combinations of
choices for \cAux and \cPrior of the form \eqref{eq:conditional-prior-gaussian}
that can be found in the literature, and we also introduce a new combination.

\noindent \underline{\texttt{C-LSSM}}  (constant linear with affine state-space model).
We assume the parameter dynamics can be modeled by a linear-Gaussian state space model (LSSM),
i.e., 
$\mathbb{E}[\vtheta_t \cond \vtheta_{t-1}] = \vF_t\,\vtheta_{t-1}
+ \vb_t$
and
${\rm Cov}[\vtheta_t \cond \vtheta_{t-1}] = \vQ_t$,
for given $(\dimstate\times\dimstate)$ dynamics matrix $\vF_t$,
$(\dimstate\times1)$ bias vector $\vb_t$,
and $(\dimstate\times\dimstate)$ positive semi-definite matrix $\vQ_t$.
We also assume
$\auxv_t = c$ is a fixed (dummy) constant,
which is equivalent to not having an auxiliary variable.
The characterisation of the conditional prior takes the form
\begin{equation}
\begin{aligned}
    g_t(c, \data_{1:t-1}) &= \vF_t\,\vmu_{t-1}
    + \vb_t,\\
    G_t(c, \data_{1:t-1}) &= \vF_t\,\vSigma_{t-1} \vF_t^\intercal + \vQ_t\,,
\end{aligned}
    \label{eqn:LSSM}
\end{equation}
This is a common baseline model that we will specialise below.

\noindent \underline{\texttt{C-OU}}  (constant with Ornstein-Uhlenbeck process).
This is a special case of the \texttt{C-LSSM} model
where
$\vF_t = \gamma \vI$, 
$\vb_t = (1-\gamma) \vmu_0$,
$\vQ_t = (1 - \gamma^2) \vSigma_0$,
$\vSigma_0=\sigma_0^2 \vI$,
$\gamma \in [0,1]$ is the fixed rate, and
$\sigma_0 \geq 0$.
% \kpm{
% If $0 < \gamma < 1$, then if we unroll the process for many steps in the absence of any observations,
% the posterior $p(\vtheta_T|\vtheta_t)$
% will revert to the prior $p(\vtheta_0)$,
% which we assume corresponds to a ``plastic'' region in which the network weights are able to learn.
% Replacing $\vQ_t$ with an arbitrary variance 
% term $\sigma_t^2$ would result in the variance of $p(\vtheta_T|\vtheta_t)$ 
% going to 0 or $\infty$.
% }
The conditional prior mean
and covariance are
a convex combination of the form
\begin{equation}
\label{eq:cprior-c-ou}
\begin{aligned}
    g(c, \data_{1:t-1}) &=  \gamma \vmu_{t-1} + (1 - \gamma) \vmu_0, \\
    G(c, \data_{1:t-1}) &=  \gamma^2 \vSigma_{t-1} + (1 - \gamma^2) \vSigma_0.
\end{aligned}
\end{equation}
This combination is used in
\cite{kurle2019continual}.
Smaller values of the rate parameter $\gamma$
correspond to a faster resetting,
i.e., the distribution of model parameters
revert more quickly to the prior belief
$(\vmu_0,\vSigma_0)$,
which means the past data will be forgotten.

\noindent \underline{\texttt{CPP-OU}}  (changepoint probability with Ornstein-Uhlenbeck process).
Here 
$\auxv_t=\upsilon_t \in [0,1]$
is the changepoint probability that we use as
the rate of an Ornstein-Uhlenbeck (OU) process,
as proposed in
\cite{titsias2023kalman,Galashov2024}.
% (\kpm{They use a vector of drift parameters, $\vgamma_t$, one per dimension of $\vtheta_t$.})
The characterisation of the conditional prior takes the form
\begin{equation}
\label{eq:cprior-cpp-d}
\begin{aligned}
    g(\upsilon_t, \data_{1:t-1}) &= \upsilon_t \vmu_{t-1} + (1 - \upsilon_t)  \vmu_0 \,,\\
    G(\upsilon_t, \data_{1:t-1}) &=  \upsilon_t^2 \vSigma_{t-1} + (1 - \upsilon_t^2)  \vSigma_0\,.
\end{aligned}
\end{equation}
An example on how to compute $\upsilon_t$
using an empirical Bayes procedure
is given in
\eqref{eqn:nut}.

\noindent \underline{\texttt{C-ACI}} (constant with additive covariance inflation).
This corresponds to a special case of \texttt{C-LSSM} in which $\vF=\vI$,
$\vb=\vzero$, and $\vQ=\alpha \vI$
for $\alpha>0$ is the amount of noise
added at each step.
This combination is used in
\cite{kuhl1990ridge,duran2022neuralbandits,chang2022diagonal,chang2023lofi} .
The characterisation of the conditional prior takes the form
\begin{equation}\label{eq:cprior-c-aci}
\begin{aligned}
    g(c,  \data_{1:t-1}) &= \vmu_{t-1},\\
    G(c, \data_{1:t-1}) &= \vSigma_{t-1} + \vQ_t.
\end{aligned}
\end{equation}
This is similar to \texttt{C-OU} with $\gamma=1$,
however, here we inject new noise at each step. % and do not revert back to prior beliefs.
%Thus, the process drifts with covariance $\vQ_t$.
Another variant of this scheme,
known as \textit{shrink-and-perturb}
\citep{Ash2020},
takes 
$g(c,  \data_{1:t-1}) = q\,\vmu_{t-1}$
and
$G(c, \data_{1:t-1}) = \vSigma_{t-1} + \vQ_t$, where $0 < q < 1$
is the shrinkage parameters,
and $\vQ_t=\sigma_0^2\,\vI$.

\noindent \underline{\texttt{C-Static}}  (constant with static parameters).
Here  $\auxv_t = c$ (with $c$ a dummy variable).
This is a special case of the \texttt{C-ACI} configuration in which $\alpha=0$.
The conditional prior is characterised by
\begin{equation}\label{eq:cprior-c-f}
\begin{aligned}
    g_t(c, \data_{1:t-1}) &= \vmu_{t-1},\\
    G_t(c, \data_{1:t-1}) &= \vSigma_{t-1}.
\end{aligned}
\end{equation}
\eat{
This
 corresponds to the classic Bayesian update rule for static parameters, namely
\begin{equation}
\begin{aligned}
    p(\vtheta_t \cond \auxv_t, \data_{1:t}) = p(\vtheta_t \cond  \data_{1:t})
    %&\propto p(\vy_t \cond \vtheta_t, \vx_t)\,\pi(\vtheta_t \cond  \data_{1:t-1})\\
    %&=
    & \propto p(\vy_t \cond \vtheta_t, \vx_t)\,{\cal N}(\vtheta_t \cond \vmu_{t-1}, \vSigma_{t-1}).
\end{aligned}
\end{equation}
}

\noindent \underline{\texttt{ME-LSSM}}  (mixture of experts with LSSM).
Here $\auxv_t = \alpha_t \in \{1,\ldots, K\}$,
and we have a bank of $K$ independent
LSSM models; the auxiliary variable specifies which model to use at each step.
The characterisation of the conditional prior takes the form
\begin{equation}
\begin{aligned}
    g_t(\alpha_t, \data_{1:t-1}) &= \vF^{(\alpha_t)}_t\,\vmu_{t-1}^{(\alpha_t)}
    + \vb_t^{(\alpha_t)}\,,
    \\
    G_t(\alpha_t, \data_{1:t-1}) &= \vF^{(\alpha_t)}_t\,\vSigma_{t-1}^{(\alpha_t)}\vF_t^\intercal + \vQ_t^{(\alpha_t)}\,.
\end{aligned}
    \label{eqn:MELSSM}
\end{equation}
%which corresponds to the mean and covariance matrix of
%$\mathbb{E}[\vtheta_{t} \cond \data_{1:t-1}]$ and
%${\rm Var}(\vtheta_t \cond \data_{1:t:-1})$.
The superscript $(\alpha_t)$ denotes the conditional prior for the $k$-th expert. More precisely, $\vmu_{t-1}^{(\alpha_t)},\vSigma_{t-1}^{(\alpha_t)}$ are the posterior at time $t-1$ using $\vF^{(\alpha_t)}_{t-1}$ and $\vQ_{t-1}^{(\alpha_t)}$ from the $k$-th expert.
This combination was introduced in \cite{chaer1997mixturekf}.

\noindent \underline{\texttt{RL-PR}}  (runlength with prior reset):
for $\auxv_t = r_t$, this choice of auxiliary variable constructs a new mean and covariance
considering the past $t - r_t$ observations. We have
\begin{equation}\label{eq:cprior-rl-pr}
\begin{aligned}
    g_t(r_t, \data_{1:t-1}) &= \vmu_0\,\mathds{1}(r_t  = 0) + \vmu_{(r_{t-1})}\mathds{1}(r_t > 0),\\
    G_t(r_t, \data_{1:t-1}) &= \vSigma_0\,\mathds{1}(r_t  = 0) + \vSigma_{(r_{t-1})}\mathds{1}(r_t > 0),\\
\end{aligned}
\end{equation}
where  $\vmu_{(r_{t-1})}, \vSigma_{(r_{t-1})}$ denotes the posterior belief computed using observations
from indices $t - r_t$ to $t - 1$.
The case $r_t = 0$ corresponds to choosing the initial pre-defined prior mean and covariance $\vmu_0$ and $\vSigma_0$.
This combination assumes that data from a single regime arrives in sequential \textit{blocks} of time
of length $r_t$.
This choice of \cPrior was first studied in \cite{adams2007bocd}.

\noindent  \underline{\texttt{RL[1]-OUPR*}}  (greedy runlength with OU and prior reset):
\label{sec:SPR}
This is a new combination we consider in this paper,
which is designed to accommodate both gradual changes and sudden changes. More precisely, we assume
$\auxv_t = r_t$, and
we choose the conditional prior as 
either a hard
reset to the prior,
if $\nu_t(r_t) > \varepsilon$,
%(if the posterior weight $\nu(r_t) = p(r_t %\cond \data_{1:t})$, for $r_t \geq 1$,
% is below a threshold $\varepsilon$),
or a convex combination
of the prior and the previous belief state (using an \texttt{OU} process),
if $\nu_t(r_t) \leq \varepsilon$.
That is,
we define the conditional prior as
\begin{equation}
\label{eq:SPR-equation-gt}
    g_t(r_t, \data_{1:t-1}) =
    \begin{cases}
        \vmu_0\,(1 - \nu_t(r_t)) + \vmu_{(r_t)}\,\nu_t(r_t) & \nu_t(r_t) > \varepsilon,\\
        \vmu_0 & \nu_t(r_t) \leq \varepsilon,
    \end{cases}
\end{equation}
\begin{equation}
\label{eq:SPR-equation-Gt}
   G_t(r_t, \data_{1:t-1}) =
    \begin{cases}
        \vSigma_0\,(1 - \nu_t(r_t)^2) + \vSigma_{(r_t)}\,\nu_t(r_t)^2 & \nu_t(r_t) > \varepsilon,\\
        \vSigma_0 & \nu_t(r_t) \leq \varepsilon.
    \end{cases}
\end{equation}
Here 
$\nu_t(r_t)=p(r_t \cond \data_{1:t})$, with $r_t=r_{t-1}+1$,
is the probability we are continuing a segment, and $\nu_t(r_t)$
with $r_t = 0$ is the probability of a changepoint.
For details on how to compute $\nu_t(r_t)$,
see  \eqref{eq:rl-oupr-weight}.
%is a  function of $r_t$ and corresponds to the choice of \cWeight; see Section \ref{sec:choice-weighting-function}.
%In \eqref{eq:SPR-equation-gt}, the convex combination is between of the posterior mean and the (initial) prior mean
%weighted by the choice of \cWeight. 
The value of the threshold parameter $\varepsilon$ controls whether an abrupt change or a gradual change should take place.
In the limit when $\varepsilon =1$, this new combination does not learn, 
since it is always doing a hard
reset to the initial beliefs at time $t=0$.
Conversely, when $\varepsilon=0$, we obtain an OU-type update weighted by $\nu_t$.
When $\varepsilon=0.5$, we revert back to prior beliefs when the most likely hypothesis is that a changepoint has just occurred. 
Finally, we remark that the above combination allows us to make use of non-Markovian choices for \cModel, as we see in Section \ref{experiment:KPM}.
This is, to the best of our knowledge, a new combination that has not been proposed
in the previous literature;
for further details see Appendix \ref{sec:rl-spr-implementation}.

%Below, we see that the comparison between $\nu(r_t)$ and $\varepsilon$ resembles that of likelihood ratio test; see \eqref{eq:rl-spr-eq-posterior}.
% \gdm{
% Be more informative on this combination!!
% Likelhood ration test is more informative that empirical Bayes (?) / keep track of runlength, we do cross
% Talk about threshold choosng.
% + "In our experiments, we observe that the likelihood ratio test is empirically more robust than the marginal likelihood used by EB".
% Choosing $\varepsilon = 1.0$ corresponds to a \texttt{CPP-OU} with a weight that depends on the runlenght.
% As we see in Section \ref{experiment:KPM}, this allows us to make use of non-Markovian choices for \cModel.
% }

\noindent \underline{\texttt{CPL-Sub}} (changepoint location with subset of data):
for $\auxv_t = s_{1:t}$,
this conditional prior constructs the mean and covariance  as
\begin{equation}
\begin{aligned}
    g_t(s_{1:t}, \data_{1:t-1}) &= \vmu_{(s_{1:t-1})},\\
    G_t(s_{1:t}, \data_{1:t-1}) &= \vSigma_{(s_{1:t-1})},
\end{aligned}
\end{equation}
where  $\vmu_{(s_{1:t-1})}, \vSigma_{(s_{1:t-1})}$  denotes the posterior belief computed using
the observations for entries where $s_{1:t-1}$ have value of $1$.
This combination assumes that data from the current regime is scattered from the past history.
That is, it assumes that data from a past regime could become relevant again at a later date.
This combination has been studied in \cite{nguyen2017vcl}.

\noindent \underline{\texttt{CPL-MCI}}  (changepoint location with multiplicative covariance matrix):
for  $\auxv_t = s_{1:t}$, 
%this choice of conditional prior reconstructs the prior mean and covariance using all information.
this choice of conditional prior maintains the prior mean, but increases
the norm of the prior covariance by a constant term $\beta \in (0,1)$. More precisely, we have that
\begin{equation}
\begin{aligned}
    g_t(s_{1:t}, \data_{1:t-1}) &= \vmu_{(s_{1:t-1)}},\\
    G_t(s_{1:t}, \data_{1:t-1}) &=  
    \begin{cases}
        \beta^{-1}\vSigma_{(s_{1:t-1})}  & s_t = 1\,,\\
        \vSigma_{(s_{1:t-1})}  & s_t = 0\,.
    \end{cases}
    %\beta^{-1}\vSigma_{(s_{1:t-1})}\mathds{1}(s_t =1) + \vSigma_{(s_{1:t-1})}\mathds{1}(s_t = 0).
\end{aligned}
\end{equation}
This combination was first proposed in \cite{li2021onlinelearning}.

\noindent \underline{\texttt{CPT-MMPR}}  (changepoint timestep using moment-matched prior reset):
% \kpm{This is so complicated, should we move to appendix?}
for  $\auxv_t = \vzeta_{t}$, with $\vzeta_t = \{\zeta_{1,t}, \ldots, \zeta_{\ell,t}\}$,
and $\zeta_{\ell,t}$ the position of the last changepoint,
the work of \cite{fearnhead2011adaptivecp}
assumes a dependence structure between changepoints.
That is, to build the conditional prior mean and covariance,  they consider the past $\data_{\zeta_{\ell,t}:t-1}$
datapoints whenever $\zeta_{\ell,t} \leq t-1$
and a moment-matched approximation to the mixture density over all possible subset densities since the last changepoint
whenever $\zeta_{\ell,t} = t$.
For an example of \texttt{MMPR} with \texttt{RL} choice of \cAux, see Appendix \ref{sec:rl-mmpr-implementation}.
% We come back to this point in Section \ref{sec:choice-weighting-function}.

% In a number of the examples above, the mean and covariance do not need to be re-computed at every timestep.
% Rather, they are updated using the previous mean $\vmu_{t-1}$ and previous covariance $\vSigma_{t-1}$.

% \subsubsection{Non-Bayesian Non-Gaussian}
% Whenever $q_{\data_{1:t}}(\d \vtheta_t;\,\auxv_t)$ is of the form \eqref{eq:measure-params-loss-based} then,
% the conditional prior takes the form of the loss function, which is modulated by $\auxv_t$.
% \begin{equation}
%     {\cal L}(\vtheta;\,\auxv_t, \data_{1:t+1})
% \end{equation}

% \subsubsection{Other assumptions}
% % I worry that by including non-Bayesian components, we might need to expand the scope
% In a non-Bayesian setting,

% \gdm{Add shirnk and perturb}
% shrink-and-perturb method of \cite{ash2020shrinkperturb}

% p(theta(t+1) | D(1:t)) = theta*(t) wp p(t) or theta(0) wp 1-p(t)
% p(t) = adaptive rule
% \gdm{Add continual backprop as inflatin parameters}
% ``continual backpropagation'' method of \cite{dohare2024continualbackprop}: add noise to the model parameters.

\subsection{Algorithm to compute the posterior over model parameters \cPosteriorraw}
\label{sec:A1}

This section presents algorithms for estimating the density $q(\vtheta_{t};\, \auxv_{t}, \data_{1:t})$;
we focus on methods that yield Gaussian posterior densities.
Specifically, we are interested in practical approaches for approximating the conditional Bayesian posterior,
as given in \eqref{eq:conditional-posterior}.

There is a vast body of literature on methods for estimating the posterior over model parameters. %a full review is beyond the scope of this paper.
% We refer interested readers to \gdm{\cite{XXX}} for a comprehensive discussion.
Here, we focus on three common approaches for computing the Gaussian posterior:
 conjugate updates (\texttt{Cj}), 
 linear-Gaussian approximations (\texttt{LG}),
 and variational Bayes (\texttt{VB}).
 For an overview of choices of \cPosterior
 and a comparison in terms of their computational complexity, see
 Table 3 in \cite{jones2024bong}.
 
\subsubsection{Conjugate updates (\texttt{Cj})}

Conjugate updates (\texttt{Cj}) provide a classical approach for computing the posterior by leveraging conjugate prior distributions.
Conjugate updates occur when the functional form of the conditional prior $\pi(\vtheta_t; \auxv_t, \data_{1:t-1})$
matches that of the measurement model $p(\vy_t \cond \vtheta, \vx_t)$ \citep[Section 3.3]{robert2007bayesianbook}.
This property allows the posterior to remain within the same family as the prior,
which leads to analytically tractable updates and facilitates efficient recursive estimation.

A common example is the conjugate Gaussian model, where the measurement model is Gaussian with known variance and the prior is a multivariate Gaussian. This results in closed-form updates for both the mean and covariance. Another example is the Beta-Bernoulli pair, where the measurement model follows a Bernoulli distribution with an unknown probability, and the prior is a Beta distribution.
See e.g., \cite{Bernardo94,West97} for details.

The recursive nature of conjugate updates makes them particularly useful for real-time or sequential learning scenarios, where fast and efficient updates are crucial.

\subsubsection{Linear-Gaussian approximation (\texttt{LG}) }

The linear-Gaussian (\texttt{LG}) method builds on the conjugate updates (\texttt{Cj}) above.
More precisely, the prior is Gaussian and the measurement model is approximated by a linear Gaussian model,
which simplifies computations.
% where both the measurement model and the prior are Gaussian.
% \texttt{LG} assumes a linear relationship between the parameters and the observations,
% which simplifies computation.
% In the nonlinear setting, we may perform approximate inference by linearizing the model, and then performing the updates. 

The prior over model parameters is taken as:
\begin{equation}
\pi(\vtheta_t;\, \auxv_t, \data_{1:t-1}) = \mathcal{N}\left(\vtheta_t \cond \vmu_{t-1}^{(\auxv_t)}, \vSigma_{t-1}^{(\auxv_t)}\right),
\end{equation}
where $\vmu_{t-1}^{(\auxv_t)}$ and $\vSigma_{t-1}^{(\auxv_t)}$ are the mean and covariance, respectively.
We use the measurement function $h$ to define a first-order approximation $\bar{h}_t$ around the prior mean which is given by
\begin{equation}
    \bar{h}_t(\vtheta_t, \vx_t) = h\left(\vmu_{t-1}^{(\auxv_t)}, \vx_t\right) +
    \vH_t\,\left(\vtheta_t - \vmu_{t-1}^{(\auxv_t)}\right)\,.
\end{equation}
Here, $\vH_t$ is the Jacobian of $h(\vtheta,\vx_t)$ with respect to $\vtheta$, evaluated at $\vmu_{t-1}^{(\auxv_t)}$.
The approximate posterior measure is given by
\begin{equation}
\begin{aligned}
    q(\vtheta_t; \auxv_t, \data_{1:t})%\d\vtheta_t
    &\propto \mathcal{N}(\vy_t \cond \bar{h}_t(\vtheta_t, \vx_t), \vR_t)\,\pi(\vtheta_t;\, \auxv_t,\data_{1:t-1})%\d\vtheta_t
    \\
    &= \mathcal{N}(\vy_t \cond \bar{h}_t(\vtheta_t, \vx_t), \vR_t)\,\mathcal{N}\left(\vtheta_t \cond \vmu_{t-1}^{(\auxv_t)}, \vSigma_{t-1}^{(\auxv_t)}\right)%\d\vtheta_t
    \\
    &\propto \mathcal{N}(\vtheta_t \cond \vmu_t^{(\auxv_t)}, \vSigma_t^{(\auxv_t)}),%\d\vtheta_t,
\end{aligned}
\end{equation}
where  $\vR_t$ is a known noise covariance matrix of the measurement $\vy_t$.
Under the \texttt{LG} algorithmic choice, the updated equations are
\begin{equation}\label{eq:ekf-update-step}
\begin{aligned}
    \hat{\vy}_t^{(\auxv_t)} &= h\left(\vmu_{t-1}^{(\auxv_t)}, \vx_t\right),\\
    \vS_t^{(\auxv_t)} &= \vH_t\,\vSigma_{t-1}^{(\auxv_t)}\,\vH_t^\intercal + \vR_t,\\
    \vK_t^{(\auxv_t)} &= \vSigma_{t-1}^{(\auxv_t)}\,\vH_t^\intercal\,\left(\vS_t^{(\auxv_t)}\right)^{-1},\\
    \vmu_t^{(\auxv_t)} &= \vmu_{t-1}^{(\auxv_t)} + \vK_t^{(\auxv_t)}\left(\vy_t - \hat{\vy}_t^{(\auxv_t)}\right),\\
    \vSigma_t^{(\auxv_t)} &=
    \vSigma_{t-1}^{(\auxv_t)} - \left(\vK_t^{(\auxv_t)}\right)\,\left(\vS_t^{(\auxv_t)}\right)\,\left(\vK_t^{(\auxv_t)}\right)^\intercal.
\end{aligned}
\end{equation}
This linear approximation  enables efficient computation of the posterior in a Gaussian form.
Examples include the extended Kalman filter (EKF) \citep{haykin2004ekfnnet},
which applies local linearisation to non-linear systems,
the exponential family EKF \citep{olllivier2018expfamekf},
which approximates the measurement model as Gaussian by matching the first two moments, and
the low-rank Kalman filter (LoFi) method \citep{chang2023lofi},
which assumes a diagonal-plus-low-rank  (DLR) posterior precision matrix.
See \cite{sarkka2023filtering}
for more details on such Gaussian filtering methods.

\subsubsection{Variational Bayes (\texttt{VB})}

Variational Bayes (VB) is a popular method for approximating a posterior distribution of model parameters by choosing a parametric
family (such as Gaussians) that is 
computationally tractable.
 The primary objective of VB is to minimise the Kullback-Leibler (KL) divergence between a candidate Gaussian distribution and the density $q_t$.
 It can be shown that we can safely
 ignore  the normalisation constant for $q_t$, which is often computationally expensive,
 so we can replace $q_t$
 with its unnormalised form.
 We have the following optimisation problem
 for the posterior variational parameters:
\begin{equation}\label{eq:variational-bayes-objective}
    (\vmu_t, \vSigma_t) =
    \argmin_{\vmu, \vSigma}
    {\boldsymbol{\rm D}}_{\rm KL}
    \left(\mathcal{N}(\vtheta_t \cond \vmu, \vSigma) \,\|\, p(\vy_t \cond \vtheta_t, \vx_t)\,\pi(\vtheta_t;\,\auxv_t, \data_{1:t-1})\right),
\end{equation}
where
$\pi_t(\vtheta_t;\,\auxv_t)$ is the chosen prior distribution \cPrior.

An example of VB for neural network models is the Bayes-by-backpropagation method (BBB) of \cite{blundell2015bbb},
which assumes a diagonal posterior covariance
(more expressive forms are also possible).
 \cite{nguyen2017vcl} extended BBB to
 non-stationary settings.
More recent approaches involve recursive estimation,
such as
the recursive variational Gaussian approximation (R-VGA) method of \cite{lambert2022rvga} which uses a full rank Gaussian variational approximation;
the low-rank RVGA (L-RVGA) method of  \cite{lambert2023lrvga}, which uses a diagonal plus low-rank  (DLR) Gaussian variational approximation;
the Bayesian online natural gradient (BONG) method of  \cite{jones2024bong},
which combines the DLR approximation with EKF-style linearisation for additional speedups;
the natural gradient Gaussian approximation (NANO) method of \cite{Cao2024}, which uses a diagonal Gaussian approximation
similar to VD-EKF in \cite{chang2022diagonal};
and the projection-based unification of last-layer and subspace estimation (PULSE) method of \cite{cartea2023sharpbayes},
which targets different posterior densities for a subspace of the hidden layers and a full-rank covariance over the final layer of a neural network.

\subsubsection{Alternative methods}

Alternative approaches for handling nonlinear 
or nonconjugate measurements have been proposed,
such as sequential Monte Carlo (SMC) methods \citep{freitas2000smcneuralnets},
and ensemble Kalman filters (EnKF) \citep{Roth2017enkf}.
These sample-based methods are particularly advantageous when the dimensionality of $\vtheta$
is large,
or when a more exact posterior approximation is required,
providing greater flexibility in non-linear and non-Gaussian environments.

Generalised Bayesian methods, such as \cite{mishkin2018slang,knoblauch2022gvi}, generalise the VB update
of \eqref{eq:variational-bayes-objective}
by allowing the right-hand side to be a loss function.
Alternatively, online gradient descent  methods like \cite{bencomo2023implicit} emulate state-space modelling via gradient-based optimisation.
% These methods are particularly useful when computational efficiency is prioritised over analytical tractability

% \subsubsection{Summary of Approaches}
% In summary, we have presented several approaches for estimating the posterior over model parameters in a Gaussian form. Variational Bayes (\texttt{VB}) offers flexibility by optimizing a divergence measure without requiring a full normalization constant. Conjugate updates (\texttt{Cj}) provide analytically tractable solutions that allow for efficient, recursive parameter updates. Linear-Gaussian approximations (\texttt{LG}) simplify computations by assuming linearity but may lose accuracy in non-linear settings. Finally, alternative methods such as SMC and OGD approaches provide additional flexibility, especially in highly non-linear scenarios where traditional methods are insufficient.

\subsection{Weighting function for auxiliary variable \cWeightraw}
\label{sec:choice-weighting-function}

The  term
$\nu_{t}(\auxv_{t})$
 defines the weights over possible values of the auxiliary variable
  \cAux.
% We have $\nu(\cdot; \data_{1:t}): \vPsi_t \to [0,1]$,
% with the condition $\int_{\auxv_t \in \vPsi_t}\nu_t(\d\auxv_t) = 1$.
% For simplification, we denote $\nu(\cdot; \data_{1:t}) = \nu_t(\cdot)$.
We compute it as the marginal posterior distribution
$\nu_{t}(\auxv_{t})=p(\auxv_t\cond \data_{1:t})$ (see e.g.,  \cite{adams2007bocd, fearnhead2007line, fearnhead2011adaptivecp, li2021onlinelearning})
or with  \textit{ad-hoc} rules (see e.g., \cite{nassar2022bam, abeles2024adaptive,titsias2023kalman}).
In the former case,
the weighting function takes the form
\begin{equation}
\label{eq:weighting-marginal-posterior}
\begin{aligned}
    \nu_t(\auxv_t)
    &= p(\auxv_t \cond \data_{1:t})\\
    &= 
    p(\vy_t \cond \vx_t, \auxv_t, \data_{1:t-1})\,
    \int_{\auxv_{t-1}\in\vPsi_{t-1}}
    p(\auxv_{t-1} \cond \data_{1:t-1})\,
    p(\auxv_t \cond \auxv_{t-1}, \data_{1:t-1}) \d \auxv_{t-1},
\end{aligned}
\end{equation}
where one assumes that $\vy_t$ is conditionally independent of $\auxv_{t-1}$,
given $\auxv_t$, and $\vx_t$ is an exogenous vector.
The first term on the right hand side of
\eqref{eq:weighting-marginal-posterior}
is known as the conditional posterior predictive,
and is given by
\begin{equation}
    p(\vy_t \cond \vx_t, \auxv_t, \data_{1:t-1}) =
    \int p(\vy_t \cond \vtheta_t, \vx_t)\,\pi(\vtheta_t;\, \auxv_t, \data_{1:t-1}) \d\vtheta_t.
\end{equation}
This integral over $\vtheta_t$ %corresponds to the normalization constant for the parameter posterior,
%and 
may require approximations,
as we discussed in Section~\ref{sec:A1}.
Furthermore,
the integral over $ \auxv_{t-1}$
in \eqref{eq:weighting-marginal-posterior}
may also require approximations,
depending on the nature of the auxiliary variable $\auxv_t$, and the modelling assumptions for
$p(\auxv_t \cond \auxv_{t-1}, \data_{1:t-1})$.
We provide some examples below.

% \subsubsection{Recursive estimation of the marginal posterior (\texttt{RP})}
% In \texttt{RP} methods,

\subsubsection{Discrete auxiliary variable (DA)}
%(\texttt{DA[inf]} and \texttt{DA[K]})}

Here we assume the 
auxiliary variable takes values in a discrete space $\auxv_t \in \vPsi_t$.
The weights for the discrete auxiliary variable (\texttt{DA}) can be computed with a fixed number of hypotheses $K\geq 1$  or with a growing number of hypotheses
if the cardinality of $\vPsi_t$ increases through time;
we denote these cases by \texttt{DA[K]} and \texttt{DA[inf]} respectively.
% Then we can find the value of %find values for 
% \eqref{eq:weighting-marginal-posterior} for each $\auxv_t \in \vPsi_t$. 
%These methods produce a set of finite values.
Below, we provide three examples that estimate the weights under \texttt{DA[inf]} recursively.

\noindent \underline{\texttt{RL}}  (runlength with Markovian assumption): for $\auxv_t = r_t$, the work of \cite{adams2007bocd}
takes
\begin{equation}\label{eq:transition-rl-markov}
    p(r_t \cond r_{t-1}, \data_{1:t-1}) =
    % p(r_t \cond r_{t-1}) =
    \begin{cases}
        1 - H(r_{t-1}) & \text{if } r_t = r_{t-1} + 1,\\
        H(r_{t-1}) & \text{if } r_t = 0, \\
        0 & \text{otherwise},
    \end{cases}
\end{equation}
where $H: \mathbb{N}_0 \to (0,1)$ is the hazard function. A popular choice is to take $H(r)=\kappa\in(0,1)$ to be a fixed
constant hyperparameter known as the hazard rate.
The choice \RLPR[inf] 
%with \texttt{DA[inf]} 
is known as the Bayesian online changepoint detection model (BOCD).

\noindent \underline{\texttt{CPL}}  (changepoint location): for $\auxv_t=s_{1:t}$, the work of \cite{li2021onlinelearning} takes
\begin{equation}\label{eq:transition-cpl-independent}
    p(\tilde{s}_{1:t} \cond s_{1:t-1}, \data_{1:t-1}) = 
    \begin{cases}
        \kappa & \text{if } \left([\tilde{s}_{1:t} \setminus \tilde{s}_t] = s_{1:t-1}\right) \text{ and } \tilde{s}_t = 1,\\
        1 - \kappa & \text{if } \left([\tilde{s}_{1:t} \setminus \tilde{s}_t] = s_{1:t-1}\right) \text{ and } \tilde{s}_t = 0,\\
        0 & \text{otherwise},
    \end{cases}
\end{equation}
i.e., the sequence of changepoints at time $t$ correspond to the sequence of changepoints up to time $t-1$,
plus a newly sampled value for $t$.
See Appendix \ref{c-aux:CPL} for details
 on how to compute $\nu_t(s_{1:t})$.

\noindent \underline{\texttt{CPT}} (changepoint timestep with Markovian assumption): for $\auxv_t = \vzeta_t$, the work of \cite{fearnhead2007line}
takes
\begin{equation}
    p(\vzeta_t \cond \vzeta_{t-1}, \data_{1:t-1}) = p(\zeta_{\ell,t} \cond \zeta_{\ell,t-1}) = J(\zeta_{\ell,t} - \zeta_{\ell,t-1}),
\end{equation}
with $J:\mathbb{N}_0 \to (0,1)$ a probability mass function.
Note that $\zeta_{\ell,t} - \zeta_{\ell,t-1}$ is the distance between two changepoints, i.e., a runlength.
In this sense, $\zeta_{\ell,t} - \zeta_{\ell,t-1} = r_t$,
which relates \texttt{CPT} to \texttt{RL}.
See their paper for details
on how to compute $\nu_t(\vzeta_{t})$.

\paragraph{Low-memory variants --- from \texttt{DA[inf]} to \texttt{DA[K]}} 
In the examples above, the number of computations to obtain 
$\sum_{\auxv_{t}} \nu_{t}(\auxv_{t})$ grows in time.
To fix the computational cost, one can restrict the sum
to be over a subset ${\cal A}_t$ of the space of $\auxv_t$ with cardinality $|{\cal A}_t| = K \geq 1$.
Each element in the set ${\cal A}_t$ is called a hypothesis and given $K\geq 1$, we 
keep the $K$ most likely elements
---according to $\nu_{t}(\auxv_{t})$--- in ${\cal A}_t$.
We then define the normalised weighting function
\begin{equation}
    \hat{\nu}_{t}(\auxv_{t}) = 
    \frac{\nu_{t}(\auxv_{t})}{\sum_{\auxv'_{t} \in {\cal A}_t} \nu_{t}(\auxv'_{t})},
\end{equation}
which we use instead of $\nu_{t}(\auxv_{t})$.
For example, in \texttt{RL} above, ${\cal A}_{t-1} = \{r_{t-1}^{(k)} : k = 1, \ldots, K\}$
are the unique $K$ most likely runlengths where the superscript represents the ranking according to $\nu_{t-1}(\cdot)$.
Then, at time $t$,
the augmented set $\bar{\cal A}_t$ becomes $({\cal A}_{t-1}+1)\cup \{0\}$, where the sum is element-wise,
and we then compute the $K$ most likely elements of $\bar{\cal A}_t$ to define ${\cal A}_t$.
In \texttt{CPL}, ${\cal A}_{t-1} $
%$= \{s_{1:t-1}^{(k)}:k=1,\ldots,K\}$
contains the $K$ most likely sequences of changepoints,    $\bar{\cal A}_{t} $ is defined as the collection of the $2\,K$ sequences where each sequence of ${\cal A}_{t-1}$ has a zero or one concatenated at the end.
%$= \{(s_{1:t-1}^{(k)},0):k=1,\ldots,K\} \cup \{(s_{1:t-1}^{(k)},1):k=1,\ldots,K\}$,
Finally, the $K$ most likely elements in $\bar{\cal A}_{t}$ define ${\cal A}_{t}$.
This style of pruning is common in segmentation methods; see, e.g., \cite{saatcci2010GPBOCD},
but other pruning are also possible, such as those proposed by \cite{li2021onlinelearning},
or sampling-based approaches; see e.g. \cite{Doucet2000}.

\paragraph{Other choices for \texttt{DA[K]}}
Finally, some choices of weighting functions are derived using ad-hoc rules,
meaning that explicit or approximate solutions to the Bayesian posterior are not needed.
One of the most popular choices of ad-hoc weighting functions are mixture of experts,
which weight different models according to a given criterion.

\noindent \underline{\texttt{ME}}  (mixture of experts with algorithmic weighting):
Consider $\auxv_t = \valpha_t$.
Let $\valpha_{t,k}=k$ denote the $k$-th \textit{configuration} over \cPrior.
Next, denote by $\vw_t = \{\vw_{t,1}, \ldots, \vw_{t,K}\}$ a set of weights,
where $\vw_{t,k}$ corresponds to the weight for the $k$-th expert at time $t$.
The work of \cite{chaer1997mixturekf}
considers the weighting function
\begin{equation}
    \nu_t(\vw_{t})_k = \frac{\exp(\vw_{t,k}^\intercal\,\vy_t)}{\sum_{j=1}^K \exp(\vw_{t,j}^\intercal\,\vy_t)},
\end{equation}
for $k=1,\ldots,K$.
The set of weights $\vw_t$ are determined by maximising the surrogate gain function
\begin{equation}
    {\cal G}_t(\vw_t)
    = p(\vy_t \cond \vx_t, \data_{1:t-1})
    = \sum_{k=1}^K p(\vy_t \cond \vx_t, \valpha_{t,k}, \data_{1:t-1})\,\nu_t(\vw_{t})_k,
\end{equation}
with respect to $\vw_{t,k}$ for all $k = 1, \ldots, K$ at every timestep $t$.

We write \texttt{DA[K]}, where \texttt{K} is the number hypothesis, for methods that use \texttt{K} hypotheses at most. 
On the other hand, we write \texttt{DA[inf]} when we do not 
impose a bound on the number of hypotheses used.
Note that even when the choice of \cWeight is built using \texttt{DA[inf]},
one can modify it to make it \texttt{DA[K]}.

\paragraph{Discrete auxiliary variable with greedy hypothesis selection (\texttt{DA[1]})}
A special case of the above is \texttt{DA[1]}, where we employ a single hypothesis.
In these scenarios, 
we set $\nu(\auxv_t) = 1$ where $\auxv_t$ is the most likely hypothesis.
% The main example in this category is below.
%a value of \cWeight is built weighting possible values in $\vPsi_t$.
%We then choose a single hypothesis and set the value $\nu(\auxv_t) = 1$.

\noindent \underline{\texttt{RL}} (Greedy runlength):
For $\auxv_t = r_t$ and $\texttt{DA[1]}$,
we take
\begin{equation}\label{eq:transition-rl-spr}
    p(r_t \cond r_{t-1}, \data_{1:t-1}) =
    % p(r_t \cond r_{t-1}) =
    \begin{cases}
        1 - \kappa & \text{if } r_t = r_{t-1} + 1,\\
        \kappa & \text{if } r_t = 0, \\
        0 & \text{otherwise}.
    \end{cases}
\end{equation}
Our choice of \cWeight 
is based on the marginal predictive likelihood ratio,
which is derived from the computation of $p(r_t \cond \data_{1:t})$
under either either
an increase in the runlength ($r_t^{(1)} = r_{t-1} + 1)$
or a reset of the runlength ($r_t^{(0)} = 0$).
Under these assumptions, the form of $\nu_t(r_t^{(1)})$
is
\begin{equation}\label{eq:rl-oupr-weight}
    \nu_t(r_t^{(1)}) = \frac{p(\vy_t \cond r_t^{(1)}, \vx_t, \data_{1:t-1})\,(1 - \kappa)}{p(\vy_t \cond r_t^{(0)}, \vx_t, \data_{1:t-1})\,\kappa + p(\vy_t \cond r_t^{(1)}, \vx_t, \data_{1:t-1})\,(1-\kappa)}.
\end{equation}
For details on the computation of \cWeight, see Appendix \ref{sec:rl-spr-implementation}.
For a detailed implementation of
\cAux \texttt{RL},
\cPrior \texttt{OUPR},
\cWeight \texttt{DA[1]}, and
\cPosterior \texttt{LG}, 
see Algorithm \ref{algo:rl-spr-step} in the Appendix.

For example, \texttt{RL[1]} is a runlength $r_t$ with a single hypothesis.
We provide another example next.

\noindent \underline{\texttt{CPL}} (changepoint location with retrospective membership):
for $\auxv_t=s_{1:t}$, the work of \cite{nassar2022bam}
evaluates the probability of past datapoints belonging in the current regime.
In this scenario,
\begin{equation}
    p(s_{1:t} \cond s_{1:t-1}, \data_{1:t-1}) = p(s_{1:t} \cond \data_{1:t-1}),
\end{equation}
so that
\begin{equation}
    p(s_{1:t} \cond \data_{1:t}) \propto
    p(s_{1:t} \cond \data_{1:t-1})\,p(\vy_t \cond \vx_t, s_{1:t}, \data_{1:t-1}).
\end{equation}

This method allows for exact computation by summing over all possible $2^t$ elements.
However, to reduce the computational cost, they propose a discrete optimisation over possible values
$\{\nu_t(s_{1:t})\,:\, s_{1:t} \in \{0,1\}^t\}$, where
$\nu_t(s_{1:t}) = p(s_{1:t} \cond \data_{1:t})$.
Then, the hypothesis with highest probability is stored and gets assigned a weight of one.

%\subsubsection{Approximations to the marginal posterior (\texttt{AP})}
\subsubsection{Continuous auxiliary variable (\texttt{CA})}
\label{sec:CA}

Here, we briefly discuss continuous auxiliary variables \texttt{(CA)}.
For some choices of $\auxv_t$ and transition densities $p(\auxv_t \cond \auxv_{t-1}, \data_{1:t-1})$,
computation of \eqref{eq:weighting-marginal-posterior} becomes infeasible.
In these scenarios, 
%$\gamma_t$ is taken as a point-wise approximation of \eqref{eq:weighting-marginal-posterior}.
we use simpler approximations.
We give an example below.
%We provide two examples below:

\noindent \underline{\texttt{CPP}} (Changepoint probability with empirical Bayes estimate):
for $\auxv_t=\upsilon_t$, consider
\begin{equation}
    p(\upsilon_t \cond \upsilon_{t-1}, \data_{1:t-1}) = p(\upsilon_t),
\end{equation}
so that
\begin{equation}
    p(\upsilon_t \cond \data_{1:t}) \propto
    p(\upsilon_t)\,p(\vy_t \cond \vx_t, \upsilon_t).
\end{equation}
% Because $p(\upsilon_t \cond \data_{1:t})$ does not have closed-form solution,
The work of \cite{titsias2023kalman} 
takes $\nu_t(\upsilon_t)=\delta(\upsilon_t - \upsilon_t^*)$, where $\delta$ is the Dirac delta function
and $\upsilon_t^*$ is a point estimate centred at
the maximum of the marginal
posterior predictive likelihood:
\begin{equation}
    \upsilon_t^* = \argmax_{\upsilon \in [0,1]} p(\vy_t \cond \vx_t, \upsilon, \data_{1:t-1}).
    \label{eqn:nut}
\end{equation}
In practice, \eqref{eqn:nut} is approximated by taking gradient steps towards the minimum.
This is a form of empirical Bayes approximation,
since we compute the most likely value of the prior 
after marginalizing out $\vtheta_t$.
The work of \cite{Galashov2024} considers a modified configuration with choice of \cAux
${\bm \upsilon}_t \in (0,1)^\dimstate$.

\section{Unified view of examples in the literature}
\label{sec:literature}

Table \ref{tab:related-bone-methods} shows that many existing methods can be written as instances
of BONE.
Rather than specifying the choice of %modelling function for 
(M.1), we instead
write the task for which it was designed,
as discussed in Section \ref{sec:tasks}.
We will experimentally compare a subset of these methods in Section \ref{section:experiments}.

The methods presented in Table \ref{tab:related-bone-methods} can be directly applied
to tackle any of the problems mentioned in Section \ref{sec:tasks}.
However, as choice of \cModel, we specify the task under which the configuration was introduced.\footnote{
In general, the components of BONE can be thought as 
the building blocks for new methods. Some of these combinations 
would not be useful, but they can be employed nonetheless.
}
\begin{table}[htb]
    \centering
    \footnotesize
    \renewcommand{\arraystretch}{1.2} % Improve spacing
    \begin{tabular}{l|c|c|c|c|c|c}
         % reference & \cModel  $\vtheta_t$ & \cAux  $\auxv_t$ & \cPrior  $\pi_t$  & \cPosterior $q_t$  & \cWeight  $\nu_t$\\
         \hline
         \textbf{Reference}  & \textbf{Task} & \textbf{\cAuxname} & \textbf{\cPriorname} & \textbf{\cPosteriorname} & \textbf{\cWeightname} \\
         \hline
         %\hline
         \cite{kalman1960filter} &  filtering & \texttt{C} & \texttt{LSSM} & \texttt{LG} & \texttt{DA[1]} \\
         \cite{magill1965optimaladaptivefilter} &  filtering & \texttt{ME} & \texttt{LSSM} & \texttt{LG} & \texttt{DA[K]}\\
         \cite{chang1978switchingkf} &  filtering & \texttt{ME} & \texttt{LSSM} & \texttt{LG} & \texttt{CA} \\
         \cite{chaer1997mixturekf} &  filtering & \texttt{ME} & \texttt{LSSM} & \texttt{LG} & \texttt{DA[K]}\\
         \cite{ghahramani2000vsssm} &  SSSM & \texttt{ME} & \texttt{Static} & \texttt{VB} & \texttt{CA} \\
         \cite{adams2007bocd} &  seg. & \texttt{RL} & \texttt{PR} & \texttt{Cj} & \texttt{DA[inf]}\\
         \cite{fearnhead2007line} &  seg. \& preq. & \texttt{CPT}/\texttt{ME} & \texttt{PR} & \texttt{Any} & \texttt{DA[inf]}\\
         \cite{wilson2010-bocd-hazard-rate} &  seg. & \texttt{RLCC} & \texttt{PR} & \texttt{Cj} & \texttt{DA[inf]}\\
         \cite{fearnhead2011adaptivecp} &  seg. & \texttt{CPT}/\texttt{ME} & \texttt{MMPR} & \texttt{Any} & \texttt{DA[inf]}\\
         \cite{mellor2013changepointthompsonsampling} &  bandits & \texttt{RL}  & \texttt{PR} & \texttt{Cj} & \texttt{DA[inf]}\\ % read in detail!
         \cite{nguyen2017vcl} &  OCL & \texttt{CPL} & \texttt{Sub} & \texttt{VB} & \texttt{DA[1]}\\
         \cite{knoblauch2018varbocd} &  seg. & \texttt{RL}/\texttt{ME} & \texttt{PR} & \texttt{Cj} & \texttt{DA[inf]}\\
         \cite{kurle2019continual} &  OCL & \texttt{CPV} & \texttt{Sub} & \texttt{VB} & \texttt{DA[1]} \\
         \cite{li2021onlinelearning} &  OCL & \texttt{CPL} & \texttt{MCI} & \texttt{VB} & \texttt{DA[inf]} \\
         \cite{nassar2022bam} &  bandits \& OCL & \texttt{CPV} & \texttt{Sub} & \texttt{LG} & \texttt{DA[1]} \\
         \cite{liu2023bdemm} &  preq. & \texttt{ME} & \texttt{C},\texttt{LSSM} & \texttt{Any} & \texttt{DA[K]} \\
         \cite{chang2023lofi} & OCL & \texttt{C} & \texttt{ACI} & \texttt{LG} & \texttt{DA[1]} \\
         \cite{titsias2023kalman} &  OCL & \texttt{CPP} & \texttt{OU} & \texttt{LG} & \texttt{CA} \\
         \cite{Galashov2024} & CL & \texttt{CPP} & \texttt{OU} & \texttt{VB} & \texttt{CA} \\
         \cite{abeles2024adaptive} &  preq. & \texttt{ME} & \texttt{LSSM} & \texttt{LG} & \texttt{DA[K]}\\
         \RLSPR (ours) & any & \texttt{RL} & \texttt{SPR} & \texttt{Any} & \texttt{DA[1]}
         % \cite{dohare2024continualbackprop} & OCL & & & & \texttt{C}\\
    \end{tabular}
    \caption{
    List of methods ordered by publication date.
    The tasks 
    %in column (M.1) 
    are discussed in Section \ref{sec:tasks}.
    We use the following abbreviations:
    SSSM means switching state space model;
    (O)CL means (online) continual learning;
    seg.~means segmentation;
    % rein.l~means reinforcement learning;
    preq.~means prequential.
    % GD.~means gradient descent.
    %Methods with \texttt{*} denote that they were introduced without any particular configuration in mind.
    Methods that consider two choices of \cAux are denoted by `\texttt{X}/\texttt{Y}'.
    This corresponds to a double expectation in \eqref{eq:bone-prediction}---one for each choice of auxiliary variable.
    }
    \label{tab:related-bone-methods}
\end{table}

%% file: diagrams/bone-ssm.tex
    \begin{tikzpicture}[
        node distance=0.5cm and 1.0cm,
        every node/.style={draw, circle, minimum size=1.2cm,
        text width=0.5cm, align=center},
        every path/.style={thick, ->, >=stealth}
    ]
        	
        % Nodes for hidden states (theta)
        \node (theta1) {$ \vtheta_{t-1} $};
        \node[right=of theta1] (theta2) {$ \vtheta_t $};
        \node[right=of theta2] (theta3) {$ \vtheta_{t+1} $};

        % Nodes for auxiliary variable
        \node[above=of theta1] (auxv1) {$\auxv_{t-1}$};
        \node[above=of theta2] (auxv2) {$\auxv_t$};
        \node[above=of theta3] (auxv3) {$\auxv_{t+1}$};

        % Nodes for observed measurements (\vy_values)
        \node[fill=lightgray, below=of theta1] (y1) {$ \vy_{t-1} $};
        \node[fill=lightgray, below=of theta2] (y2) {$ \vy_t $};
        \node[fill=lightgray, below=of theta3] (y3) {$ \vy_{t+1} $};

        % Nodes for latent states (x)
        \node[rectangle, fill=lightgray, below=of y1, xshift=-3mm] (x1) {$ \vx_{t-1} $};
        \node[rectangle, fill=lightgray, below=of y2, xshift=-3mm] (x2) {$ \vx_t $};
        \node[rectangle, fill=lightgray, below=of y3, xshift=-3mm] (x3) {$ \vx_{t+1} $};
                
        % Arrows for hidden state transitions
        \path (theta1) edge (theta2);
        \path (theta2) edge (theta3);
%        \path (theta3) edge (emptytheta);
        
        % Arrows for observations
        \path (theta1) edge (y1);
        \path (theta2) edge (y2);
        \path (theta3) edge (y3);
        
        % Arrows for observations
        \path (x1) edge (y1);
        \path (x2) edge (y2);
        \path (x3) edge (y3);
        
        % Arrows for auxiliary variables
        \path (auxv1) edge (theta1);
        \path (auxv2) edge (theta2);
        \path (auxv3) edge (theta3);

        % Arrows for auxiliary transitioin
        \path (auxv1) edge (auxv2);
        \path (auxv2) edge (auxv3);
%        \path (auxv3) edge (emptyauxv);
        
        % Arrows for auxiliary transitioin
        \path (auxv1) edge[bend right=40, dotted] (y1);
        \path (auxv2) edge[bend right=40, dotted] (y2);
        \path (auxv3) edge[bend right=40, dotted] (y3);
    \end{tikzpicture}

%% file: sections/experiments.tex
\section{Experiments}
\label{section:experiments}

In this section we experimentally evaluate 
different algorithms 
within the BONE framework
on a number of tasks.
 
Each experiment consists of a \textit{warmup} period where the hyperparameters are chosen,
and a \textit{deploy} period where sequential predictions and updates are performed.
In each experiment, we fix the choice of measurement model $h$ \cModel and posterior inference method \cPosterior,
and then compare different methods with respect to their choice of \cAux, \cPrior, and \cWeight. 
For \texttt{DA} methods, we append the number of hypotheses in brackets to determine
how many hypotheses are being considered.
For example,
\RLPR[1] denotes one hypothesis,
\RLPR[K] denotes $K$ hypotheses,
and \RLPR[inf] denotes all possible hypotheses.
In all experiments, unless otherwise stated, we consider a single hypothesis for choices of \texttt{DA}.
See Table \ref{tab:rosetta-methods} for the methods we compare.

\begin{table}[htb]
    \centering
    \footnotesize
    \begin{tabular}{c|c|c|p{8cm}|p{1.5cm}}
        \textbf{M.2-M.3} & \textbf{Eq.} & \textbf{A.2} & \textbf{Description} & \textbf{Sections}\\
         \hline
         \hline
        \multicolumn{4}{c}{static}\\
         \hline
         \namemethod{C-Static} & \eqref{eq:cprior-c-f} & - &   
         {\scriptsize
         This corresponds to the static case with a classical Bayesian update.
         This method does not assume changes in the environment.
         }
         & {\scriptsize
         \ref{experiment:KPM}, \ref{experiment:heavy-tail-regression}
         }
         \\
         \hline
        \multicolumn{4}{c}{abrupt changes}\\
         \hline
         \texttt{RL-PR} & \eqref{eq:cprior-rl-pr} & \texttt{DA[inf]} & 
         {\scriptsize
         This approach, commonly referred to as Bayesian online changepoint detection (\textbf{BOCD}),
         assumes that non-stationarity arises from independent  blocks of time, each with stationary data. Estimates are made using data from the current block.
         See Appendix \ref{sec:rl-pr-implementation} for more details.
         }
         & 
         {\scriptsize
         \ref{experiment:hour-ahead-forecasting}, \ref{exp:logistic-reg}, \ref{experiment:bandits}, \ref{experiment:KPM}, \ref{experiment:heavy-tail-regression}, 
         }
         \\
          \newmethod{WoLF+RL-PR*} & \eqref{eq:cprior-rl-pr} & \texttt{DA[inf]} &
         {\scriptsize
         Special case of \namemethod{RL-PR} with explicit choice of \cModel which makes it robust to outliers.
         }
         & {\scriptsize \ref{experiment:heavy-tail-regression} }
         \\
         \hline
        \multicolumn{4}{c}{gradual changes}\\
         \hline
         \CPPD & \eqref{eq:cprior-cpp-d} & \texttt{CA}  & 
         {\scriptsize
         Updates are done using a discounted mean and covariance according to the probability estimate that a change has occurred.
         }
         & {\scriptsize
         \ref{experiment:hour-ahead-forecasting}, \ref{exp:logistic-reg}, \ref{experiment:bandits}
         }
         \\
         \namemethod{C-ACI} & \eqref{eq:cprior-c-aci} &  - &  
         {\scriptsize
         At each timestep,
         this method assumes that the parameters
         evolve according to a linear map $\vF_t$,
         at a rate given by a known positive semidefinite covariance matrix $\vQ_t$.
         } 
         & 
         {\scriptsize
        \ref{experiment:hour-ahead-forecasting}, \ref{exp:logistic-reg}, \ref{experiment:bandits}, 
         }
         \\
        \hline
        \multicolumn{4}{c}{abrupt \& gradual changes}\\
         \hline
          \namemethod{RL-MMPR} & \eqref{eq:rl-mmpr-prior-reset} & \texttt{DA[inf]} & 
         {\scriptsize 
            Modification of \namemethod{CPT-MMPR} that assumes dependence between any two consecutive blocks of time
            and with choice of \texttt{RL}.
            This combination employs a moment-matching approach when evaluating the prior mean and covariance under a changepoint.
            See Appendix \ref{sec:rl-mmpr-implementation} for more details.
         }
         &
         {\scriptsize 
        \ref{experiment:KPM}
         }
         \\
         \texttt{RL-OUPR} & \eqref{eq:SPR-equation-gt}& \texttt{DA[1]} & 
         {\scriptsize
        Depending on the threshold parameter, updates involve either (i) a convex combination of the prior belief with the previous mean and covariance based on the estimated probability of a change (given the run length), or (ii) a hard reset of the mean and covariance, reverting them to prior beliefs.
        See Appendix \ref{sec:rl-spr-implementation} for more details.
        } &
        {\scriptsize 
        \ref{experiment:hour-ahead-forecasting}, \ref{exp:logistic-reg}, 
        \ref{experiment:bandits}, 
        \ref{experiment:KPM}
        }
    \end{tabular}
    \caption{
    List of methods we compare in our experiments.
    The first column, \textbf{M.2--M.3}, is defined by the choices of \cAux and \cPrior.
    The second column, \textbf{Eq.}, references the equation that define M.2--M.3.
    The third column, \textbf{A.2}, determines the  choice of \cWeight.
    The fourth column, \textbf{Description}, provides a brief summary of the method.
    The fifth column, \textbf{Sections}, shows the sections where the method is evaluated.
    The choice of \cModel and \cPosterior are defined on a per-experiment basis. 
    (The only exception being \newmethod{WolF+RL-PR}).
    For \cAux the acronyms are as follows: \texttt{RL} means runlength, \texttt{CPP} means changepoint probability,
    \texttt{C} means constant, and \texttt{CPT} means changepoint timestep.
    For \cPrior the acronyms are as follows:
    \texttt{PR} means prior reset,
    \texttt{OU} means Ornstein–Uhlenbeck,
    \texttt{LSSM} means linear state-space model,
    \texttt{Static} means full Bayesian update,
    \texttt{MMR} means moment-matched prior reset,
    and \texttt{OUPR} means Ornstein–Uhlenbeck and prior reset.
    We use the convention in \cite{huvskova1999gradualabruptchange} for the terminology abrupt/gradual changes.
    % For \cWeight the acronyms are as follows: \texttt{RP} means recursive estimation of the marginal posterior, \texttt{AP} means approximation of the marginal posterior, and \texttt{AW} means algorithmic weighting function.
    }
    \label{tab:rosetta-methods}
\end{table}

\subsection{Prequential prediction}
\label{experiment:prequential}

In this section, we give several examples of non-stationary prequential prediction problems.

\subsubsection{Online regression for hour-ahead electricity forecasting}
\label{experiment:hour-ahead-forecasting}

In this experiment,  we consider the task of predicting the hour-ahead electricity load
before and after the Covid pandemic.
We use the dataset presented in \cite{farrokhabadi2020electricitycovid}, which has 31,912 observations;
each observation
contains 7 features $\vx_t$ and a single target variable $\vy_t$.
The 7 features correspond to
pressure (kPa), cloud cover (\%), humidity (\%), temperature (C) , wind direction (deg), and wind speed (KmH).
The target variable is the hour-ahead electricity load (kW).
To preprocess the data,
we normalise the target variable $\vy_t$ by subtracting an exponentially weighted moving average (EWMA) mean with a half-life of 20 hours,
then dividing the resulting series by an EWMA standard deviation with the same half-life.
To normalise the features $\vx_t$, we divide each by a 20-hour half-life EWMA.
The features are lagged by one hour.

Our choice of measurement model $h$  is a two-hidden layer multilayered perceptron (MLP)
with four units per layer and a ReLU activation function.
% For the auxiliary weighting term (A.2),
% we choose the \texttt{AW} method.

For this experiment, we consider
\RLSPR  (our proposed method),
\RLPR (a classical method),
\CACI (a simple benchmark),
and \CPPD (a modern method).
For computational convenience, we plug in a point-estimate (MAP estimate)
of the neural network parameters when making predictions using $h$.
More precisely, given $\auxv_t$, we use $h(\vtheta_t^*, \vx_{t+1})$ to make a (conditional) prediction,
where $\vtheta_t^* = \argmax_\vtheta q(\vtheta;\, \auxv_t, \data_{1:t})$.
For  a fully Bayesian treatment of neural network predictions, see \cite{immer2021improving};
we leave the implementation of these approaches for future work.

The hyperparameters of each method are found using the first 300 observations (around 13 days)
and deployed on the remainder of the dataset.
Specifically, during the warmup period we tune the value of the probability of a changepoint for \RLSPR and \RLPR. For \CACI we tune  $\vQ_t$, and  for \CPPD we tune the learning rate. See the open-source notebooks for more details.

\begin{figure}[htb]
    \centering
    \includegraphics[width=0.80\linewidth]{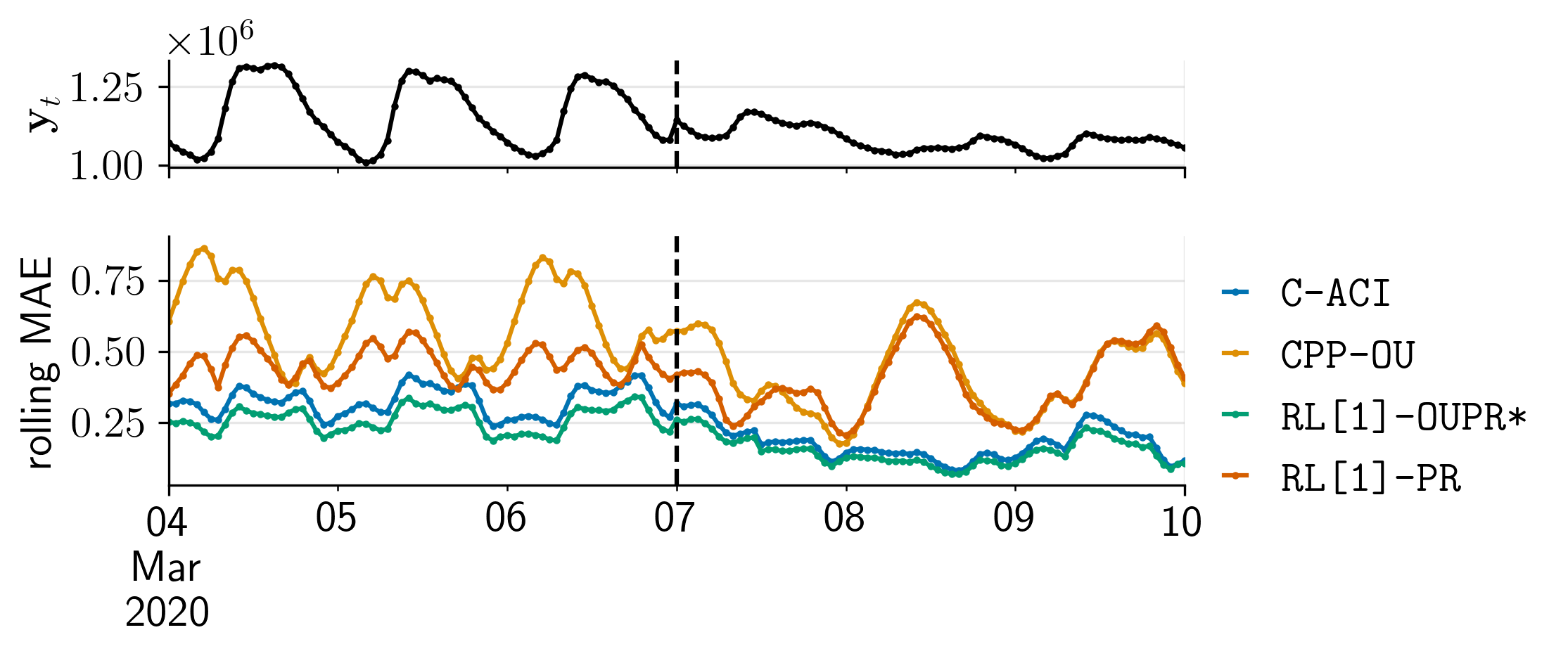}
    \caption{
    The \textbf{top panel} shows the target variable (electricity consumption) 
    from March 1 2020 to March 12 2020.
    The \textbf{bottom panel} shows the twelve-hour rolling relative absolute error of predictions
        for the same time window.
    The dotted black line corresponds to March 7 2020, when Covid lockdown began.
    %The shaded areas represent abrupt changes (pink) and gradual changes (yellow).
    }
    \label{fig:day-ahead-plot}
\end{figure}

In the top panel of Figure \ref{fig:day-ahead-plot} 
we show the evolution of the target variable $\vy_t$ between March 3 2020 and March 10 2020.
The bottom panel of Figure \ref{fig:day-ahead-plot}
shows the 12-hour rolling mean absolute error (MAE) of predictions made by the methods.
We see that there is a changepoint around March 7 2020 as pointed out in \cite{farrokhabadi2020electricitycovid}. 
This is likely due to the introduction of Covid lockdown rules.
Among the methods considered, \CACI and \RLSPR adapt the quickest after the changepoint and maintain
a low rolling MAE compared to \RLPR and \CPPD.

\begin{figure}[htb]
    \centering
    \includegraphics[width=0.80\linewidth]{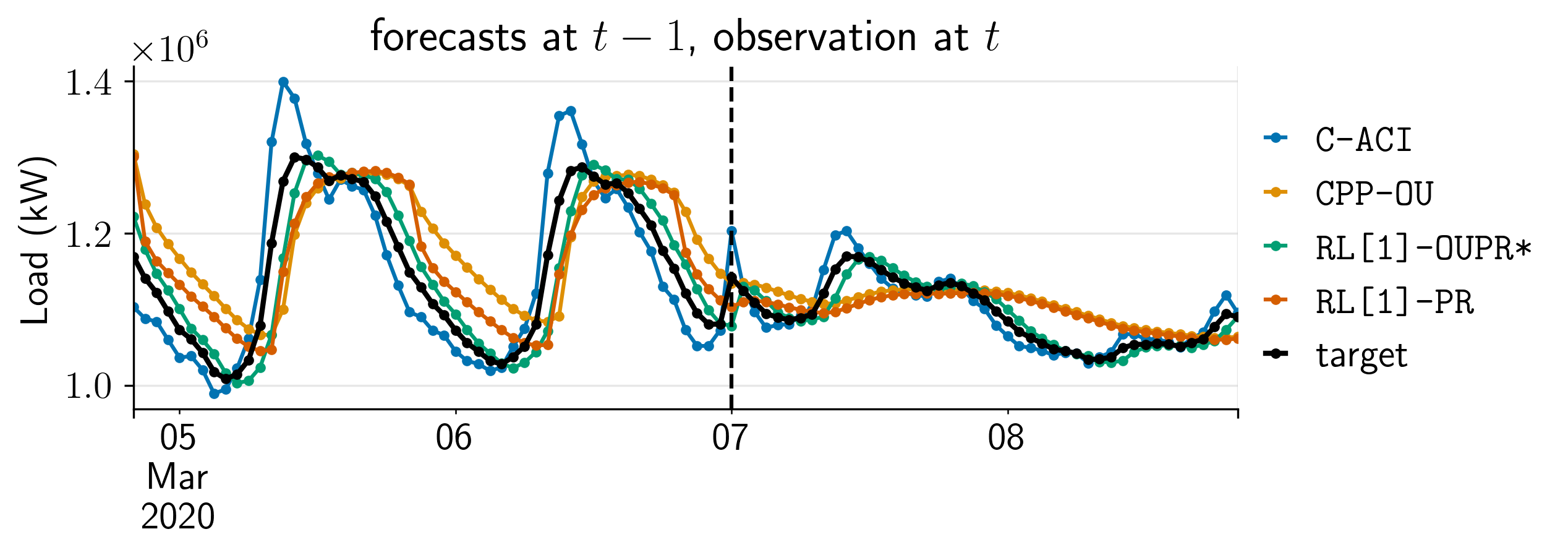}
    \caption{
    One day ahead electricity forecasting results
    for Figure \ref{fig:day-ahead-plot}.
    The dotted black line corresponds to  March 7 2020.
    }
    \label{fig:day-ahead-results-zoomed}
\end{figure}

%\paragraph{Results}
Next, Figure \ref{fig:day-ahead-results-zoomed},
shows the forecasts made by each method between March 4 2020 and March March 8 2020.
We observe a clear cyclical pattern before March 7 2020 but less so afterwards,
indicating a change in daily electricity usage from diurnal to constant.

\begin{figure}[htb]
    \centering
    \includegraphics[width=0.65\linewidth]{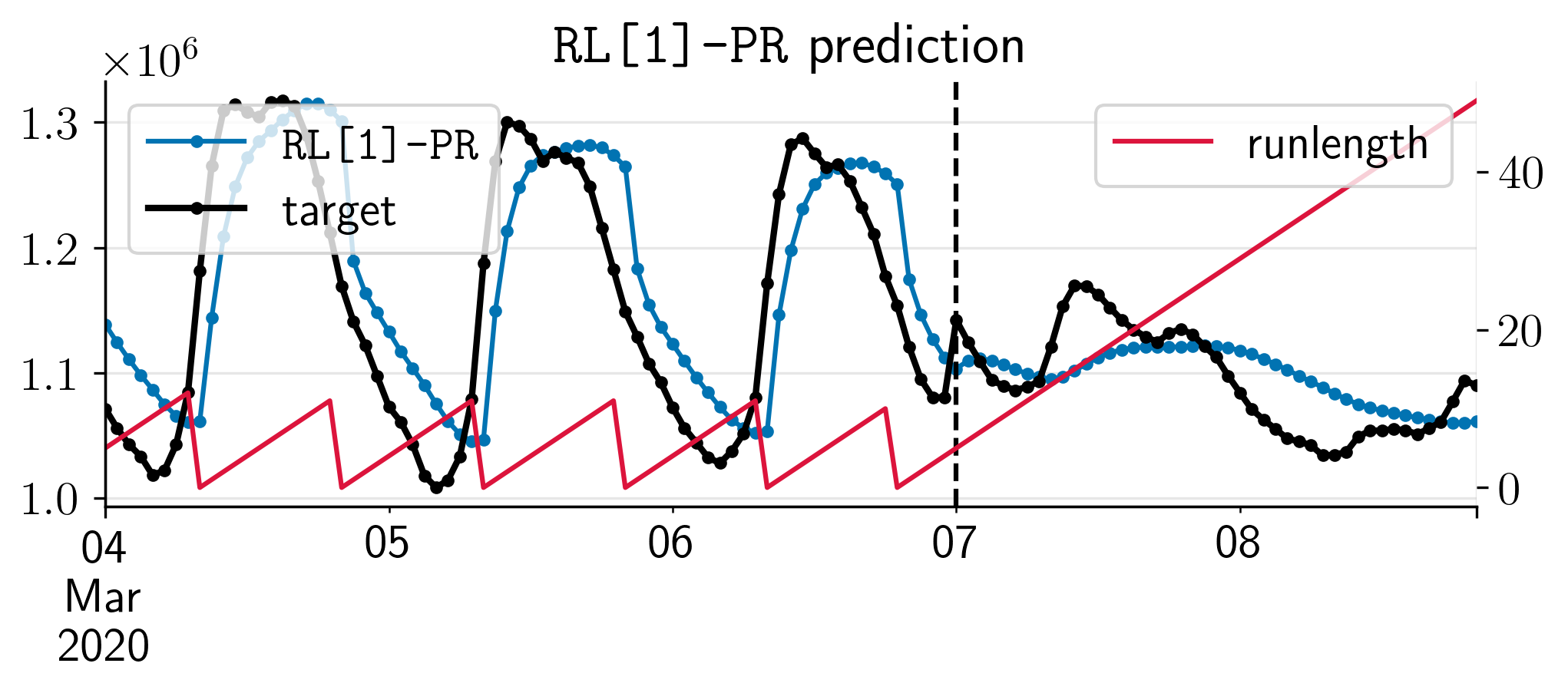}
    \caption{
    One day ahead electricity forecasting results for \RLPR together with the target variable
    on the left y-axis, and the value for runlength (\texttt{RL}) on the right y-axis.
    We see that after the 7 March changepoint, the runlength monotonically increases,
    indicating a stationary regime.
    }
    \label{fig:day-ahead-rlpr-predictions}
\end{figure}

We also observe that \RLPR and \CPPD slow-down their rate of adaptation.
One possible explanation of this behaviour is that the changes are
not abrupt enough to be captured by the algorithms.
To provide evidence for this hypothesis, Figure \ref{fig:day-ahead-rlpr-predictions}
shows, on the left $y$-axis, the predictions for \RLPR and the target variable $\vy_t$.
On the right $y$-axis, we show the estimated runlength.

We see that \RLPR resets approximately twice every day until the time of the changepoint.
After that, there is no evidence of a changepoint (as provided by the hyperparameters and the modelling choices),
so \RLPR does not reset which translates to less adaptation for the period to the right of the changepoint.

\begin{figure}[htb]
    \centering
    \includegraphics[width=0.65\linewidth]{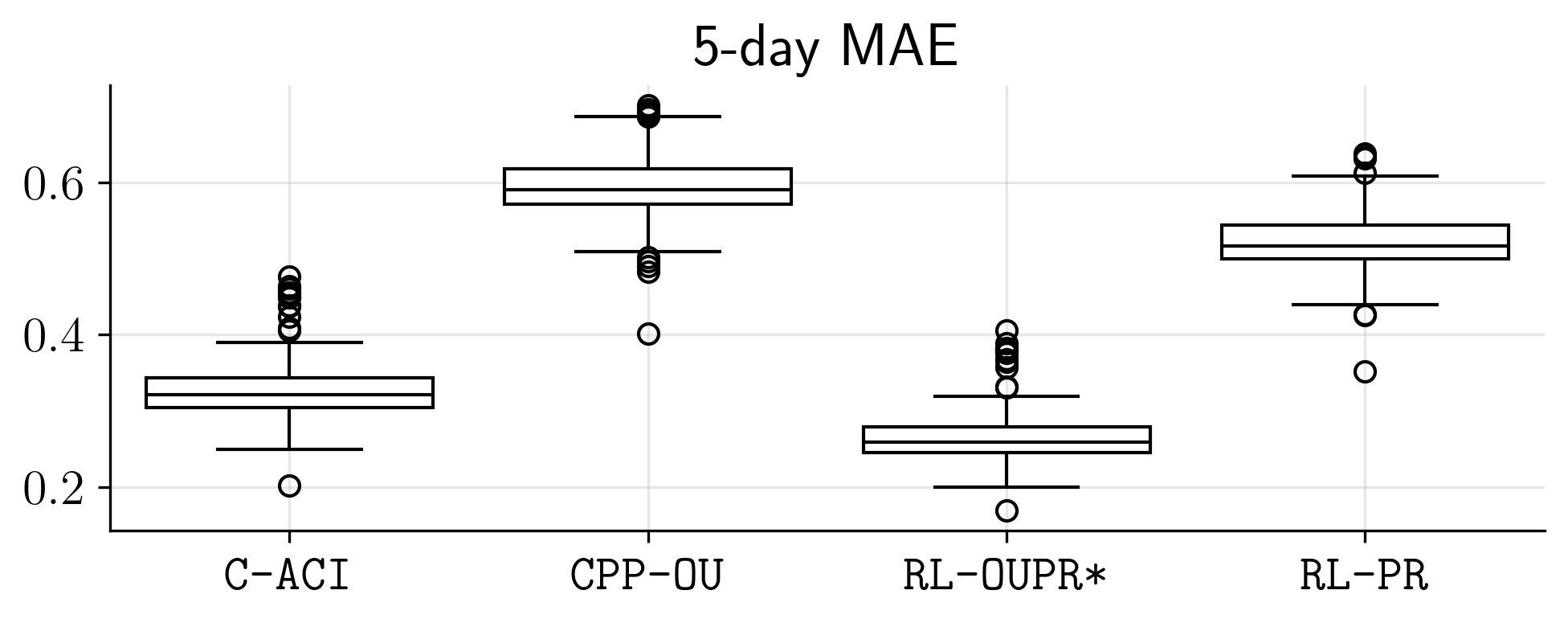}
    \caption{
    Distribution of the 5-day mean absolute error (MAE) for each of the competing methods on electricity forecasting over the entire period. For this calculation we split the  dataset into consecutive buckets containing  five days of data each, and for a given bucket we compute  the average absolute error of the predictions and observations that fall within the bucket.
    }
    \label{fig:day-ahead-results}
\end{figure}

Finally, we compare the error of predictions made by the competing methods.
This is quantified in 
Figure \ref{fig:day-ahead-results},
which shows a box-plot of the five-day MAE for each of the competing methods over the whole dataset,
from March 2017 to November 2020.
Our new \RLSPR method has the lowest MAE.

\subsubsection{Online classification with periodic drift}
\label{exp:logistic-reg}
\label{sec:periodic-drifts}

In this section we study the performance of \CACI, \CPPD, 
\RLPR, and \RLSPR for the classification experiment of Section 6.2 in \cite{kurle2019continual}.
More precisely, in this experiment $x_{t,i}\sim \mathrm{Unif}[-3,3]$ for $i \in \{1,2\}$,
$\vx_t = (x_{t,1}, x_{t,2}) \in\mathbb{R}^2$, 
$y_t \sim \mathrm{Bernoulli}(\sigma(\vtheta_t^\intercal\,\vx_t) )$
with $\vtheta^{(1)}_t = 10\,\sin(5^\circ\,t)$ and $\vtheta^{(2)}_t = 10\,\cos(5^\circ\,t)$.
Thus the unknown values of model parameters are slowly drifting deterministically according to sine and cosine functions.
The timesteps go from 0 to 720.

%below, in Figure \ref{fig:clf-rlpr-comparison} we carry out a robustness check on the number of hypothesis.

\begin{figure}[H]
    \centering
    \includegraphics[width=0.65\linewidth]{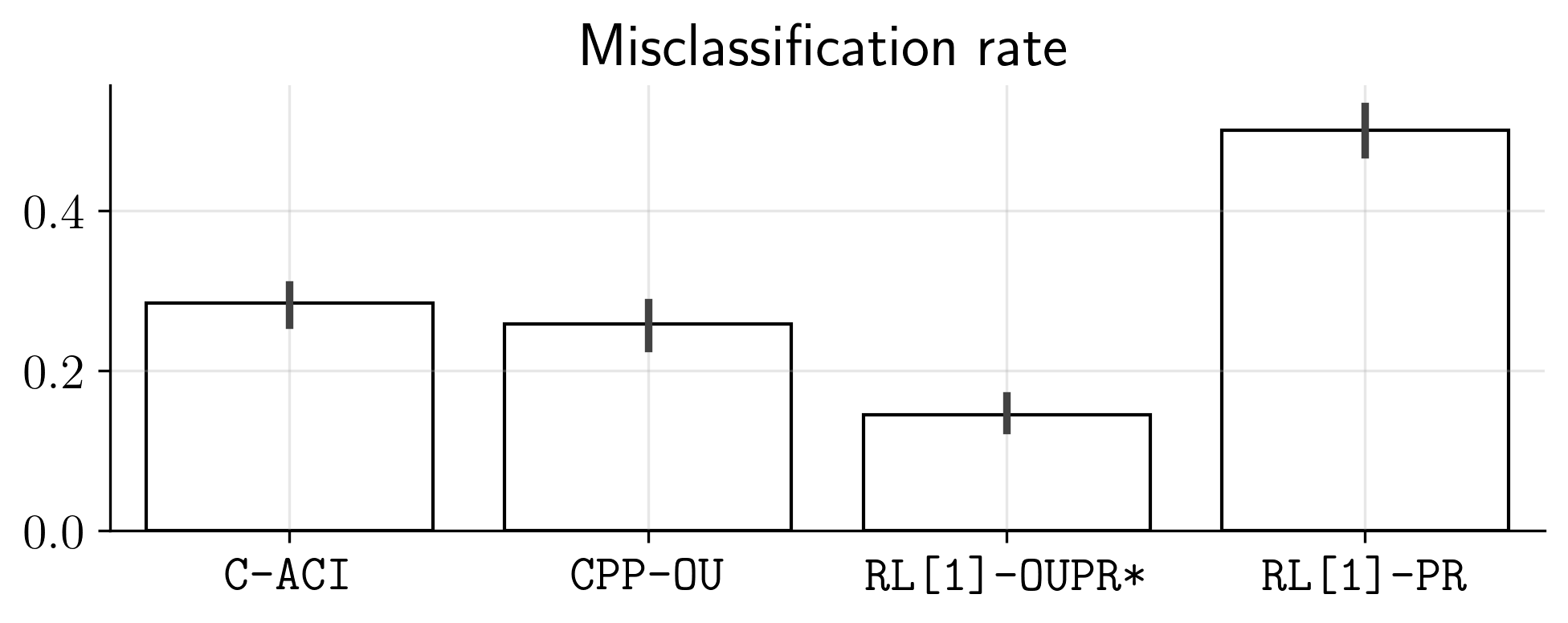}
    \caption{
    Misclassification rate of various methods on the online classification with periodic drift task.
    }
    \label{fig:clf-results}
\end{figure}

Figure \ref{fig:clf-results} summarises the results of the experiment where we show the misclassification rate (which is one minus the accuracy) for the competing methods.
% (For \RLPR we employ two hypotheses for the runlength and for \RLSPR we employ one hypothesis for the runlength.)
%
Our \RLSPR method works the best, and signifcantly outperforms \RLPR,
since we use an OU drift process with a  soft prior reset
rather than assuming constant parameter
with a hard prior rset.

\begin{figure}[H]
    \centering
    \includegraphics[width=0.60\linewidth]{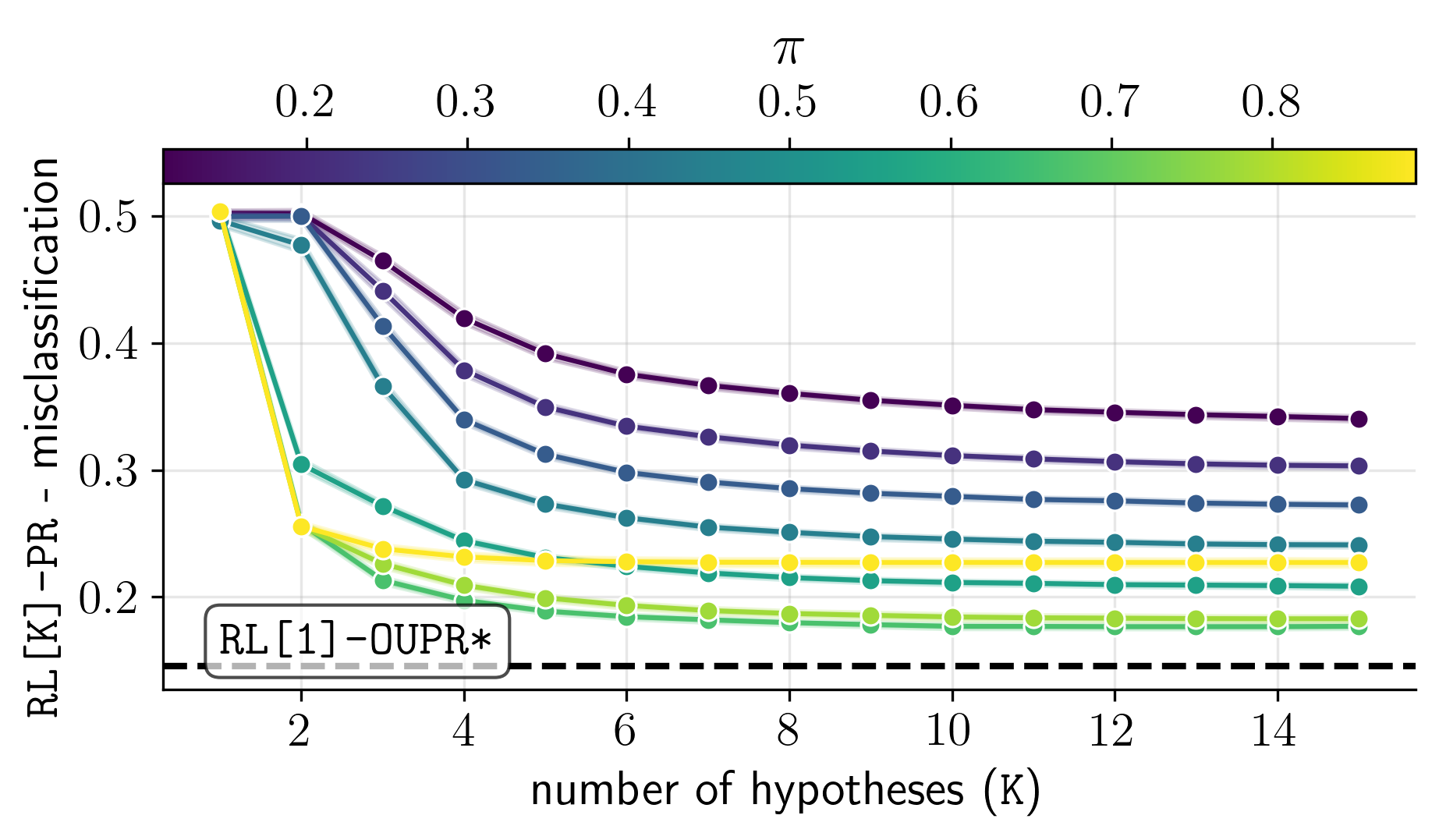}
    \caption{
    Accuracy of predictions for \RLPR as a function of the number of hypothesis and
    the prior probability of a changepoint $\kappa$.
    The black dotted line is the performance of \RLSPR reported in Figure \ref{fig:clf-results}.
    }
    \label{fig:clf-rlpr-comparison}
\end{figure}

We can improve the performance of \RLPRK  if the number of hypotheses $K$ increases,
 and if we vary the changepoint probability threshold $\kappa$,
as shown in Figure \ref{fig:clf-rlpr-comparison}.
 %We see that its optimal performance is when the probability of a changepoint is roughly 60\%. %We remark that the performance of \RLSPR is a lower bound for all the above combinations.
However, even then the performance of this method  still does not match our method.

\subsubsection{Online classification with drift and jumps}
\label{exp:classification-jumps}

In this section we study the performance of \CACI, \CPPD, 
\RLPR, and \RLSPR for an experiment with drift and sudden changes.
More precisely, we assume that the parameters of a logistic regression problem evolve according to
\begin{equation}
\vtheta_t =
\begin{cases}
\vtheta_{t-1} + \vepsilon_t & \text{w.p. } 1 - p_\epsilon,\\
{\cal U}[-2, 2]^2 & \text{w.p. } p_\epsilon,
\end{cases}
\end{equation}
with $p_\epsilon = 0.01$,
$\vtheta_0 \sim {\cal U}[-2, 2]^2$, and
$\vepsilon_t$ is  a zero-mean distributed random vector with isotropic covariance matrix $(0.01)^2\,\vI_2$
(where $\vI_2$ is a $2\times 2$ identity matrix).
Intuitively, this experiment has model parameters that drift slowly with occasional abrupt changes (at a rate of $0.01$).

\begin{figure}[H]
    \centering
    \includegraphics[width=0.6\linewidth]{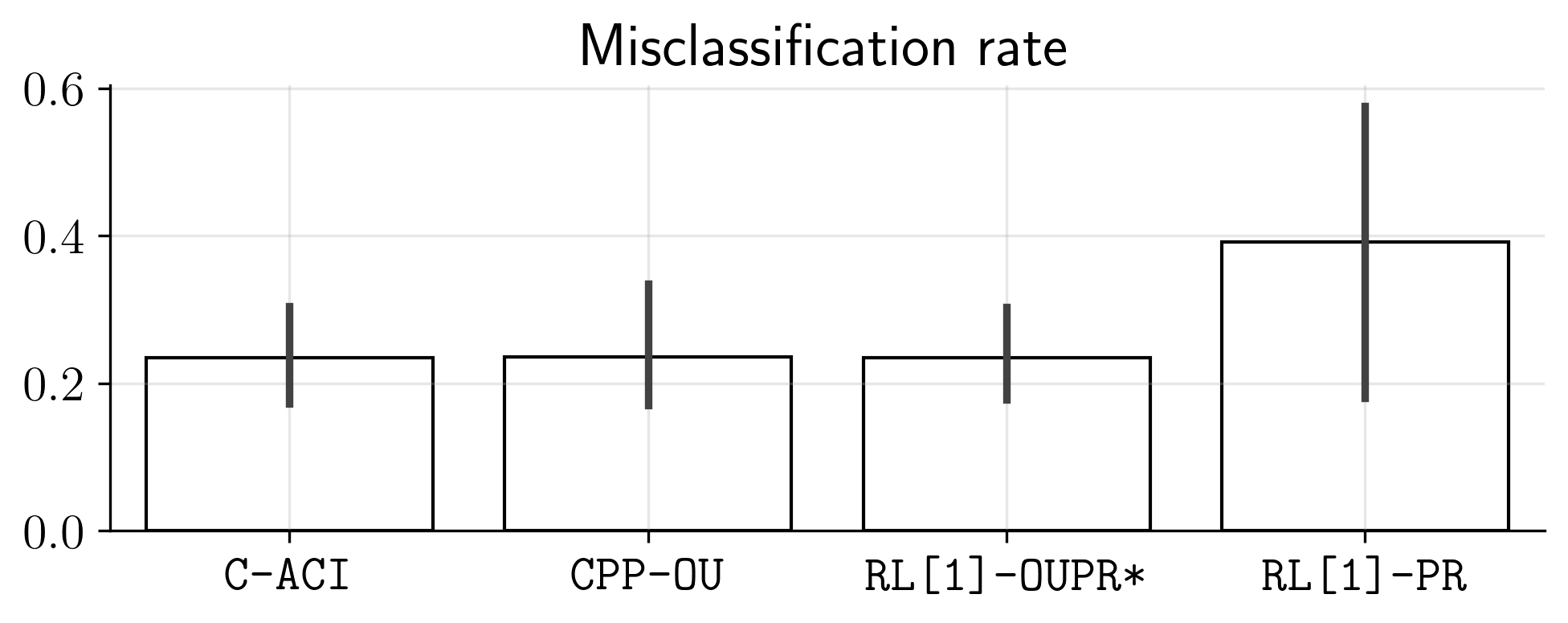}
    \caption{
        Misclassification rate of various methods on the online classification with drift and jumps task. 
    }
    \label{fig:clf-results-abrupt}
\end{figure}

Figure \ref{fig:clf-results-abrupt} shows the missclasification rate among the competing methods.
We observe that \CACI, \CPPD, and \RLSPR have comparable performance, whereas \RLPR  is the method with highest misclassification rate. 

\begin{figure}[H]
    \centering
    \includegraphics[width=0.60\linewidth]{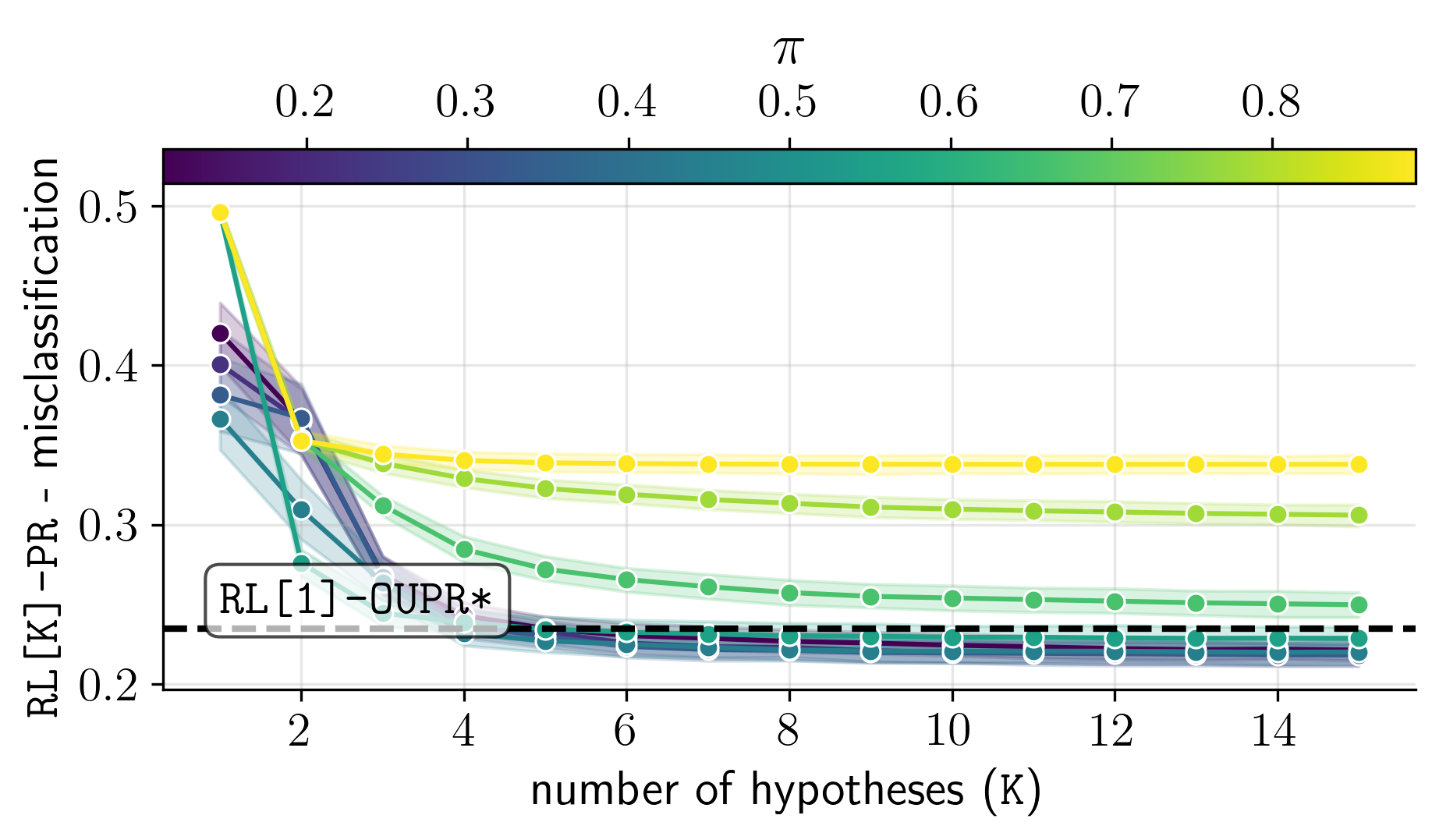}
    \caption{
    Accuracy of predictions for \RLPR[K] as a function of the number of hypotheses ($\texttt{K}$)
    and the probability of a changepoint $\kappa$.
    The black dotted line is the performance of \RLSPR reported in Figure \ref{fig:clf-results-abrupt}.
    }
    \label{fig:clf-rlpr-comparison-abrupt}
\end{figure}
To explain this behaviour, Figure \ref{fig:clf-rlpr-comparison-abrupt}
shows the performance of \RLPR[K] as a function of number of hypotheses and prior
probability of a changepoint $\kappa$.
We observe that up to three hypotheses, the lowest misclassification error of \RLPR[K] is higher than that of \RLSPR,
which only considers one hypothesis.
However, as we increase the number of hypotheses, the best performance for \RLPR[K]  obtains a lower misclassification rate than \RLSPR.
This is in contrast to the results in Figure 5.
Here, we see that with more hypotheses \RLPR[K] outperforms our new method at the expense of being  more memory intensive.

\subsection{Contextual bandits}
\label{experiment:bandits}

In this section, we study the performance of \CACI, \CPPD, \RLPR, and \RLSPR for the 
simple  Bernoulli bandit
from Section 7.3 of \cite{mellor2013changepointthompsonsampling}. 
%(In their example, the context is a constant,
%$\vx_t=c$, which does not affect the reward,
%rendering this a non-contextual bandit.)
More precisely, we consider a multi-armed bandit problem with 10 arms, 10,000 steps per simulation, and 100 simulations. The payoff of a given arm is the outcome of a Bernoulli random variable with unknown probability $\vtheta_t = \min\{\max\{\vtheta_{t-1} + 0.03\,Z_t,0\},1\}$ for $\{Z_t\}_{t\in\{1,2,\dots,10,000\}}$ independent and identically distributed standard normal random variables. We take $\vtheta_0\sim \mathrm{Unif}[0,1]$ and use the same formulation for all ten arms with independence across arms. The observations are the rewards and there are no features (non-contextual).

\begin{figure}[H]
    \centering
    \includegraphics[width=0.6\linewidth]{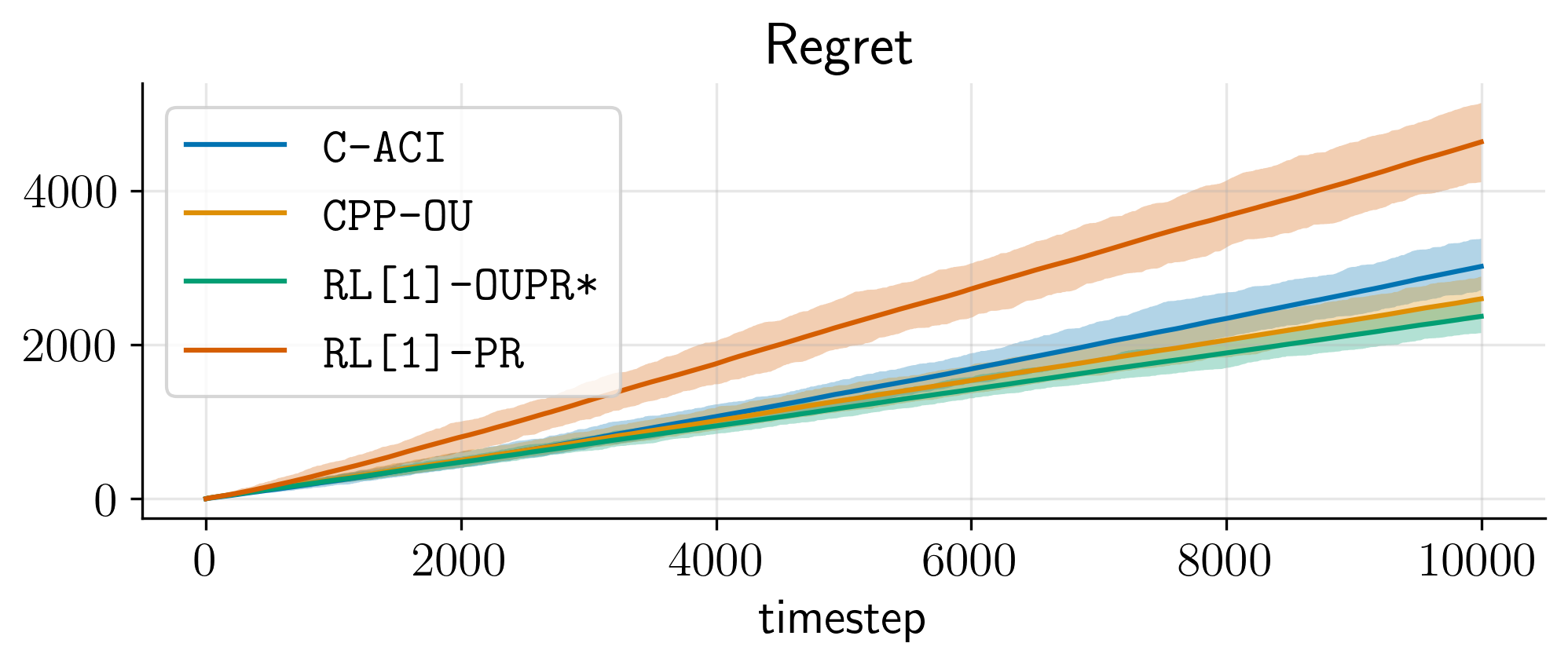}
    \caption{
    Regret of competing methods on the contextual bandits task. Confidence bands are computed with one hundred simulations.
    }
    \label{fig:bandit-results}
\end{figure}

The idea of using \RLPR in multi-armed bandits problems was introduced in \cite{mellor2013changepointthompsonsampling}. With this experiment, we extend the concept to other members of the BONE framework. We use Thompson sampling for each of the competing methods. Figure \ref{fig:bandit-results} shows the regret of using \CACI, \CPPD, \RLPR, and \RLSPR for the above multi-armed bandits  problem.
The results we obtain are similar to those of Section \ref{sec:periodic-drifts}. This is because both problems have a similar drift structure. 

\subsection{Segmentation and prediction}

In this section, we evaluate methods both in terms of their ability to ``correctly'' segment the observed output signal,
and to do one-step-ahead predictions.
Note that by ``correct segmentation'',
we mean one that matches the ground truth data generating process.
This metric can only be applied to synthetic data.

\subsubsection{Autoregression with dependence across the segments}
\label{experiment:KPM}
\label{experiment:segmentation}

% In this experiment we use two well-known datasets with dependence across segments.  First, we consider   the ``Blocks, Heavisine, Bumps, and Doppler'' datasets
% from \cite{donoho1994segmentation-dataset},
% which has been used in \cite{denison1998automatic,fearnhead2007line}. For a description of the dataset see Table 1 in  \cite{donoho1994segmentation-dataset}. 
% We also consider the synthetic dataset introduced in Section 2 of

\begin{figure}[htb]
    \centering
    \includegraphics[width=0.45\linewidth]{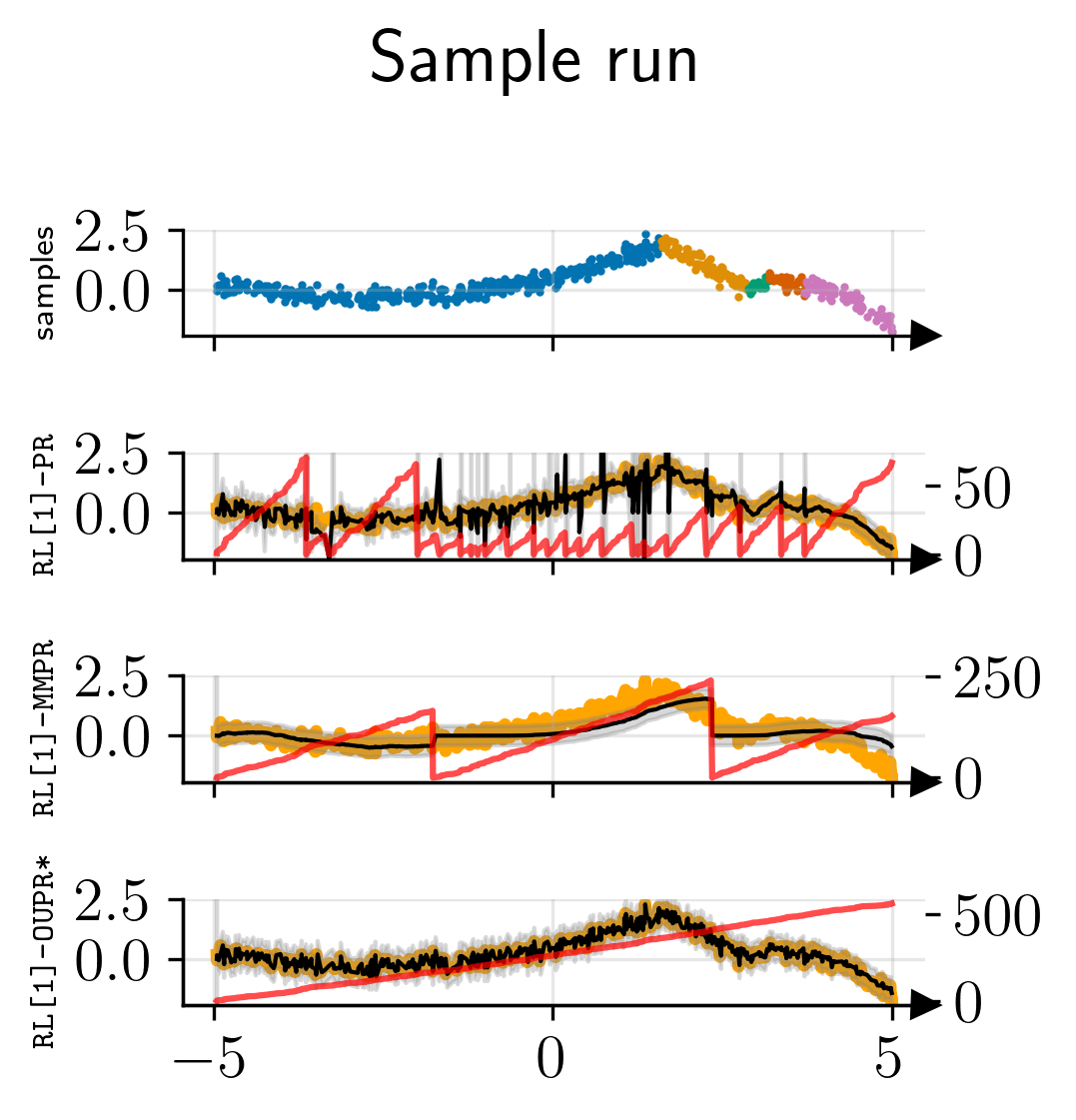}
    \includegraphics[width=0.45\linewidth]{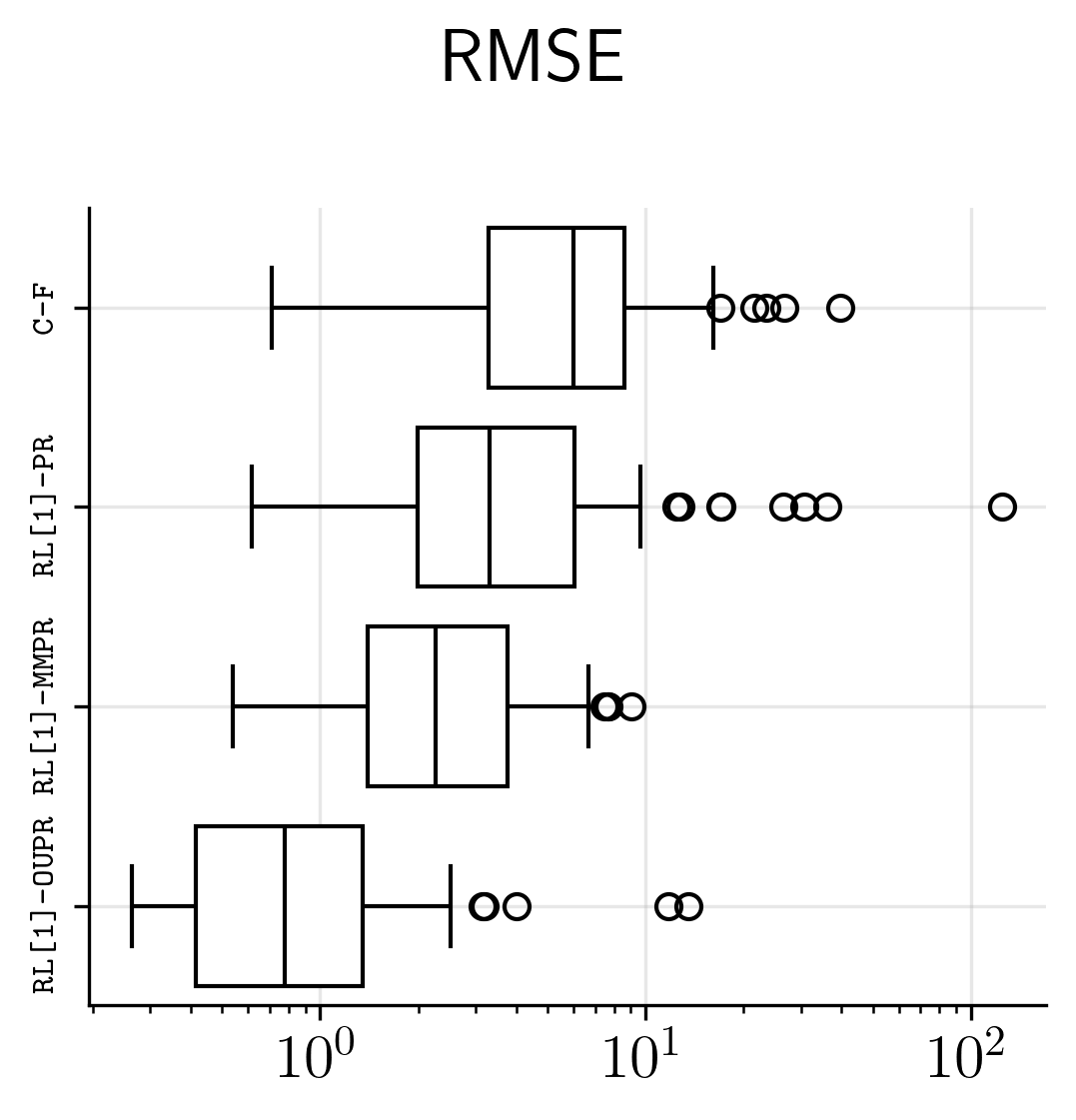}
    \caption{
        The \textbf{left panel} shows
        a sample run of the piecewise polynomial regression with dependence across segments.
        The $x$-axis is for the features,
        the (left) $y$-axis is for measurements together with the  estimations made by \RLPR,
        \namemethod{RL-MMPR}, and \RLSPR,
        the (right) $y$-axis is for the value of $r_t$ under each model.
        The orange line 
        denotes the true data-generating process and the red line denotes the
        value of the hypothesis \texttt{RL}.
        The \textbf{right panel} shows the 
        RMSE of predictions over 100 trials.
    }
    \label{fig:sements-dependency-lr}
\end{figure}

In this experiment,
we consider the synthetic autoregressive dataset introduced in Section 2 of
\cite{fearnhead2011adaptivecp},
consisting of a set of one dimensional polynomial curves that are constrained to match up at segmentation boundaries,
as shown in the top left of Figure \ref{fig:sements-dependency-lr}.

We compare the performance of the three methods in the previous subsection.
For this experiment, we employ a probability of a changepoint $\kappa = 0.01$. 
% For \RLSPR we use the threshold value $\varepsilon=0.5$. We employ one hypothesis for the trajectory of the runlength.
Since this dataset has dependence of the parameters across segments,
 we allow  for the choice of \cModel
to be influenced by the choice of \cAux, i.e., our choice of model is given by $h(\vtheta_t; \auxv_t, \vx_t)$. 
For this experiment, we take \cAux to be \texttt{RL} and our choice of \cModel becomes
\begin{equation}\label{eq:h_fct_aux}
    h(\vtheta_t; r_t, \vx_{1:t}) = \vtheta_t^\intercal\,\vh(\vx_{1:t}, r_t),
\end{equation}
with $\vh(\vx_{1:t}, r_t) = [1, \Delta, \Delta^2]$, $\Delta = (x_t - x_{r_t})$, and $x_{r_t} \geq x_t$.
Intuitively this represents a quadratic curve fit to the beginning $x_{r_t}$ and end points $x_t$  of the current segment.
Given the form of \cModel in  \eqref{eq:h_fct_aux}, here we do not consider \CACI nor \CPPD.
Instead, we use 
runlength with moment-matching prior reset, i.e.,
\namemethod{RL-MMPR} (see Table \ref{tab:rosetta-methods}) which was designed for segmentation with dependence.
%This choice of measurement model $h$ is in similar spirit to a hard version of the attention mechanism in the Transformer model.
% \gdm{Similar in spirit to a hard-version of the attention mechanism of the Transformer model}
% \gdm{See Os96 for more sophisticated models}

%for the experiment in \cite{fearnhead2011adaptivecp}.

Figure \ref{fig:sements-dependency-lr} shows the results.
On the right, 
we observe that \RLSPR has the lowest RMSE.
On the left, we plot the predictions of each method, so we can visualise the nature of their errors.
For  \RLPR, the spikes occur
because the method has many false positive beliefs in a changepoint occurring,
and this causes breaks in the predictions
due  the explicit dependence of $h$ on $r_t$ and the hard parameter reset upon changepoints.
For  \namemethod{RL-MMPR},
the slow adaptation 
(especially when $x_t\in[1,5]$)
is because
 the method does not adjust beliefs as quickly as it should.
Our \RLSPR  method strikes a good compromise.

\begin{figure}[H]
    \centering
    \includegraphics[width=0.60\linewidth]{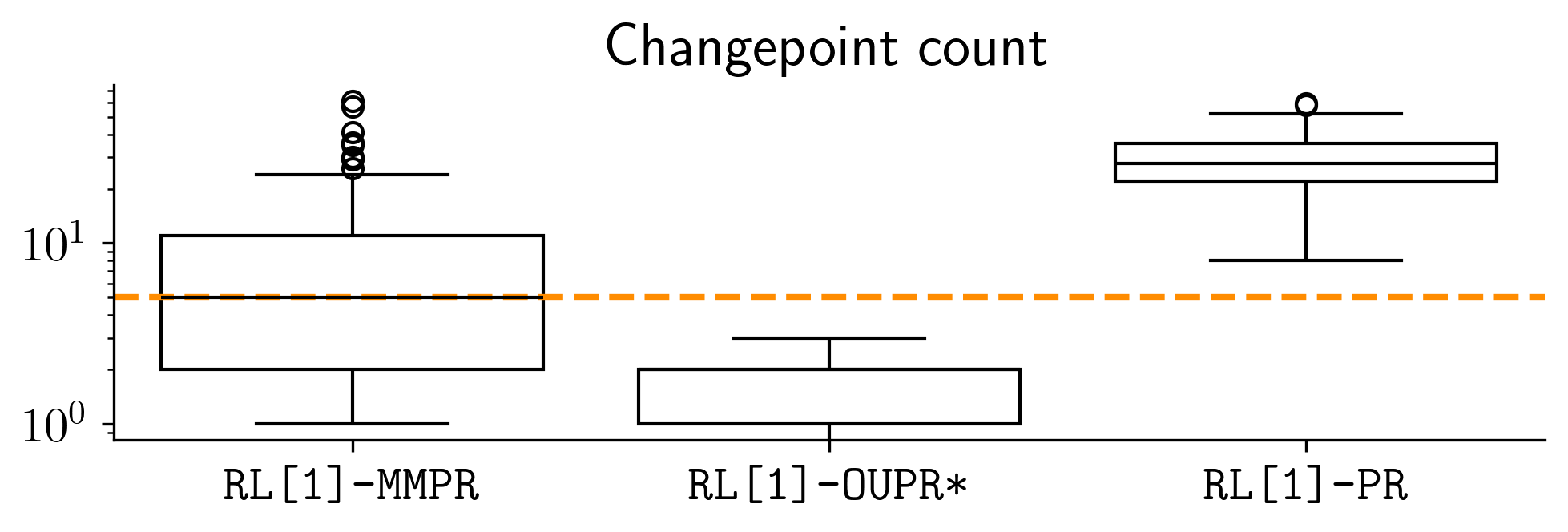}
    \caption{
    Count of changepoints over an experiment for 100 trials.
    The orange line shows the true number of changepoints for all trials.
    }
    \label{fig:segments-count}
\end{figure}

Figure \ref{fig:segments-count} shows the distribution (over 100 simulations) of the number of detected changepoints,
i.e., instances where $\nu_t(r_t)$ with $r_t = 0$ is the highest.
We observe that superior predictive performance 
in Figure \ref{fig:sements-dependency-lr}
does not necessarily translate to a better segmentation capability. For example,  the distribution produced by \namemethod{RL-MMPR} sits around the actual number of changepoints (better at segmenting) whereas \RLSPR, which is detecting far fewer changepoints, is the best performing prediction method.
This reflects the discrepancy between the objectives of segmentation and prediction.
For a more thorough analysis and evaluation of changepoint detection methods on time-series data,
see \cite{van2020evaluation}.
%\gdm{Add explanation on bias term and RL-MMPR. Likelihood can be non-markov if we use RL.}

\subsubsection{Non-stationary heavy-tailed regression with \texttt{DA[inf]}}
\label{experiment:heavy-tail-regression}

It is well-known that the combination \texttt{RL-PR} is sensitive to outliers if the choice of \cModel is misspecified,
since an observation that is ``unusual'' may trigger a changepoint unnecessarily.
As a consequence,
various works have proposed outlier-robust variants to the \RLPR[inf] for segmentation
\citep{knoblauch2018doublyrobust-bocd,fearnhead2019robustchangepoint, altamirano2023robust,sellier2023robustbocdgp}
and for filtering \citep{reimann2024changingfiltering}.
\eat{
In this experiment, however, we show that an outlier-robust variant of \RLPR[inf] for prediction can be created from the
filtering literature by changing the choice of  \cPosterior.
Specifically, because the choice of \cPosterior does not depend on \cPrior or \cAux,
we can make use of the vast literature for outlier-robust filtering in measurement space
\citep[see e.g.,][]{ting2007learning, agamennoni2012, piche2012, huang2016, wang2018}
to construct a robust and adaptive online learner by fixing \cAux and \cPrior to be \RLPR[inf] and changing \cPosterior.
}
In what follows, we show how we can easily accomodate robust methods into the BONE framework by changing the way we compute the likelihood
and/or posterior.
In particular, we 
consider the WoLF-IMQ method of \cite{duranmartin2024-wlf}.
We use  WoLF-IMQ 
because it is a provably robust algorithm and it is a 
straightforward modification of the  linear Gaussian posterior update equations.
% As written in Table \ref{tab:rosetta-methods},
We denote
\RLPR[inf] with \cPosterior taken to be \texttt{LG} as \RLPRKF and
\RLPR[inf] with \cPosterior taken to be WoLF-IMQ as \RLPRWoLF.
% For this experiment, we consider the full set of hypotheses for each \texttt{RL} method,
% i.e., the total number of computations increases at a linear rate in time,
% as specified in Table \ref{tab:auxv-time-complexity}.

\begin{figure}[htb]
    \centering
    \includegraphics[width=0.8\linewidth]{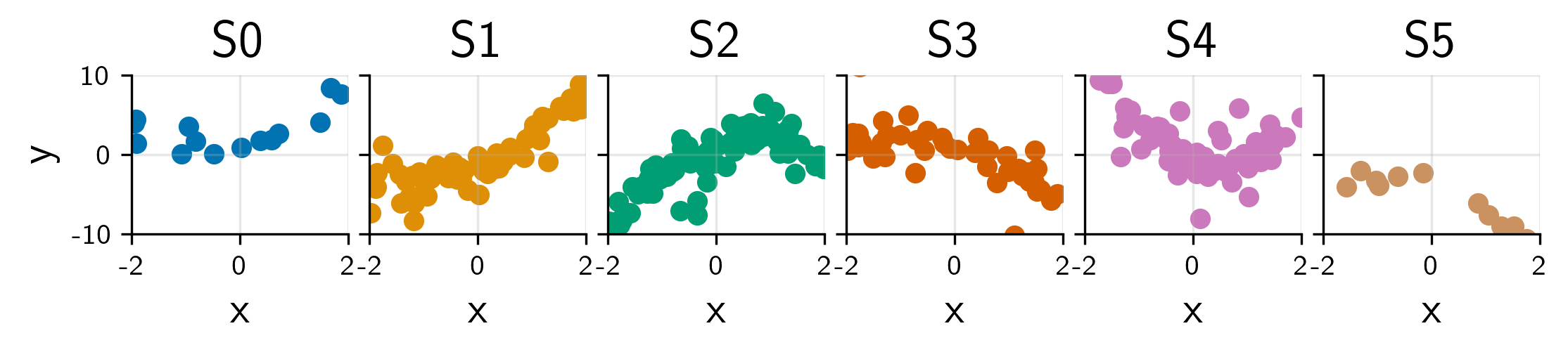}
    \vspace{-2em}
    \caption{
        Sample run of the heavy-tailed-regression process.
        Each box corresponds to the samples within a segment.
    }
    \label{fig:segements-tdlist-lr}
\end{figure}

To demonstrate the utility of a robust method,
we consider a 
piecewise linear regression model with Student-$t$ errors,
where the measurement are sampled according to
$\vx_t \sim {\cal U}[-2, 2]$,
$ \vy_t \sim {\rm St}\big( \phi(\vx_t)^\intercal\vtheta_t, 1,\,\, 2.01 \big)$
a Student-$t$ distribution with location $ \phi(\vx_t)^\intercal\vtheta_t$,
scale $1$,
degrees of freedom $2.01$, and
$\phi(\vx_t) = (1,\,x,\,x^2)$.
At every timestep, the parameters take the value
\begin{equation}
\vtheta_t =
\begin{cases}
\vtheta_{t-1} & \text{w.p. } 1 - p_\epsilon,\\
{\cal U}[-3, 3]^3 & \text{w.p. } p_\epsilon,
\end{cases}
\end{equation}
with $p_\epsilon = 0.01$, and $\vtheta_0 \sim {\cal U}[-3, 3]^3$. 
Intuitively, at each timestep, there is probability $p_\epsilon$  of a changepoint, and conditional on a changepoint occurring, the each of the entries of the new parameters $\vtheta_t$ are sampled from a uniform in $[-3,3]$. 
Figure \ref{fig:segements-tdlist-lr} shows a sample data generated by this process.

To process this data, our choice of \cModel is
$h(\vtheta_t, \vx_t) = \vtheta_t^\intercal\,\phi(\vx_t)$
with 
\begin{equation}
    \ell(\vy_t; \vtheta_t, \vx_t)
    = -W^2\big(\vy_t, h(\vtheta_t, \vx_t)\big)\,\log{\cal N}(\vy_t \cond h(\vtheta_t, \vx_t), 1.0),
\end{equation}
a weighted Gaussian log-likelihood and
$W(u, z) = (1 + \frac{(u - z)^2}{c^2})^{-1/2}$ the inverse multi-quadratic (IMQ) function
with soft threshold value $c=4$, 
representing four standard deviations of tolerance to outliers.
Here $u,z \in \real$.

\begin{figure}[htb]
    \centering
    \includegraphics[width=0.48\linewidth]{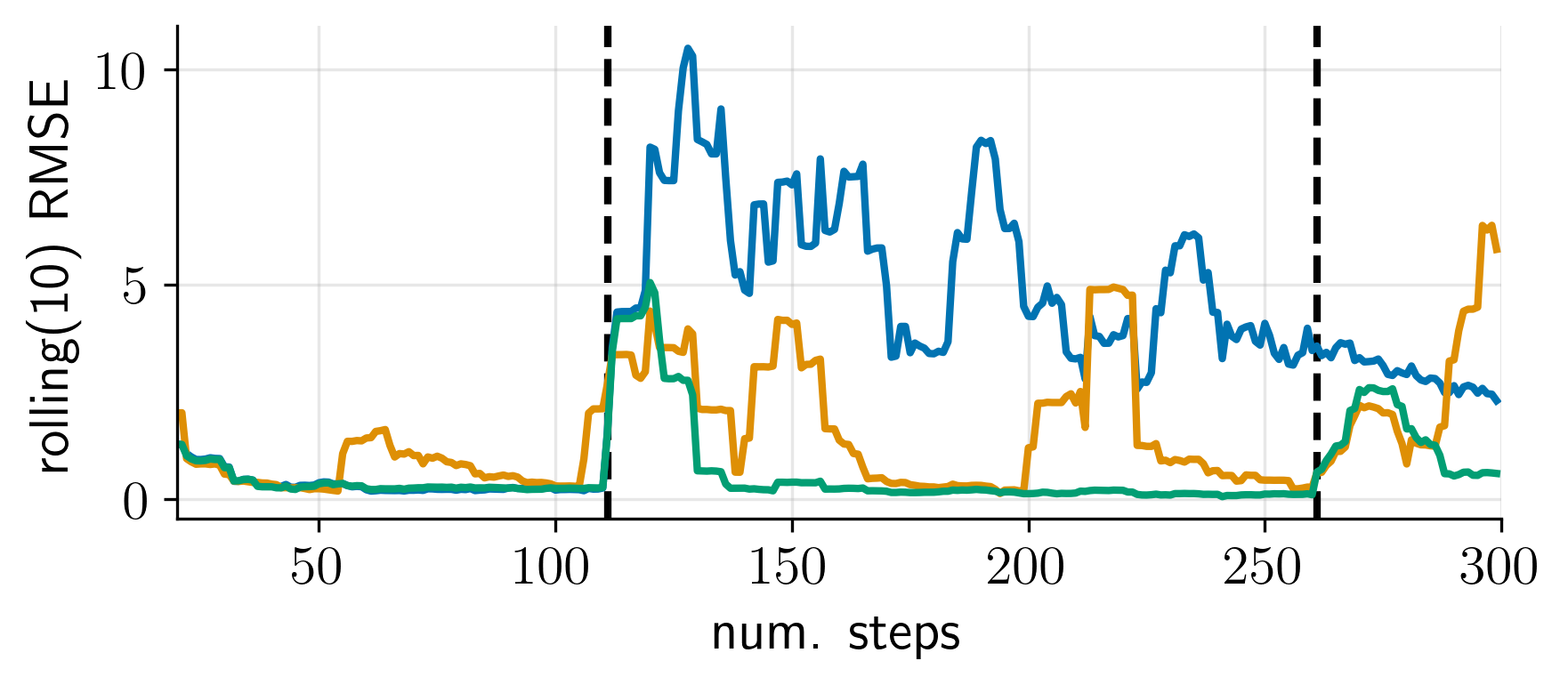}
    \includegraphics[width=0.48\linewidth]{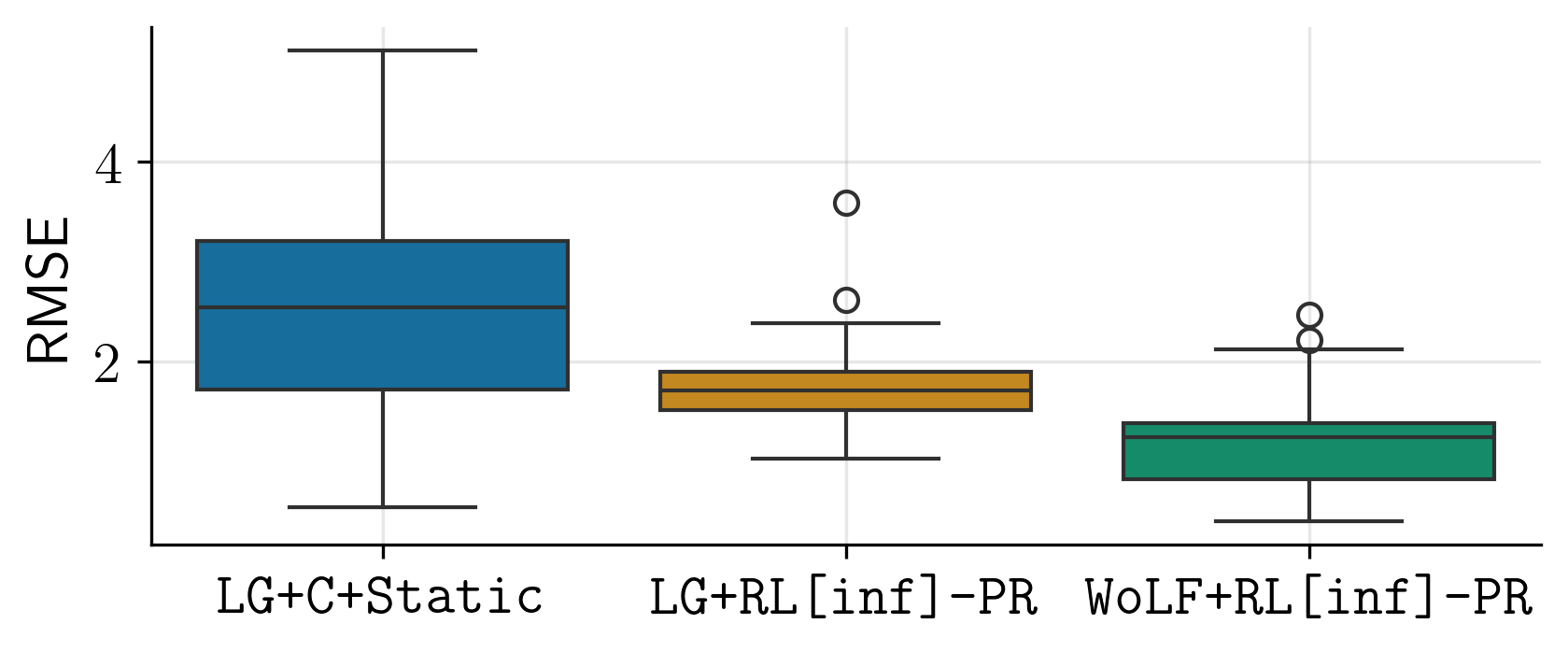}
    \caption{
    The \textbf{left panel} shows the rolling RMSE using a window of the 10 previous observations.
    The \textbf{right panel} shows the distribution of final RMSE over 30 runs.
    The vertical dotted line denotes a change in the true model parameters.
    }
    \label{fig:outliers-lr-res}
\end{figure}

The left panel in Figure \ref{fig:outliers-lr-res} shows
the rolling mean (with a window of size 10) of the RMSE for
\RLPRKF, \RLPRWoLF, and \staticKF.
The right panel in Figure \ref{fig:outliers-lr-res}
shows the distribution of the RMSE for all methods after 30 trials.

The left panel of Figure \ref{fig:outliers-lr-res} shows that
\staticKF has a lower rolling RMSE error than \RLPRKF up to first changepoint (around 100 steps).
% The higher rolling RMSE of \RLPR[inf] before the \textit{true} changepoint
% is a consequence of the false positives of a changepoints, which induces a parameter reset.
The performance of \staticKF significantly deteriorates afterwards.
Next, \RLPRKF wrongly detects changepoints and resets its parameters frequently.
This results in periods of increased rolling RMSE.
Finally, \RLPRWoLF has the lowest error among the methods.
After the regime change, its error increases at a similar rate to the other methods,
however, it correctly adapts to the regime and its error decreases soon after the changepoint.

\begin{figure}[htb]
    \centering
    \includegraphics[width=0.48\linewidth]{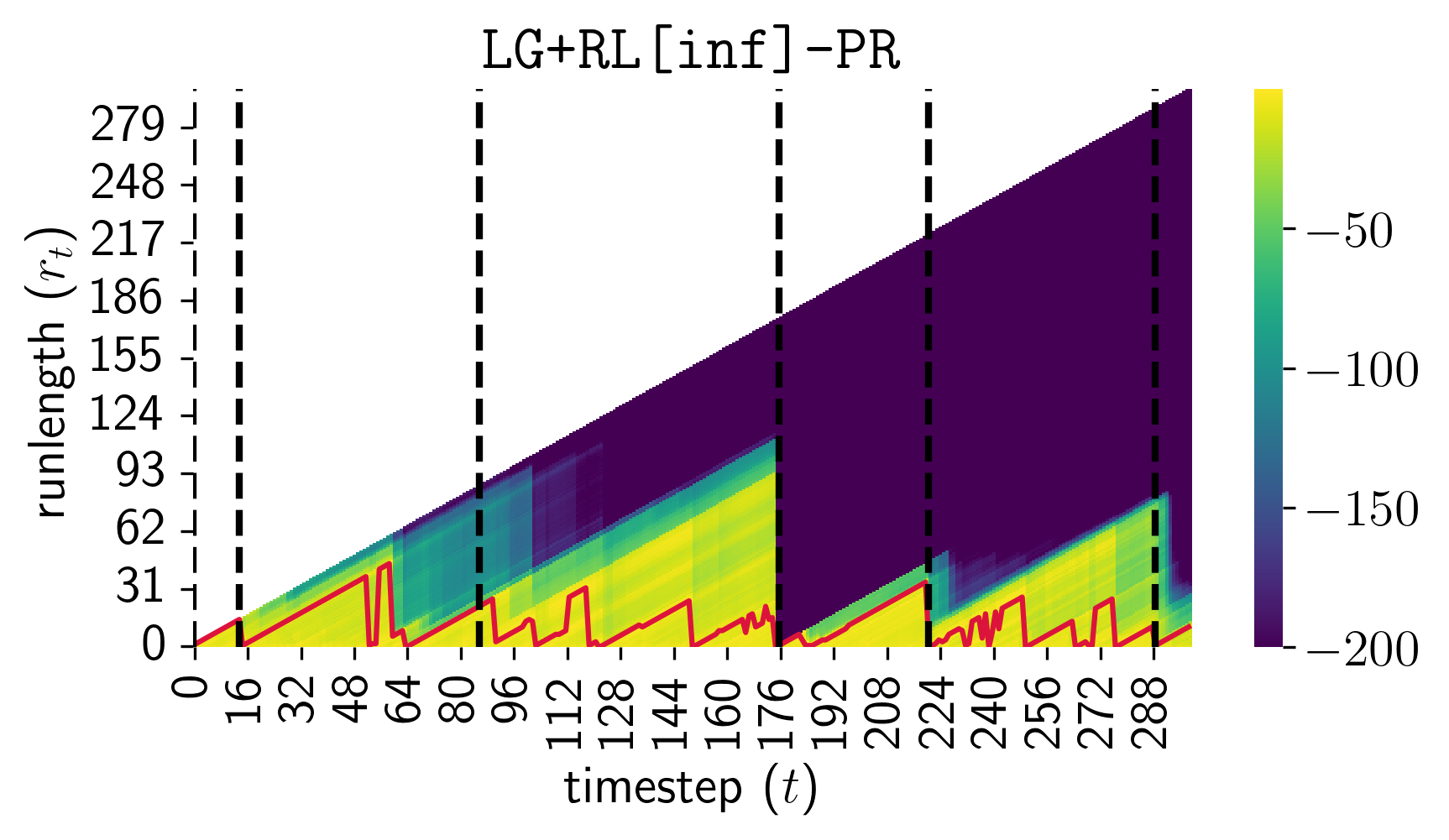}
    \includegraphics[width=0.48\linewidth]{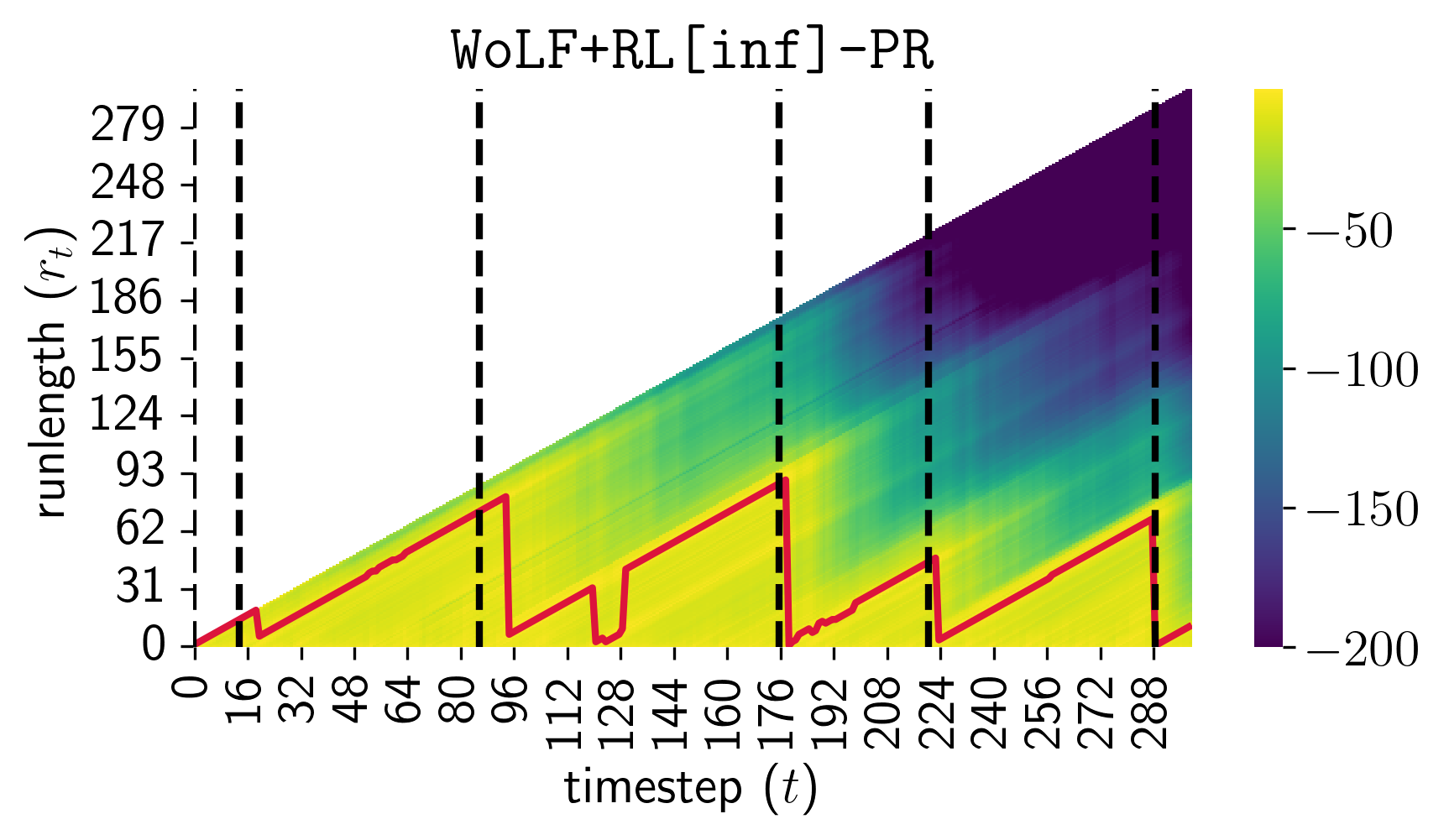}
    \caption{
        Segmentation of the non-stationary linear regression problem.
        The left panel shows the segmentation done by \RLPRKF.
        The right panel shows the segmentation done by \RLPRWoLF.
        The $x$-axis is the timestep $t$, the $y$-axis is the runlength $r_t$ (note that it is always the case that $r_t\leq t$), and the color bar shows the value $\log\,p(r_t \cond \vy_{1:t})$.
        The red line in either plot is the trajectory of the mode, i.e.,
        the set $r_{1:t}^* = \{\argmax_{r_{1}} p(r_1 \cond \data_{1}),  \ldots, \argmax_{r_{t}} p(r_t \cond \data_{1:t})\}$.
        Note that the non-robust method (left) oversegments the signal.
        See \href{https://youtu.be/hLY93RTQejQ}{this url}
        for a video comparison between \RLPRKF and \RLPRWoLF.
    }
    \label{fig:bocd-lr-stress}
\end{figure}

Figure \ref{fig:bocd-lr-stress} shows
the posterior belief of the value of the runlength  
using \RLPRKF and \RLPRWoLF.
The constant reaction to outliers in the case of \RLPRKF means that the parameters keep reseting back to the initial prior belief.
As a consequence, the RMSE of \RLPRKF deteriorates. 
On the other hand, \RLPRWoLF  resets less often, and accurately adjusts to the regime changes when they do happen. This results in the lowest RMSE among the three methods.

%% file: sections/conclusions.tex
\section{Conclusions}

% We carried out a literature review on methods that perform online predictions in non-stationary environments.
% We identified three inductive biases that drive a large portion of this literature.
% The inductive biases are: a model over parameters, a model over measurements, and a latent variable that tracks non-stationarity. In general, the framework we put forward is composed of three modelling choices and two algorithmic choices.

We introduced a unified Bayesian framework to  perform online predictions in non-stationary environments,
and showed how it covers many prior works.
We also used our framework
to design a new method,  \RLSPR, which is suited to tackle prediction problems
when the observations exhibit both abrupt and gradual changes.
In future work, we aim to investigate other novel variants and applications to newer model architectures
such as transformers and graph neural networks \citep{moreno-pino2024roughtransformer,arroyo2025vanishing}.
% and \newmethod{WoLF+RL-PR} (or more generally, \namemethod{WoLF} and any member of the BONE framework).
%In particular, \RLSPR performs well when the observations exhibit both abrupt and gradual changes.
%, and
% \newmethod{WolF+RL-PR} does well when there are abrupt changes and there are outliers in the data.

\section*{Aknowledgments}
We thank
Matias Altamirano,
François-Xavier Briol,
Patrick Chang,
Alex Galashov,
Matt Jones,
Jeremias Knoblauch, and
Radford Neal
for insightful comments and discussions
on earlier versions of this paper.

%% file: sections/appendix/weights-caux.tex
\section{Worked examples for BONE methods}
\label{sec:caux-weights}
In this section, we provide a detailed calculation of $\nu_{t}(\auxv_t)$ for some choices of $\auxv_t$.
We consider a choice of \cModel to be linear Gaussian with known observation variance $\vR_t$, i.e.,
\begin{equation}\label{eq:linear-gaussian-measurement-model}
    p(\vy_t \cond \vtheta, \vx_t) = {\cal N}(\vy_t \cond \vx_t^\intercal\,\vtheta_t, \vR_t).
\end{equation}

\subsection{Runlength with prior reset (\texttt{RL-PR})}
\label{sec:rl-pr-implementation}

\subsubsection{Unbounded number of hypotheses \texttt{RL[inf]-PR}}
The work in \cite{adams2007bocd} takes
$\auxv_t = r_t$ to be the runlength, with $r_t \in \{0, 1, \ldots, t\}$, that
that counts the number of steps since the last changepoint.
Assume the runlength follows the dynamics \eqref{eq:transition-rl-markov}.
We consider $\nu_{t}(r_{t}) = p(r_t | \data_{1:t})$ such that
\begin{equation}
    p(r_t \cond \data_{1:t}) =
    \frac{p(r_t, \data_{1:t})}{\sum_{\hat{r}_t=0}^t p(\hat{r}_t, \data_{1:t})},
\end{equation}
for $r_t \in \{0, \ldots, t\}$.
The \texttt{RL-PR} method estimates $p(r_t, \data_{1:t})$ for all $r_t \in \{0, \ldots, t\}$ at every timestep.
To estimate this value recursively, we  sum over all possible previous runlengths as follows
\begin{equation}\label{eq:prrl-lj}
\begin{aligned}
    &p(r_t,  \data_{1:t})\\
    &= \sum_{r_{t-1}=0}^{t-1} p(r_t,  r_{t-1}, \data_{1:t-1}, \data_t)\\
    &= \sum_{r_{t-1}=0}^{t-1} p(r_{t-1} , \data_{1:t-1})
    \,p(r_t \cond r_{t-1}, \data_{1:t-1})\, p(\vy_t \cond r_t, r_{t-1},\vx_t, \data_{1:t-1})\\
    &= p(\vy_t \cond r_t , \vx_t, \data_{1:t-1}) \sum_{r_{t-1}=0}^{t-1}
    p(r_{t-1} , \data_{1:t-1})
    p(r_t\cond r_{t-1}).
\end{aligned}
\end{equation}
In the last equality, there are two implicit assumptions,
(i) the runlength at time $t$ is conditionally independent of the data $\data_{1:t-1}$ given the runlength at time $t-1$, and
(ii) the model is Markovian in the runlength, that is, conditioned on $r_t$, the value of $r_{t-1}$ bears no information.
Mathematically, this means that (i)  $p(r_t \cond r_{t-1}, \data_{1:t-1}) = p(r_t \cond r_{t-1})$
and (ii) $p(\vy_t \cond r_t, r_{t-1}, \data_{1:t-1}) = p(\vy_t \cond r_t, \data_{1:t-1})$. 
From \eqref{eq:prrl-lj}, we observe there are only two possible scenarios for the value of $r_t$. Either 
$r_t = 0$ or $r_t = r_{t-1} + 1$
with $r_{t-1} \in \{0, \ldots, t-1\}$.
Thus, 
%\eqref{eq:prrl-lj} 
$p(r_t, \data_{1:t})$ becomes
\begin{equation}\label{eq:bocd-joint}
    \begin{aligned}
        p(r_t, \data_{1:t}) &= p(\vy_t \cond r_t, \vx_t, \data_{1:t-1})\, p(r_{t-1}, \data_{1:t-1})\, p(r_t \cond r_{t-1})
        & \text{ if }r_t \geq 1\\
        p(r_t, \data_{1:t}) &= p(\vy_t \cond  r_t, \vx_t, \data_{1:t-1}) \sum_{r_{t-1}=0}^{t-1} p(r_{t-1}, \data_{1:t-1}) \, p(r_t \cond r_{t-1}) & \text{ if } r_t = 0\,.
    \end{aligned}
\end{equation}
The joint density \eqref{eq:bocd-joint} considers two possible scenarios:
either we stay in a regime considering the past $r_t \geq 1$ observations,
or we are in a new regime, in which $r_t = 0$.
Finally, note that \eqref{eq:bocd-joint} depends on three terms:
(i) the transition probability $p(r_t \cond r_{t-1})$, which it is assumed to be known,
(ii) the previous log-joint $p(r_{t-1}, \data_{1:t-1})$, with $r_{t-1} \in \{0, 1, \ldots, t-1\}$,
which is estimated at the previous timestep, and
(iii) the prior predictive density
\begin{equation}\label{eq:rl-predictive}
    p(\vy_t \cond r_t, \vx_t, \data_{1:t-1}) = \int
    p(\vy_t \cond \vtheta_{t}, \vx_t)
    \, p(\vtheta_{t} \cond r_t, \data_{1:t-1}) \d\vtheta_t.
\end{equation}
For a choice of \cModel given by \eqref{eq:linear-gaussian-measurement-model} and
a choice of \cPrior given by \eqref{eq:cprior-rl-pr},
the posterior predictive \eqref{eq:rl-predictive} takes the form.
\begin{equation}
\begin{aligned}
    p(\vy_t \cond r_t, \vx_t, \data_{1:t-1})
    &= \int {\cal N}\left(\vy_t \cond \vx_t^\intercal\,\vtheta_t, \vR_t\right)\,{\cal N}\left(\vtheta_t \cond \vmu_{t-1}^{(r_t)}, \vSigma_{t-1}^{(r_t)}\right) \d\vtheta_t\\
    &= {\cal N}\left(\vy_t \cond \vx_t^\intercal \vmu_{t-1}^{(r_{t})},\,\vx_t^\intercal\,\vSigma_{t-1}^{(r_{t})}\,\vx_t + \vR_t\right),
\end{aligned}
\end{equation}
with $r_t \in \{0, \ldots, t-1\}$.
Here,
$\left(\vmu_{t-1}^{(r_t)}, \vSigma_{t-1}^{(r_t)}\right)$ are the posterior mean and covariance
at time $t-1$ built using the last $r_t \geq 1$ observations.
If $r_t = 0$, then $(\vmu_{t-1}^{(r_t)}, \vSigma_{t-1}^{(r_{t})}) = \left(\vmu_0, \vSigma_0\right)$.

\subsubsection{Bounded number of hypotheses \texttt{RL[K]-PR}}
If we maintain a set of $K$ possible hypotheses, then $\vPsi_t = \{r_{t-1}^{(1)}, \ldots, r_{t-1}^{(K)}\} \in \{0, \ldots, t-1\}^K$
is a collection of $K$ unique runlengths obtained at time $t-1$.
Next, \eqref{eq:prrl-lj} takes the form
\begin{align}
        p(r_t, \data_{1:t}) &= p(\vy_t \cond r_t, \vx_t, \data_{1:t-1})\, p(r_{t-1}, \data_{1:t-1})\, p(r_t \cond r_{t-1})
        & \text{ if }r_t \geq 1, \label{eq:bocd-joint-fixed-K-up}\\
        p(r_t, \data_{1:t}) &= p(\vy_t \cond  r_t, \vx_t, \data_{1:t-1}) \sum_{r_{t-1} \in \vPsi_{t-1}} p(r_{t-1}, \data_{1:t-1}) \, p(r_t \cond r_{t-1}) & \text{ if } r_t = 0 \label{eq:bocd-joint-fixed-K-down}\,.
\end{align}
Here, we have that either $r_t = r_{t-1} + 1$ when $r_{t-1} \in \vPsi_{t-1}$ or $r_{t} = 0$. 
After computing $p(r_t, \data_{1:t})$ for all $K+1$ possibles values of $r_t$, a choice is made to keep $K$ hypotheses.
For timesteps $t \leq K$, we evaluate all possible hypotheses until $t > K$.

Algorithm \ref{algo:rl-pr-step} shows an update step under this process when we maintain a set of $K$ possible
hypotheses.

\subsection{Runlength with moment-matched prior reset (\texttt{RL-MMPR})}
\label{sec:rl-mmpr-implementation}
Here, we consider a modified version of the method introduced in \cite{fearnhead2011adaptivecp}.
We consider the choice of \texttt{RL} and
adjust the choice of \cPrior for \texttt{RL-PR} introduced in Appendix \ref{sec:rl-pr-implementation} whenever $r_t =0$.
In this combination, for $r_t = 0$, we take
$\tau(\vtheta_t \cond r_t, \data_{1:t-1}) = p(\vtheta_t \cond r_t, \data_{1:t-1})$.
Next
\begin{equation}\label{eq:mmpr-conditional-prior}
\begin{aligned}
    p(\vtheta_t \cond r_t, \data_{1:t-1})
    &= \sum_{r_{t-1}=0}^{t-1} p(\vtheta_t, r_{t-1} \cond r_t, \data_{1:t-1})\\
    &= \sum_{r_{t-1}=0}^{t-1} p(r_{t-1} \cond \data_{1:t-1})\,p(r_t \cond r_{t-1})\,p(\vtheta_t \cond r_t, r_{t-1}, \vy_{1:t-1})\\
    &= \sum_{r_{t-1}=0}^{t-1} p(r_{t-1} \cond \data_{1:t-1})\,p(r_t \cond r_{t-1})\,{\cal N}(\vtheta_t \cond \vmu_{t-1}^{(r_{t-1})}, \vSigma_{t-1}^{(r_{t-1})}).
\end{aligned}
\end{equation}
Because \eqref{eq:mmpr-conditional-prior} is a mixture model, we choose a conditional prior to be Gaussian that approximates the
first two moments.
We obtain
\begin{equation}\label{eq:rl-mmpr-first-moment}
    \mathbb{E}[\vtheta_t \cond r_t, \vy_{1:t-1}] =
    \sum_{r_{t-1}=0}^{t-1} p(r_{t-1} \cond \data_{1:t-1})\,p(r_t \cond r_{t-1})\,\vmu_{t-1}^{(r_{t-1})}
\end{equation}
for the first moment, and
\begin{equation}\label{eq:rl-mmpr-second-moment}
    \mathbb{E}[\vtheta_t\,\vtheta_t^\intercal \cond r_t, \vy_{1:t-1}]
    \sum_{r_{t-1}=0}^{t-1} p(r_{t-1} \cond \data_{1:t-1})\,p(r_t \cond r_{t-1})
    \left(\vSigma_{t-1}^{(r_{t-1})} + \vmu_{t-1}^{(r_{t-1})}\,\vmu_{t-1}^{(r_{t-1})\intercal}\right)
\end{equation}
for the second moment.
The conditional prior mean and prior covariance under $r_t = 0$ take the form
\begin{equation}\label{eq:rl-mmpr-prior-reset}
\begin{aligned}
    \vmu_t^{(0)} &= \mathbb{E}[\vtheta_t \cond r_t, \vy_{1:t-1}],\\
    \vSigma_t^{(0)} &= \mathbb{E}[\vtheta_t\,\vtheta_t^\intercal \cond r_t, \vy_{1:t-1}] - \left(\mathbb{E}[\vtheta_t \cond r_t, \vy_{1:t-1}]\right)\left(\mathbb{E}[\vtheta_t \cond r_t, \vy_{1:t-1}]\right)^\intercal.
\end{aligned}
\end{equation}
Algorithm \ref{algo:rl-mmpr-step} shows an update step under this process when we maintain a set of $K$ possible
hypotheses.

\subsection{Runlength with OU dynamics and prior reset (\texttt{RL[1]-OUPR*})}
\label{sec:rl-spr-implementation}
In this section, we provide pseudocode for the new hybrid method we propose.
Specifically, our choices in BONE are:
\RLSPR for \cAux and \cPrior,
\texttt{LG} for \cPosterior, and
\texttt{DA[1]} for \cWeight.
Because of our choice of \cWeight, \RLSPR considers a single hypothesis (or runlength) which,
at every timestep, is either increased by one or set back to zero, according to the probability of a changepoint
and a threshold $\epsilon \in (0,1)$.

In essence, \RLSPR follows the logic behind \RLPR introduced in Section \ref{sec:rl-pr-implementation}
with $K=1$ hypothesis and different choice of \cPrior.
To derive the algorithm for \RLSPR at time $t > 1$, suppose $r_{t-1}$ is available
(the only hypothesis we track).
Denote by $r_t^{(1)}$ the hypothesis of a runlength increase, i.e., $r_t = r_{t-1} + 1$ and
denote by $r_t^{(0)}$ the hypothesis of a runlenght reset, i.e., $r_t = 0$.
The probability of a runlength increase under a single hypothesis takes the 
form:
\begin{equation}
\begin{aligned}
    \nu_t(r_t^{(1)})
    &= p(r_{t}^{(1)} \cond \data_{1:t})\\
    &= \frac{p(r_t^{(1)}, \data_{1:t})}{p(r_t^{(1)}, \data_{1:t}) + p(r_t^{(0)}, \data_{1:t})}\\
    &=
    \frac
    {p(\vy_t \cond r_t^{(1)}, \vx_t, \data_{1:t-1})\,p(r_{t-1}, \data_{1:t-1})\,(1 - \kappa)}
    {p(\vy_t \cond r_t^{(0)}, \vx_t, \data_{1:t-1})\,p(r_{t-1}, \data_{1:t-1})\,\kappa
    + p(\vy_t \cond r_t^{(1)}, \vx_t, \data_{1:t-1})\,p(r_{t-1}, \data_{1:t-1})\,(1 - \kappa)}
    \\
    &= 
    \frac{p(\vy_t \cond r_t^{(1)}, \vx_t, \data_{1:t-1})\,(1 - \kappa)}{p(\vy_t \cond r_t^{(0)}, \vx_t, \data_{1:t-1})\,\kappa + p(\vy_t \cond r_t^{(1)}, \vx_t, \data_{1:t-1})\,(1-\kappa)}.
    \label{eq:rl-spr-eq-posterior}
\end{aligned}
\end{equation}
where $\kappa=p(r_t \cond r_{t-1})$ with $r_t = 0$ is the prior probability of a changepoint and
and $1-\kappa=p(r_t \cond r_{t-1}$ with $r_t=r_{t-1}+1$
is the probability of continuation of the current segment.

Next, we use $\nu_t(r_t)$ to decide whether to update our parameters 
or reset them according to a prior belief according to some threshold $\epsilon$.
This implements our choice of \cPrior given in \eqref{eq:SPR-equation-gt}  and \eqref{eq:SPR-equation-Gt}.
Because we maintain a single hypothesis, the weight at the end of the update step is set to $1$.
Algorithm \ref{algo:rl-spr-step} shows an update step for \RLSPR under the choice of \cModel given by \eqref{eq:linear-gaussian-measurement-model}.

\subsection{Changepoint location with multiplicative covariance inflation \texttt{CPL-MCI}}
\label{c-aux:CPL}
The work in \cite{li2021onlinelearning}
takes $\auxv_t = s_{1:t}$ to be a $t$-dimensional vector where the
$i$-th element is a binary vector that determines a changepoint at time $t$.
Then, the sum of the entries of $s_{1:t}$ represents the total number of changepoints up to, and including, time $t$.

We take $\nu_{t}(s_{1:t}) = p(s_{1:t} | \data_{1:t})$,
which is recursively expressed as
\begin{equation}\label{eq:bcm-recursive-changepoint}
\begin{aligned}
    p(s_{1:t} \cond \data_{1:t})
    &= p(s_t, s_{1:t-1} \cond \vy_{t}, \vx_t, \data_{1:t-1})\\
    &= p(s_{1:t-1} \cond \data_{1:t-1}) p(s_t \cond s_{1:t-1}, \vx_t, \vy_t, \data_{1:t-1}).
\end{aligned}
\end{equation}
Here, $p(s_{1:t-1} \cond \data_{1:t-1})$ is inferred at the previous timestep $t-1$.
The estimate of a changepoint conditioned on the past changes and the  measurements  is
\begin{equation}
\begin{aligned}
    &p(s_t = 1 \cond s_{1:t-1}, \data_{1:t})\\
    &\qquad = \frac{p(s_t = 1)p(\vy_t \cond \vx_t, s_{1:t-1}, s_t=1, \data_{1:t-1})}
    {p(s_t = 1)p(\vy_t \cond s_t=1, \vx_t, s_{1:t-1}, \vy_{1:t-1}) + p(s_t = 0)p(\vy_t \cond s_t=0, \vx_t, s_{1:t-1}, \data_{1:t-1})}\\
    &\qquad =
    \left(1+ {\exp\left(-\log\left(
    \frac{p(s_t = 1)p(\vy_t \cond s_t=1, \vx_t, s_{1:t-1}, \vy_{1:t-1})}{p(s_t = 0)p(\vy_t \cond s_t=0, \vx_t, s_{1:t-1}, \data_{1:t-1})}
    \right)\right)}\right)^{-1} = \sigma(m_t),
\end{aligned}
\end{equation}
where $\sigma(x)= 1/(1+\exp(-x))$ and 
\begin{equation}
    m_t
    = \log\left(\frac{p(\vy_t \cond s_t=1, \vx_t, s_{1:t-1}, \data_{1:t-1})}{p(\vy_t \cond s_t=0, \vx_t, s_{1:t-1}, \data_{1:t-1})}\right)
    + \log\left(\frac{p(s_t=1)}{p(s_t=0)}\right),
\end{equation}
and similarly,
\begin{equation}
    p(s_t = 0 \cond s_{1:t-1}, \data_{1:t}) = 1 - \sigma(m_t).
\end{equation}
Finally, the transition between states is given by $p(s_{1:t} \cond s_{1:t-1}) = p(s_t)$.

% \begin{algorithm}[htb]
% \begin{algorithmic}
%     \REQUIRE $\omega_{t-1, i}$ for $i\in\{1, \ldots, 2^{t-1}\}$, $\vy_t \in \real^\dimobs$
%     \FOR{$\ell \in \{0, \ldots, 2^{t-1}\}$}
%     \STATE{
%         $m_t \gets $
%         $a_{t,\ell} \gets p(\vy_t \cond r_t =\ell, \vy_{1:t-1})$ // \text{following \eqref{eq:rl-predictive}}
%     }
%     \STATE $\pi_{t,\ell} \gets p(r_t = \ell \cond  r_{t-1}=\ell - 1)$ // \text{following \eqref{eq:prior-runlength}}
%     \ENDFOR
% \end{algorithmic}
% \caption{
%     \texttt{CPL} weight update for $t \geq 1$
% }
% \label{algo:cpl-step}
% \end{algorithm}

%% file: sections/appendix/algorithms.tex
\section{Algorithms}

\begin{algorithm}[H]
    \small
    \begin{algorithmic}[1]
        \REQUIRE $(\vmu_0, \vSigma_0)$ // default prior beliefs
        \REQUIRE $\data_{t} = (\vx_t, \vy_t)$  // current observation
        \REQUIRE $\{r_{t-1}^{(k)}\}_{k=1}^K \in \{0, \ldots, t-1\}^K$ // bank of runlengths at time $t-1$
        \REQUIRE $\{p(r_{t-1}^{(k)}, \data_{1:t})\}_{k=1}^K$ // joint from past hypotheses
        \REQUIRE $\left\{(\vmu_{t-1}^{(k)}, \vSigma_{t-1}^{(k)})\right\}_{k=1}^K$ // beliefs from past hypotheses
        \REQUIRE $\vx_{t+1}$ // next-step observation
        \REQUIRE $p(\vy \cond \vtheta, \vx) = {\cal N}(\vy \cond \vtheta^\intercal\vx, \vR_t)$ // Choice of \cModel
        \STATE // Evaluate hypotheses if there is no changepoint
        \FOR{$k=1,\ldots,K$}
            \STATE $r_t^{(k)} \gets r_{t-1}^{(k)} + 1$
            \STATE $p(\vy_t \cond r_t^{(k)}, \vx_t, \data_{1:t-1}) \gets {\cal N}(\vy_t  \cond \vx_t^\intercal\,\vmu_{t-1}^{(k)},\,\vx_t^\intercal\,\vSigma_{t-1}^{(k)}\,\vx_t + \vR_t)$ // posterior predictive for $k$-th hypothesis
            \STATE $p(r_t^{(k)},\,\data_{1:t}) \gets p(\vy_t \cond r_t^{(k)}, \vx_t, \data_{1:t-1})\,p(r_{t-1}^{(k)}, \data_{1:t-1})\,p(r_t^{(k)} \cond r_{t-1}^{(k)})$ // update joint density
            \STATE $(\bar{\vmu}_t^{(k)}, \bar{\vSigma}_t^{(k)}) \gets (\vmu_{t-1}^{(k)}, \vSigma_{t-1}^{(k)})$
           \STATE $\tau_t(\vtheta_t; r_t^{(k)}) \gets {\cal N}(\vtheta_t \cond \bar{\vmu}_t, \bar{\vSigma}_t)$ // choice of \cPrior
           \STATE $q_t(\vtheta_t;\, r_t^{(k)}) \propto \tau_t(\vtheta_t; r_t^{(k)})\,p(\vy_t \cond \vtheta^\intercal\vx_t, \vR_t) \propto {\cal N}(\vtheta_t \cond \vmu_t^{(k)}, \vSigma_t^{(k)})$ // following \eqref{eq:ekf-update-step}
        \ENDFOR
        \STATE // Evaluate hypothesis under a changepoint
        \STATE $r_t^{(k+1)} \gets 0$
        \STATE $p(\vy_t \cond r_t^{(k+1)}, \vx_t, \data_{1:t-1}) \gets {\cal N}(\vy_t  \cond \vx_t^\intercal\,\vmu_0,\,\vx_t^\intercal\,\vSigma_0\,\vx_t + \vR_t)$ // posterior predictive for $k$-th hypothesis
        \STATE $p(r_t^{(k+1)}, \data_{1:t}) \gets p(\vy_t \cond r_t^{(k+1)},\,\vx_t,\,\data_{1:t-1})\sum_{k=1}^K p(r_t^{(k)}, \data_{1:t})\,p(r_t^{(t+1)} \cond r_t^{(k)} - 1)$
        \STATE // Extend number of hypotheses to $K+1$ and keep top $K$ hypotheses
        \STATE $I_{1:k} = {\rm top.k}(\{p(r_t^{(1)},\,\data_{1:t}), \ldots, p(r_{t}^{(k+1)}, \data_{1:t})\},\,K)$
        \STATE $\{p(r_t^{(k)}, \data_{1:t})\}_{k=1}^K \gets {\rm slice.at}(\{p(r_t^{(k)},\,\data_{1:t})\}_{k=1}^{K+1},\,I_{1:K})$
        \STATE $\{(\vmu_t^{(k)}, \vSigma_t^{(k)})\}_{k=1}^K \gets {\rm slice.at}(\{(\vmu_t^{(k)},\,\vSigma_t^{(k)})\}_{k=1}^{K+1},\,I_{1:K})$
        \STATE // build weight and make prequential prediction
        \STATE $\nu_t(r_t^{(k)}) \gets \frac{p(r_t^{(k)}, \data_{1:t})}{\sum_{j=1}^K p(r_t^{(j)}, \data_{1:t})}$ for $k=1,\ldots, K$
        \STATE $\hat{\vy}_{t+1} \gets \vx_{t+1}^\intercal\,\left(\sum_{k=1}^K \nu_t(r_t^{(k)})\vmu_t^{(k)}\right)$ // prequential prediction under a linear-Gaussian model
       \RETURN $\{(\vmu_{t}^{(k)}, \vSigma_{t}^{(k)}, r_t^{(k)})\}_{k=1}^K$, $\hat{\vy}_{t+1}$
        \end{algorithmic}
    \caption{
        Implementation of \RLPR[K].
        We consider an update at time $t$ and one-step ahead forecasting at time $t+1$
        under a Gaussian linear model with known observation variance.
    }
    \label{algo:rl-pr-step}
\end{algorithm}
In Algorithm \ref{algo:rl-pr-step},
the function ${\rm top.k}(A,K)$ returns the indices of the top $K \geq 1$ elements of $A$ with highest value.
The function  ${\rm slice.at}(A, B)$ returns the elements in $A$ according to the list of indices $B$.
If $|A| \leq |B|$, we return all elements in $A$.

\begin{algorithm}[H]
    \small
    \begin{algorithmic}[1]
        \REQUIRE $\data_{t} = (\vx_t, \vy_t)$  // current observation
        \REQUIRE $\{r_{t-1}^{(k)}\}_{k=1}^K \in \{0, \ldots, t-1\}^K$ // bank of runlengths at time $t-1$
        \REQUIRE $\{p(r_{t-1}^{(k)}, \data_{1:t})\}_{k=1}^K$ // joint from past hypotheses
        \REQUIRE $\left\{(\vmu_{t-1}^{(k)}, \vSigma_{t-1}^{(k)})\right\}_{k=1}^K$ // beliefs from past hypotheses
        \REQUIRE $\vx_{t+1}$ // next-step observation
        \REQUIRE $p(\vy \cond \vtheta, \vx) = {\cal N}(\vy \cond \vtheta^\intercal\vx, \vR_t)$ // Choice of \cModel
        \STATE // Evaluate hypotheses if there is no changepoint
        \FOR{$k=1,\ldots,K$}
            \STATE $r_t^{(k)} \gets r_{t-1}^{(k)} + 1$
            \STATE $p(\vy_t \cond r_t^{(k)}, \vx_t, \data_{1:t-1}) \gets {\cal N}(\vy_t  \cond \vx_t^\intercal\,\vmu_{t-1}^{(k)},\,\vx_t^\intercal\,\vSigma_{t-1}^{(k)}\,\vx_t + \vR_t)$ // posterior predictive for $k$-th hypothesis
            \STATE $p(r_t^{(k)},\,\data_{1:t}) \gets p(\vy_t \cond r_t^{(k)}, \vx_t, \data_{1:t-1})\,p(r_{t-1}^{(k)}, \data_{1:t-1})\,p(r_t^{(k)} \cond r_{t-1}^{(k)})$ // update joint density
            \STATE $(\bar{\vmu}_t^{(k)}, \bar{\vSigma}_t^{(k)}) \gets (\vmu_{t-1}^{(k)}, \vSigma_{t-1}^{(k)})$
           \STATE $\tau_t(\vtheta_t; r_t^{(k)}) \gets {\cal N}(\vtheta_t \cond \bar{\vmu}_t, \bar{\vSigma}_t)$ // choice of \cPrior
           \STATE $q_t(\vtheta_t;\, r_t^{(k)}) \propto \tau_t(\vtheta_t; r_t^{(k)})\,p(\vy_t \cond \vtheta^\intercal\vx_t, \vR_t) \propto {\cal N}(\vtheta_t \cond \vmu_t^{(k)}, \vSigma_t^{(k)})$ // following \eqref{eq:ekf-update-step}
        \ENDFOR
        \STATE // Evaluate hypothesis under a changepoint
        \STATE $r_t^{(k+1)} \gets 0$
        \STATE $\vmu_0 \gets \mathbb{E}[\vtheta_t \cond r_t, \vy_{1:t-1}]$ // following \eqref{eq:rl-mmpr-first-moment}
        \STATE $\vSigma_0 \gets \mathbb{E}[\vtheta_t\,\vtheta_t^\intercal \cond r_t, \vy_{1:t-1}] - \left(\mathbb{E}[\vtheta_t \cond r_t, \vy_{1:t-1}]\right)\left(\mathbb{E}[\vtheta_t \cond r_t, \vy_{1:t-1}]\right)^\intercal$ // following \eqref{eq:rl-mmpr-first-moment} and \eqref{eq:rl-mmpr-second-moment}
        \STATE $p(\vy_t \cond r_t^{(k+1)}, \vx_t, \data_{1:t-1}) \gets {\cal N}(\vy_t  \cond \vx_t^\intercal\,\vmu_0,\,\vx_t^\intercal\,\vSigma_0\,\vx_t + \vR_t)$ // posterior predictive for $k$-th hypothesis
        \STATE $p(r_t^{(k+1)}, \data_{1:t}) \gets p(\vy_t \cond r_t^{(k+1)},\,\vx_t,\,\data_{1:t-1})\sum_{k=1}^K p(r_t^{(k)}, \data_{1:t})\,p(r_t^{(t+1)} \cond r_t^{(k)} - 1)$
        \STATE // Extend number of hypotheses to $K+1$ and keep top $K$ hypotheses
        \STATE $I_{1:k} = {\rm top.k}(\{p(r_t^{(1)},\,\data_{1:t}), \ldots, p(r_{t}^{(k+1)}, \data_{1:t})\},\,K)$
        \STATE $\{p(r_t^{(k)}, \data_{1:t})\}_{k=1}^K \gets {\rm slice.at}(\{p(r_t^{(k)},\,\data_{1:t})\}_{k=1}^{K+1},\,I_{1:K})$
        \STATE $\{(\vmu_t^{(k)}, \vSigma_t^{(k)})\}_{k=1}^K \gets {\rm slice.at}(\{(\vmu_t^{(k)},\,\vSigma_t^{(k)})\}_{k=1}^{K+1},\,I_{1:K})$
        \STATE // build weight and make prequential prediction
        \STATE $\nu_t(r_t^{(k)}) \gets \frac{p(r_t^{(k)}, \data_{1:t})}{\sum_{j=1}^K p(r_t^{(j)}, \data_{1:t})}$ for $k=1,\ldots, K$
        \STATE $\hat{\vy}_{t+1} \gets \vx_{t+1}^\intercal\,\left(\sum_{k=1}^K \nu_t(r_t^{(k)})\vmu_t^{(k)}\right)$ // prequential prediction under a linear-Gaussian model
       \RETURN $\{(\vmu_{t}^{(k)}, \vSigma_{t}^{(k)}, r_t^{(k)})\}_{k=1}^K$, $\hat{\vy}_{t+1}$
        \end{algorithmic}
    \caption{
        Implementation of \texttt{RL[K]-MMPR}.
        We consider an update at time $t$ and one-step ahead forecasting at time $t+1$
        under a Gaussian linear model with known observation variance.
    }
    \label{algo:rl-mmpr-step}
\end{algorithm}

\begin{algorithm}[H]
\small
\begin{algorithmic}[1]
    \REQUIRE $\data_{t} = (\vx_t, \vy_t)$  // current observation
    \REQUIRE $\vx_{t+1}$ // next-step observation
    \REQUIRE $\epsilon \in (0,1)$ // restart threshold
    \REQUIRE $r_{t-1} \in \{0, \ldots, t-1\}$ // runlength at time $t-1$
    \REQUIRE $(\vmu_0, \vSigma_0)$ // default prior beliefs
    \REQUIRE $(\vmu_{t-1}, \vSigma_{t-1})$ // beliefs from prior step
    \REQUIRE $p(\vy \cond \vtheta, \vx) = {\cal N}(\vy \cond \vtheta^\intercal\vx, \vR_t)$ // Choice of \cModel
    %$h(\vtheta;\, \vx_{t})$
   \STATE $(r_t^{(0)}, r_t^{(1)}) \gets (0, r_{t-1} + 1)$ // choice of \cAux
   \STATE $p(\vy_t \cond r_t^{(0)}, \vx_t, \data_{1:t-1}) \gets {\cal N}(\vy_t  \cond \vx_t^\intercal\,\vmu_0,\,\vx_t^\intercal\,\vSigma_0\,\vx_t + \vR_t)$ // posterior predictive at changepoint
   \STATE $p(\vy_t \cond r_t^{(1)}, \vx_t, \data_{1:t-1}) \gets {\cal N}(\vy_t  \cond \vx_t^\intercal\,\vmu_{t-1},\,\vx_t^\intercal\,\vSigma_{t-1}\,\vx_t + \vR_t)$ // posterior predictive if no changepoint
   \STATE $\nu_t(r^{(1)}) \gets \frac{p(\vy_t \cond r_t^{(1)}, \vx_t, \data_{1:t-1})(1 - \pi)}
   {p(\vy_t \cond r_t^{(1)}, \vx_t, \data_{1:t-1})\,(1 - \pi) + p(\vy_t \cond r_t^{(0)}, \vx_t, \data_{1:t-1})\,\pi}$ // probability of no-changepoint at timestep $t$
   \STATE 
   \IF{$\nu(r_t^{(1)}) > \epsilon$}
       \STATE $r_t \gets r_t^{(1)}$
       \STATE $\bar\vmu_t^{(r_t)} \gets \vmu_{t-1}^{(r_{t-1})}\,\nu(r_t^{(1)}) + \vmu_0 \, \left(1 - \nu(r_t^{(1)})\right)$
       \STATE $\bar\vSigma_t^{(r_t)} \gets \vSigma_{t-1}^{(r_{t-1})}\,\nu(r_t^{(1)})^2 + \vSigma_0 \, \left(1 - \nu(r_t^{(1)})^2\right)$
    \ELSIF{$\nu(r_t^{(1)}) \leq \epsilon$}
        \STATE $r_t \gets r_t^{(0)}$
        \STATE $\bar\vmu_t^{(r_t)} \gets \vmu_0$
        \STATE $\bar\vSigma_t^{(r_t)} \gets \vSigma_0$
   \ENDIF
   \STATE $\tau_t(\vtheta_t; r_t) \gets {\cal N}(\vtheta_t \cond \bar{\vmu}_t, \bar{\vSigma}_t)$ // choice of \cPrior
   \STATE $q_t(\vtheta_t;\, r_t) \propto {\cal N}(\vtheta_t \cond \bar{\vmu}_t, \bar{\vSigma}_t)\,p(\vy_t \cond \vtheta^\intercal\vx_t, \vR_t) \propto {\cal N}(\vtheta_t \cond \vmu_t, \vSigma_t)$ // choice of \cPosterior --- via \eqref{eq:ekf-update-step}
    \STATE $\hat{\vy}_{t+1} \gets \vx_{t+1}^\intercal\,\vmu_t$ // prequential prediction (given linear-Gaussian model)
   \RETURN $(\vmu_{t}, \vSigma_{t}, r_t)$, $\hat{\vy}_{t+1}$
\end{algorithmic}
\caption{
    Implementation of \RLSPR, with update at time $t$ and for one-step ahead forecasting at time $t+1$,
    under a Gaussian linear model with known observation variance.
}
\label{algo:rl-spr-step}
\end{algorithm}

%% file: sections/appendix/full-prediction-in-bayes.tex
\section{Bayesian inference when the dynamics are known}\label{app:full-bayes-prediction}

If the dynamics of model parameters $p(\vtheta_{t+1}, \cond \vtheta_{t}, \auxv_{t+1})$ and
the dynamics for auxiliary variable $p(\auxv_{t+1} \cond \auxv_t)$ are known,
then a full-Bayesian treatment of the prediction at time $t+1$ is given 
by \textit{predicting} the expected model parameters and then performing inference.

Specifically, given the prior belief state
$p(\auxv_t,\vtheta_t\cond\data_{1:t})$
and the new input $\vx_{t+1}$,
the  predictive distribution 
over the output $\vy_{t+1}$
is given by 
\begin{equation}
\begin{aligned}
    p(\vy_{t+1}\cond\vx_{t+1},\data_{1:t})
    &= \sum_{\auxv_{t+1}}
    \int_{\vtheta_{t+1}}
    p(\vy_{t+1}\cond\vx_{t+1},\vtheta_{t+1})
    p(\auxv_{t+1},\vtheta_{t+1}\cond\data_{1:t}) \,\d \vtheta_{t+1},
    \end{aligned}
\end{equation}
where the parameter predictive
distribution is
\begin{equation}
\begin{aligned}
    p(\auxv_{t+1},\vtheta_{t+1}\cond\data_{1:t})
     &= \sum_{\auxv_t} \int_{\vtheta_t}
        p(\auxv_{t+1},\vtheta_{t+1}\cond
        \auxv_t,\vtheta_t)
            p(\auxv_t,\vtheta_t\cond\data_{1:t})  \,\d \vtheta_{t}.
    \end{aligned}
\end{equation}
Here, the parameter dynamics factorises
as
\begin{equation}
\begin{aligned}
    p(\auxv_{t+1},\vtheta_{t+1}\cond
        \auxv_t,\vtheta_t)
         &= p(\auxv_{t+1}\cond\auxv_t)
         p(\vtheta_{t+1}\cond\vtheta_t,\auxv_{t+1}),
\end{aligned}
\end{equation}
and the prior belief state factorises as
\begin{equation}
\begin{aligned}
    p(\auxv_{t},\vtheta_{t}\cond\data_{1:t})
         &= p(\auxv_{t}\cond\data_{1:t})
         p(\vtheta_t\cond\auxv_t,\data_{1:t}).
\end{aligned}
\end{equation}

% Given the above expressions, 
% we see that the main modeling assumptions are:
% (M.1) the choice of likelihood
% $p(y_t\cond\vx_t,\vtheta_t)$,
% which is governed by
% $h(\vtheta_t,\vx_t)$;
% (M.2) the choice of auxiliary variable
% $\auxv_t$ and its dynamics
% $p(\auxv_t\cond\auxv_{t-1})$,
% as well as the choice of parameters
% $\vtheta_t$ and their dynamics
% $p(\vtheta_t\cond\vtheta_{t-1},\auxv_{t})$;
% and
% (M.3) the choice of representation for the belief state $p(\auxv_t\cond\data_{1:t})$
% and $p(\vtheta_t\cond\auxv_t,\data_{1:t})$.

Combining all the above equations, we can derive an expression for the posterior predicted mean
\begin{equation}
\begin{aligned}
    \hat{\vy}_{t+1} &=
    \mathbb{E}[\vy_{t+1}\cond\vx_{t+1},\data_{1:t}]
    \\
    &=
    \sum_{\auxv_{t+1}}
    \sum_{\auxv_t}
    p(\auxv_{t+1}\cond\auxv_t)
    p(\auxv_t\cond\data_{1:t}) \\
    &\qquad  \times
    \int_{\vtheta_{t+1}}
        \int_{\vtheta_{t}}
        h(\vtheta_{t+1},\vx_{t+1})
    p(\vtheta_{t+1}\cond\vtheta_t,\auxv_{t+1})
    p(\vtheta_t\cond\auxv_t,\data_{1:t}) \,\d \vtheta_{t}\, \,\d \vtheta_{t+1}.
\end{aligned}
\end{equation}

After making the prediction, we observe the outcome $\vy_{t+1}$, and then update the belief state 
for the parameter $\vtheta_{t+1}$
(for each value
of $\vpsi_{t+1}$) 
using 
\begin{equation}
\begin{aligned}
    p(\vtheta_{t+1}\cond\auxv_{t+1},\data_{1:t+1})
        &= \frac{1}{Z_{t+1}^{\psi_{t+1}}}
    p(\vy_{t+1}\cond\vx_{t+1},\vtheta_{t+1})
    p(\vtheta_{t+1}\cond\auxv_{t+1},\data_{1:t}), \\
    Z_{t+1}^{\auxv_{t+1}} &= 
    p(\vy_{t+1}\cond\vx_{t+1},\data_{1:t},\auxv_{t+1}) = \int_{\vtheta_{t+1}}
    p(\vy_{t+1}\cond\vx_{t+1},\vtheta_{t+1})
    p(\vtheta_{t+1}\cond\auxv_{t+1},\data_{1:t}) \,\d \vtheta_{t+1},
    \\
   p(\vtheta_{t+1}\cond\auxv_{t+1},\data_{1:t}) 
    &= \int_{\vtheta_t} p(\vtheta_{t+1}\cond\vtheta_t,\auxv_{t+1})
    p(\vtheta_t\cond\data_{1:t}) \,\d \vtheta_{t}.
\end{aligned}
\end{equation}

Similarly we update the belief state for the auxiliary variable using 
\begin{equation}
\begin{aligned}
    p(\auxv_{t+1}\cond\data_{1:t+1})
        &= \frac{1}{Z_{t+1}}
    p(\vy_{t+1}\cond\vx_{t+1},\auxv_{t+1})
    p(\auxv_{t+1}\cond\data_{1:t}), \\
    Z_{t+1} &= 
    p(\vy_{t+1}\cond\vx_{t+1},\data_{1:t})
    = \sum_{\auxv_{t+1}}
    Z_{t+1}^{\auxv_{t+1}}
    p(\auxv_{t+1}\cond\data_{1:t}) ,
    \\
       p(\auxv_{t+1}\cond\data_{1:t}) 
    &= \sum_{\auxv_t} p(\auxv_{t+1}\cond\auxv_{t})\, p(\auxv_t\cond\data_{1:t}).
\end{aligned}
\end{equation}

This full Bayesian treatment is different from BONE, because we do not assume well-specified
dynamics. Thus, BONE 
makes predictions at time $t+1$ with only the posterior at time $t$, i.e.,
it does not require a \textit{predict} step.

%% file: main.bbl
\begin{thebibliography}{88}
\providecommand{\natexlab}[1]{#1}
\providecommand{\url}[1]{\texttt{#1}}
\expandafter\ifx\csname urlstyle\endcsname\relax
  \providecommand{\doi}[1]{doi: #1}\else
  \providecommand{\doi}{doi: \begingroup \urlstyle{rm}\Url}\fi

\bibitem[Abélès et~al.(2024)Abélès, de~Vilmarest, and Wintemberger]{abeles2024adaptive}
Baptiste Abélès, Joseph de~Vilmarest, and Olivier Wintemberger.
\newblock Adaptive time series forecasting with markovian variance switching, 2024.

\bibitem[Adams \& MacKay(2007)Adams and MacKay]{adams2007bocd}
Ryan~Prescott Adams and David J.~C. MacKay.
\newblock Bayesian online changepoint detection, 2007.

\bibitem[Agudelo-Espa{\~n}a et~al.(2020)Agudelo-Espa{\~n}a, Gomez-Gonzalez, Bauer, Sch{\"o}lkopf, and Peters]{agudelo2020bocdprediction}
Diego Agudelo-Espa{\~n}a, Sebastian Gomez-Gonzalez, Stefan Bauer, Bernhard Sch{\"o}lkopf, and Jan Peters.
\newblock Bayesian online prediction of change points.
\newblock In \emph{Conference on Uncertainty in Artificial Intelligence}, pp.\  320--329. PMLR, 2020.

\bibitem[Alami(2023)]{alami2023banditnonstationary}
Reda Alami.
\newblock Bayesian change-point detection for bandit feedback in non-stationary environments.
\newblock In \emph{Asian Conference on Machine Learning}, pp.\  17--31. PMLR, 2023.

\bibitem[Alami et~al.(2020)Alami, Maillard, and F{\'e}raud]{alami2020restartedbocd}
R{\'e}da Alami, Odalric Maillard, and Raphael F{\'e}raud.
\newblock Restarted bayesian online change-point detector achieves optimal detection delay.
\newblock In \emph{International conference on machine learning}, pp.\  211--221. PMLR, 2020.

\bibitem[Altamirano et~al.(2023)Altamirano, Briol, and Knoblauch]{altamirano2023robust}
Matias Altamirano, François-Xavier Briol, and Jeremias Knoblauch.
\newblock Robust and scalable bayesian online changepoint detection, 2023.

\bibitem[Aminikhanghahi \& Cook(2017)Aminikhanghahi and Cook]{aminikhanghahi2017changepointsurvey}
Samaneh Aminikhanghahi and Diane~J Cook.
\newblock A survey of methods for time series change point detection.
\newblock \emph{Knowledge and information systems}, 51\penalty0 (2):\penalty0 339--367, 2017.

\bibitem[Arroyo et~al.(2025)Arroyo, Gravina, Gutteridge, Barbero, Gallicchio, Dong, Bronstein, and Vandergheynst]{arroyo2025vanishing}
{\'A}lvaro Arroyo, Alessio Gravina, Benjamin Gutteridge, Federico Barbero, Claudio Gallicchio, Xiaowen Dong, Michael Bronstein, and Pierre Vandergheynst.
\newblock On vanishing gradients, over-smoothing, and over-squashing in gnns: Bridging recurrent and graph learning.
\newblock \emph{arXiv preprint arXiv:2502.10818}, 2025.

\bibitem[Ash \& Adams(2020)Ash and Adams]{Ash2020}
Jordan~T Ash and Ryan~P Adams.
\newblock On warm-starting neural network training.
\newblock In \emph{NIPS}, 2020.
\newblock URL \url{http://arxiv.org/abs/1910.08475}.

\bibitem[Barry \& Hartigan(1992)Barry and Hartigan]{barry1992ppm}
Daniel Barry and John~A Hartigan.
\newblock Product partition models for change point problems.
\newblock \emph{The Annals of Statistics}, pp.\  260--279, 1992.

\bibitem[Basseville et~al.(1993)Basseville, Nikiforov, et~al.]{basseville1993onlinedetection}
Michele Basseville, Igor~V Nikiforov, et~al.
\newblock \emph{Detection of abrupt changes: theory and application}, volume 104.
\newblock Prentice hall Englewood Cliffs, 1993.

\bibitem[Battin(1982)]{battin1982apollo}
Richard~H Battin.
\newblock Space guidance evolution-a personal narrative.
\newblock \emph{Journal of Guidance, Control, and Dynamics}, 5\penalty0 (2):\penalty0 97--110, 1982.

\bibitem[Beal et~al.(2001)Beal, Ghahramani, and Rasmussen]{beal2001infinite}
Matthew Beal, Zoubin Ghahramani, and Carl Rasmussen.
\newblock The infinite hidden markov model.
\newblock \emph{Advances in neural information processing systems}, 14, 2001.

\bibitem[Bencomo et~al.(2023)Bencomo, Snell, and Griffiths]{bencomo2023implicit}
Gianluca~M. Bencomo, Jake~C. Snell, and Thomas~L. Griffiths.
\newblock Implicit maximum a posteriori filtering via adaptive optimization, 2023.

\bibitem[Bernardo \& Smith(1994)Bernardo and Smith]{Bernardo94}
J.~Bernardo and A.~Smith.
\newblock \emph{Bayesian Theory}.
\newblock John Wiley, 1994.

\bibitem[Bissiri et~al.(2016)Bissiri, Holmes, and Walker]{bissiri2016generalbayes}
Pier~Giovanni Bissiri, Chris~C Holmes, and Stephen~G Walker.
\newblock A general framework for updating belief distributions.
\newblock \emph{Journal of the Royal Statistical Society: Series B (Statistical Methodology)}, 78\penalty0 (5):\penalty0 1103--1130, 2016.

\bibitem[Blundell et~al.(2015)Blundell, Cornebise, Kavukcuoglu, and Wierstra]{blundell2015bbb}
Charles Blundell, Julien Cornebise, Koray Kavukcuoglu, and Daan Wierstra.
\newblock Weight uncertainty in neural network.
\newblock In Francis Bach and David Blei (eds.), \emph{Proceedings of the 32nd International Conference on Machine Learning}, volume~37 of \emph{Proceedings of Machine Learning Research}, pp.\  1613--1622, Lille, France, 07--09 Jul 2015. PMLR.
\newblock URL \url{https://proceedings.mlr.press/v37/blundell15.html}.

\bibitem[Bradbury et~al.(2018)Bradbury, Frostig, Hawkins, Johnson, Leary, Maclaurin, Necula, Paszke, Vander{P}las, Wanderman-{M}ilne, and Zhang]{jax2018github}
James Bradbury, Roy Frostig, Peter Hawkins, Matthew~James Johnson, Chris Leary, Dougal Maclaurin, George Necula, Adam Paszke, Jake Vander{P}las, Skye Wanderman-{M}ilne, and Qiao Zhang.
\newblock {JAX}: composable transformations of {P}ython+{N}um{P}y programs, 2018.
\newblock URL \url{http://github.com/google/jax}.

\bibitem[Cai et~al.(2021)Cai, Sener, and Koltun]{cai2021ocl}
Zhipeng Cai, Ozan Sener, and Vladlen Koltun.
\newblock Online continual learning with natural distribution shifts: An empirical study with visual data.
\newblock In \emph{Proceedings of the IEEE/CVF international conference on computer vision}, pp.\  8281--8290, 2021.

\bibitem[Cao et~al.(2024)Cao, Zhang, Sun, Liu, Yau, and Li]{Cao2024}
Wenhan Cao, Tianyi Zhang, Zeju Sun, Chang Liu, Stephen S-T Yau, and Shengbo~Eben Li.
\newblock Nonlinear bayesian filtering with natural gradient gaussian approximation.
\newblock \emph{arXiv [eess.SY]}, October 2024.
\newblock URL \url{http://arxiv.org/abs/2410.15832}.

\bibitem[Cartea et~al.(2023{\natexlab{a}})Cartea, Drissi, and Osselin]{cartea2023bandits}
{\'A}lvaro Cartea, Fay{\c{c}}al Drissi, and Pierre Osselin.
\newblock Bandits for algorithmic trading with signals.
\newblock \emph{Available at SSRN 4484004}, 2023{\natexlab{a}}.

\bibitem[Cartea et~al.(2023{\natexlab{b}})Cartea, Duran-Martin, and Sánchez-Betancourt]{cartea2023sharpbayes}
Álvaro Cartea, Gerardo Duran-Martin, and Leandro Sánchez-Betancourt.
\newblock Detecting toxic flow, 2023{\natexlab{b}}.

\bibitem[Chaer et~al.(1997)Chaer, Bishop, and Ghosh]{chaer1997mixturekf}
Wassim~S Chaer, Robert~H Bishop, and Joydeep Ghosh.
\newblock A mixture-of-experts framework for adaptive kalman filtering.
\newblock \emph{IEEE Transactions on Systems, Man, and Cybernetics, Part B (Cybernetics)}, 27\penalty0 (3):\penalty0 452--464, 1997.

\bibitem[Chang \& Athans(1978)Chang and Athans]{chang1978switchingkf}
Chaw-Bing Chang and Michael Athans.
\newblock State estimation for discrete systems with switching parameters.
\newblock \emph{IEEE Transactions on Aerospace and Electronic Systems}, AES-14\penalty0 (3):\penalty0 418--425, 1978.

\bibitem[Chang et~al.(2022)Chang, Murphy, and Jones]{chang2022diagonal}
Peter~G Chang, Kevin~Patrick Murphy, and Matt Jones.
\newblock On diagonal approximations to the extended kalman filter for online training of bayesian neural networks.
\newblock In \emph{Continual Lifelong Learning Workshop at ACML 2022}, 2022.

\bibitem[Chang et~al.(2023)Chang, Duran-Martin, Shestopaloff, Jones, and Murphy]{chang2023lofi}
Peter~G Chang, Gerardo Duran-Martin, Alex Shestopaloff, Matt Jones, and Kevin~Patrick Murphy.
\newblock Low-rank extended kalman filtering for online learning of neural networks from streaming data.
\newblock In \emph{Conference on Lifelong Learning Agents}, pp.\  1025--1071. PMLR, 2023.

\bibitem[Chen et~al.(2003)]{chen2003bayesianfiltersurvey}
Zhe Chen et~al.
\newblock Bayesian filtering: From kalman filters to particle filters, and beyond.
\newblock \emph{Statistics}, 182\penalty0 (1):\penalty0 1--69, 2003.

\bibitem[de~Freitas et~al.(2000)de~Freitas, Niranjan, Gee, and Doucet]{freitas2000smcneuralnets}
Joao~FG de~Freitas, Mahesan Niranjan, Andrew~H. Gee, and Arnaud Doucet.
\newblock Sequential monte carlo methods to train neural network models.
\newblock \emph{Neural computation}, 12\penalty0 (4):\penalty0 955--993, 2000.

\bibitem[de~Vilmarest \& Wintenberger(2021)de~Vilmarest and Wintenberger]{vilmarest2021viking}
Joseph de~Vilmarest and Olivier Wintenberger.
\newblock Viking: Variational bayesian variance tracking.
\newblock \emph{arXiv preprint arXiv:2104.10777}, 2021.

\bibitem[Dohare et~al.(2024)Dohare, Hernandez-Garcia, Lan, Rahman, Mahmood, and Sutton]{dohare2024continualbackprop}
Shibhansh Dohare, J~Fernando Hernandez-Garcia, Qingfeng Lan, Parash Rahman, A~Rupam Mahmood, and Richard~S Sutton.
\newblock Loss of plasticity in deep continual learning.
\newblock \emph{Nature}, 632\penalty0 (8026):\penalty0 768--774, 2024.

\bibitem[Doucet et~al.(2000)Doucet, de~Freitas, Murphy, and Russell]{Doucet2000}
Arnaud Doucet, Nando de~Freitas, Kevin Murphy, and Stuart Russell.
\newblock {Rao-Blackwellised} particle filtering for dynamic bayesian networks.
\newblock In \emph{{UAI}}, 2000.
\newblock URL \url{http://arxiv.org/abs/1301.3853}.

\bibitem[Duran-Martin et~al.(2022)Duran-Martin, Kara, and Murphy]{duran2022neuralbandits}
Gerardo Duran-Martin, Aleyna Kara, and Kevin Murphy.
\newblock Efficient online bayesian inference for neural bandits.
\newblock In \emph{International Conference on Artificial Intelligence and Statistics}, pp.\  6002--6021. PMLR, 2022.

\bibitem[Duran-Martin et~al.(2024)Duran-Martin, Altamirano, Shestpaloff, S{\'a}nchez-Betancourt, Knoblauch, Jones, Fran{\c c}ois-Xavier, and Murphy]{duranmartin2024-wlf}
Gerardo Duran-Martin, Matias Altamirano, Alexander~Y. Shestpaloff, Leandro S{\'a}nchez-Betancourt, Jeremias Knoblauch, Matt Jones, Briol Fran{\c c}ois-Xavier, and Kevin~P. Murphy.
\newblock Outlier-robust kalman filtering through generalised bayes.
\newblock In \emph{International Conference on Machine Learning}. PMLR, 2024.

\bibitem[Evensen(1994)]{evensen1994enkf}
Geir Evensen.
\newblock Sequential data assimilation with a nonlinear quasi-geostrophic model using monte carlo methods to forecast error statistics.
\newblock \emph{Journal of Geophysical Research: Oceans}, 99\penalty0 (C5):\penalty0 10143--10162, 1994.

\bibitem[Farrokhabadi et~al.(2022)Farrokhabadi, Browell, Wang, Makonin, Su, and Zareipour]{farrokhabadi2020electricitycovid}
Mostafa Farrokhabadi, Jethro Browell, Yi~Wang, Stephen Makonin, Wencong Su, and Hamidreza Zareipour.
\newblock Day-ahead electricity demand forecasting competition: Post-covid paradigm.
\newblock \emph{IEEE Open Access Journal of Power and Energy}, 9:\penalty0 185--191, 2022.
\newblock \doi{10.1109/OAJPE.2022.3161101}.

\bibitem[Fearnhead \& Liu(2007)Fearnhead and Liu]{fearnhead2007line}
Paul Fearnhead and Zhen Liu.
\newblock On-line inference for multiple changepoint problems.
\newblock \emph{Journal of the Royal Statistical Society Series B: Statistical Methodology}, 69\penalty0 (4):\penalty0 589--605, 2007.

\bibitem[Fearnhead \& Liu(2011)Fearnhead and Liu]{fearnhead2011adaptivecp}
Paul Fearnhead and Zhen Liu.
\newblock Efficient bayesian analysis of multiple changepoint models with dependence across segments.
\newblock \emph{Statistics and Computing}, 21:\penalty0 217--229, 2011.

\bibitem[Fearnhead \& Rigaill(2019)Fearnhead and Rigaill]{fearnhead2019robustchangepoint}
Paul Fearnhead and Guillem Rigaill.
\newblock Changepoint detection in the presence of outliers.
\newblock \emph{Journal of the American Statistical Association}, 114\penalty0 (525):\penalty0 169--183, 2019.

\bibitem[Fox et~al.(2007)Fox, Sudderth, Jordan, and Willsky]{fox2007sticky}
Emily~B Fox, Erik~B Sudderth, Michael~I Jordan, and Alan~S Willsky.
\newblock The sticky hdp-hmm: Bayesian nonparametric hidden markov models with persistent states.
\newblock \emph{Arxiv preprint}, 2, 2007.

\bibitem[Galashov et~al.(2024)Galashov, Titsias, György, Lyle, Pascanu, Whye, and Sahani]{Galashov2024}
Alexandre Galashov, Michalis~K Titsias, András György, Clare Lyle, Razvan Pascanu, Teh~Yee Whye, and Maneesh Sahani.
\newblock Non-stationary learning of neural networks with automatic soft parameter reset.
\newblock In \emph{NIPS}, November 2024.
\newblock URL \url{https://arxiv.org/abs/2411.04034}.

\bibitem[Gama et~al.(2008)Gama, Aguilar-Ruiz, and Klinkenberg]{gama2008streamlearn}
Joao Gama, Jesus Aguilar-Ruiz, and Ralf Klinkenberg.
\newblock Knowledge discovery from data streams.
\newblock \emph{Intelligent Data Analysis}, 12\penalty0 (3):\penalty0 251--252, 2008.

\bibitem[Ghahramani \& Hinton(2000)Ghahramani and Hinton]{ghahramani2000vsssm}
Zoubin Ghahramani and Geoffrey~E Hinton.
\newblock Variational learning for switching state-space models.
\newblock \emph{Neural computation}, 12\penalty0 (4):\penalty0 831--864, 2000.

\bibitem[Gunasekara et~al.(2023)Gunasekara, Pfahringer, Gomes, and Bifet]{gunasekara2023surveyocl}
Nuwan Gunasekara, Bernhard Pfahringer, Heitor~Murilo Gomes, and Albert Bifet.
\newblock Survey on online streaming continual learning.
\newblock In \emph{IJCAI}, pp.\  6628--6637, 2023.

\bibitem[Gupta et~al.(2024)Gupta, Wadhvani, and Rasool]{gupta2024changepointsurvey}
Muktesh Gupta, Rajesh Wadhvani, and Akhtar Rasool.
\newblock Comprehensive analysis of change-point dynamics detection in time series data: A review.
\newblock \emph{Expert Systems with Applications}, pp.\  123342, 2024.

\bibitem[Haykin(2004)]{haykin2004ekfnnet}
Simon Haykin.
\newblock \emph{Kalman filtering and neural networks}.
\newblock John Wiley \& Sons, 2004.

\bibitem[Hu{\v{s}}kov{\'a}(1999)]{huvskova1999gradualabruptchange}
M~Hu{\v{s}}kov{\'a}.
\newblock Gradual changes versus abrupt changes.
\newblock \emph{Journal of Statistical Planning and Inference}, 76\penalty0 (1-2):\penalty0 109--125, 1999.

\bibitem[Immer et~al.(2021)Immer, Korzepa, and Bauer]{immer2021improving}
Alexander Immer, Maciej Korzepa, and Matthias Bauer.
\newblock Improving predictions of bayesian neural nets via local linearization.
\newblock In \emph{International conference on artificial intelligence and statistics}, pp.\  703--711. PMLR, 2021.

\bibitem[Jones et~al.(2024)Jones, Chang, and Murphy]{jones2024bong}
Matt Jones, Peter Chang, and Kevin Murphy.
\newblock Bayesian online natural gradient ({BONG}).
\newblock In \emph{Advances in Neural Information Processing Systems}, May 2024.
\newblock URL \url{http://arxiv.org/abs/2405.19681}.

\bibitem[Kalman(1960)]{kalman1960filter}
Rudolph~Emil Kalman.
\newblock A new approach to linear filtering and prediction problems.
\newblock \emph{Transactions of the ASME--Journal of Basic Engineering}, 82\penalty0 (Series D):\penalty0 35--45, 1960.

\bibitem[Khan \& Rue(2023)Khan and Rue]{khan2023bayesian}
Mohammad~Emtiyaz Khan and H{\aa}vard Rue.
\newblock The bayesian learning rule.
\newblock \emph{Journal of Machine Learning Research}, 24\penalty0 (281):\penalty0 1--46, 2023.

\bibitem[Knoblauch \& Damoulas(2018)Knoblauch and Damoulas]{knoblauch2018varbocd}
Jeremias Knoblauch and Theodoros Damoulas.
\newblock Spatio-temporal bayesian on-line changepoint detection with model selection.
\newblock In \emph{International Conference on Machine Learning}, pp.\  2718--2727. PMLR, 2018.

\bibitem[Knoblauch et~al.(2018)Knoblauch, Jewson, and Damoulas]{knoblauch2018doublyrobust-bocd}
Jeremias Knoblauch, Jack~E Jewson, and Theodoros Damoulas.
\newblock Doubly robust bayesian inference for non-stationary streaming data with $\beta$-divergences.
\newblock In S.~Bengio, H.~Wallach, H.~Larochelle, K.~Grauman, N.~Cesa-Bianchi, and R.~Garnett (eds.), \emph{Advances in Neural Information Processing Systems}, volume~31. Curran Associates, Inc., 2018.
\newblock URL \url{https://proceedings.neurips.cc/paper_files/paper/2018/file/a3f390d88e4c41f2747bfa2f1b5f87db-Paper.pdf}.

\bibitem[Knoblauch et~al.(2022)Knoblauch, Jewson, and Damoulas]{knoblauch2022gvi}
Jeremias Knoblauch, Jack Jewson, and Theodoros Damoulas.
\newblock An optimization-centric view on bayes' rule: Reviewing and generalizing variational inference.
\newblock \emph{Journal of Machine Learning Research}, 23\penalty0 (132):\penalty0 1--109, 2022.
\newblock URL \url{http://jmlr.org/papers/v23/19-1047.html}.

\bibitem[Kuhl(1990)]{kuhl1990ridge}
Mark~R Kuhl.
\newblock Ridge regression signal processing.
\newblock \emph{NASA, Langley Research Center, Joint University Program for Air Transportation Research, 1989-1990}, 1990.

\bibitem[Kurle et~al.(2019)Kurle, Cseke, Klushyn, Van Der~Smagt, and G{\"u}nnemann]{kurle2019continual}
Richard Kurle, Botond Cseke, Alexej Klushyn, Patrick Van Der~Smagt, and Stephan G{\"u}nnemann.
\newblock Continual learning with bayesian neural networks for non-stationary data.
\newblock In \emph{International Conference on Learning Representations}, 2019.

\bibitem[Lambert et~al.(2022)Lambert, Bonnabel, and Bach]{lambert2022rvga}
Marc Lambert, Silv{\`e}re Bonnabel, and Francis Bach.
\newblock The recursive variational gaussian approximation (r-vga).
\newblock \emph{Statistics and Computing}, 32\penalty0 (1):\penalty0 10, 2022.

\bibitem[Lambert et~al.(2023)Lambert, Bonnabel, and Bach]{lambert2023lrvga}
Marc Lambert, Silv{\`e}re Bonnabel, and Francis Bach.
\newblock The limited-memory recursive variational gaussian approximation (l-rvga).
\newblock \emph{Statistics and Computing}, 33\penalty0 (3):\penalty0 70, 2023.

\bibitem[Li et~al.(2021)Li, Boyd, Smyth, and Mandt]{li2021onlinelearning}
Aodong Li, Alex Boyd, Padhraic Smyth, and Stephan Mandt.
\newblock Detecting and adapting to irregular distribution shifts in bayesian online learning, 2021.

\bibitem[Li et~al.(2010)Li, Chu, Langford, and Schapire]{li2010contextualbandits}
Lihong Li, Wei Chu, John Langford, and Robert~E Schapire.
\newblock A contextual-bandit approach to personalized news article recommendation.
\newblock In \emph{Proceedings of the 19th international conference on World wide web}, pp.\  661--670, 2010.

\bibitem[Linderman et~al.(2017)Linderman, Johnson, Miller, Adams, Blei, and Paninski]{linderman2017rslds}
Scott Linderman, Matthew Johnson, Andrew Miller, Ryan Adams, David Blei, and Liam Paninski.
\newblock Bayesian learning and inference in recurrent switching linear dynamical systems.
\newblock In \emph{Artificial intelligence and statistics}, pp.\  914--922. PMLR, 2017.

\bibitem[Liu(2023)]{liu2023bdemm}
Bin Liu.
\newblock Robust sequential online prediction with dynamic ensemble of multiple models: A review.
\newblock \emph{Neurocomputing}, pp.\  126553, 2023.

\bibitem[Liu et~al.(2023)Liu, Van~Roy, and Xu]{liu2023nonstationarybandits}
Yueyang Liu, Benjamin Van~Roy, and Kuang Xu.
\newblock Nonstationary bandit learning via predictive sampling.
\newblock In \emph{International Conference on Artificial Intelligence and Statistics}, pp.\  6215--6244. PMLR, 2023.

\bibitem[Lu et~al.(2018)Lu, Liu, Dong, Gu, Gama, and Zhang]{lu2018preqnonstationarysurvey}
Jie Lu, Anjin Liu, Fan Dong, Feng Gu, Joao Gama, and Guangquan Zhang.
\newblock Learning under concept drift: A review.
\newblock \emph{IEEE transactions on knowledge and data engineering}, 31\penalty0 (12):\penalty0 2346--2363, 2018.

\bibitem[Luo et~al.(2024)Luo, Cho, Demmel, Kozachenko, Li, and Liu]{luo2024-globalbayesopt}
Hengrui Luo, Younghyun Cho, James~W Demmel, Igor Kozachenko, Xiaoye~S Li, and Yang Liu.
\newblock Non-smooth bayesian optimization in tuning scientific applications.
\newblock \emph{The International Journal of High Performance Computing Applications}, 38\penalty0 (6):\penalty0 633--657, 2024.
\newblock \doi{10.1177/10943420241278981}.
\newblock URL \url{https://doi.org/10.1177/10943420241278981}.

\bibitem[Magill(1965)]{magill1965optimaladaptivefilter}
David Magill.
\newblock Optimal adaptive estimation of sampled stochastic processes.
\newblock \emph{IEEE Transactions on Automatic Control}, 10\penalty0 (4):\penalty0 434--439, 1965.

\bibitem[Mellor \& Shapiro(2013)Mellor and Shapiro]{mellor2013changepointthompsonsampling}
Joseph Mellor and Jonathan Shapiro.
\newblock Thompson sampling in switching environments with bayesian online change point detection.
\newblock \emph{arXiv preprint arXiv:1302.3721}, 2013.

\bibitem[Mishkin et~al.(2018)Mishkin, Kunstner, Nielsen, Schmidt, and Khan]{mishkin2018slang}
Aaron Mishkin, Frederik Kunstner, Didrik Nielsen, Mark Schmidt, and Mohammad~Emtiyaz Khan.
\newblock Slang: Fast structured covariance approximations for bayesian deep learning with natural gradient.
\newblock \emph{Advances in neural information processing systems}, 31, 2018.

\bibitem[Moreno-Pino et~al.(2024)Moreno-Pino, Arroyo, Waldon, Dong, and Cartea]{moreno-pino2024roughtransformer}
Fernando Moreno-Pino, \'{A}lvaro Arroyo, Harrison Waldon, Xiaowen Dong, and \'{A}lvaro Cartea.
\newblock Rough transformers: Lightweight and continuous time series modelling through signature patching.
\newblock In A.~Globerson, L.~Mackey, D.~Belgrave, A.~Fan, U.~Paquet, J.~Tomczak, and C.~Zhang (eds.), \emph{Advances in Neural Information Processing Systems}, volume~37, pp.\  106264--106294. Curran Associates, Inc., 2024.
\newblock URL \url{https://proceedings.neurips.cc/paper_files/paper/2024/file/bfe167fee3be2d862a56af82dee77720-Paper-Conference.pdf}.

\bibitem[Nassar et~al.(2022)Nassar, Brennan, Evans, and Lowrey]{nassar2022bam}
Josue Nassar, Jennifer Brennan, Ben Evans, and Kendall Lowrey.
\newblock Bam: Bayes with adaptive memory.
\newblock \emph{arXiv preprint arXiv:2202.02405}, 2022.

\bibitem[Nguyen et~al.(2017)Nguyen, Li, Bui, and Turner]{nguyen2017vcl}
Cuong~V Nguyen, Yingzhen Li, Thang~D Bui, and Richard~E Turner.
\newblock Variational continual learning.
\newblock \emph{arXiv preprint arXiv:1710.10628}, 2017.

\bibitem[Ollivier(2018)]{olllivier2018expfamekf}
Yann Ollivier.
\newblock {Online natural gradient as a Kalman filter}.
\newblock \emph{Electronic Journal of Statistics}, 12\penalty0 (2):\penalty0 2930 -- 2961, 2018.
\newblock \doi{10.1214/18-EJS1468}.
\newblock URL \url{https://doi.org/10.1214/18-EJS1468}.

\bibitem[Ostendorf et~al.(1996)Ostendorf, Digalakis, and Kimball]{ostendorf1996hmm}
Mari Ostendorf, Vassilios~V Digalakis, and Owen~A Kimball.
\newblock From hmm's to segment models: A unified view of stochastic modeling for speech recognition.
\newblock \emph{IEEE Transactions on speech and audio processing}, 4\penalty0 (5):\penalty0 360--378, 1996.

\bibitem[Reimann(2024)]{reimann2024changingfiltering}
Hans Reimann.
\newblock Towards robust inference for bayesian filtering of linear gaussian dynamical systems subject to additive change.
\newblock masterthesis, Universit{\"a}t Potsdam, 2024.

\bibitem[Riquelme et~al.(2018)Riquelme, Tucker, and Snoek]{riquelme2018banditshowdown}
Carlos Riquelme, George Tucker, and Jasper Snoek.
\newblock Deep bayesian bandits showdown: An empirical comparison of bayesian deep networks for thompson sampling, 2018.

\bibitem[Robert et~al.(2007)]{robert2007bayesianbook}
Christian~P Robert et~al.
\newblock \emph{The Bayesian choice: from decision-theoretic foundations to computational implementation}, volume~2.
\newblock Springer, 2007.

\bibitem[Roth et~al.(2017)Roth, Hendeby, Fritsche, and Gustafsson]{Roth2017enkf}
Michael Roth, Gustaf Hendeby, Carsten Fritsche, and Fredrik Gustafsson.
\newblock The ensemble kalman filter: a signal processing perspective.
\newblock \emph{{EURASIP J. Adv. Signal Processing}}, 2017\penalty0 (1):\penalty0 56, 2017.
\newblock URL \url{https://doi.org/10.1186/s13634-017-0492-x}.

\bibitem[Saat{\c{c}}i et~al.(2010)Saat{\c{c}}i, Turner, and Rasmussen]{saatcci2010GPBOCD}
Yunus Saat{\c{c}}i, Ryan~D Turner, and Carl~E Rasmussen.
\newblock Gaussian process change point models.
\newblock In \emph{Proceedings of the 27th International Conference on Machine Learning (ICML-10)}, pp.\  927--934, 2010.

\bibitem[S{\"a}rkk{\"a} \& Svensson(2023)S{\"a}rkk{\"a} and Svensson]{sarkka2023filtering}
Simo S{\"a}rkk{\"a} and Lennart Svensson.
\newblock \emph{Bayesian filtering and smoothing}, volume~17.
\newblock Cambridge university press, 2023.

\bibitem[Scalzo et~al.(2021)Scalzo, Arroyo, Stankovi{\'c}, and Mandic]{scalzo2021nonstationary}
Bruno Scalzo, Alvaro Arroyo, Ljubi{\v{s}}a Stankovi{\'c}, and Danilo~P Mandic.
\newblock Nonstationary portfolios: Diversification in the spectral domain.
\newblock In \emph{ICASSP 2021-2021 IEEE International Conference on Acoustics, Speech and Signal Processing (ICASSP)}, pp.\  5155--5159. IEEE, 2021.

\bibitem[Schirmer et~al.(2024)Schirmer, Zhang, and Nalisnick]{schirmer2024ssmtta}
Mona Schirmer, Dan Zhang, and Eric Nalisnick.
\newblock Test-time adaptation with state-space models.
\newblock \emph{arXiv preprint arXiv:2407.12492}, 2024.

\bibitem[Sellier \& Dellaportas(2023)Sellier and Dellaportas]{sellier2023robustbocdgp}
Jeremy Sellier and Petros Dellaportas.
\newblock Bayesian online change point detection with hilbert space approximate student-t process.
\newblock In \emph{International Conference on Machine Learning}, pp.\  30553--30569. PMLR, 2023.

\bibitem[Thompson(1933)]{thompson1933sampling}
William~R Thompson.
\newblock On the likelihood that one unknown probability exceeds another in view of the evidence of two samples.
\newblock \emph{Biometrika}, 25\penalty0 (3-4):\penalty0 285--294, 1933.

\bibitem[Titsias et~al.(2024)Titsias, Galashov, Rannen-Triki, Pascanu, Teh, and Bornschein]{titsias2023kalman}
Michalis~K Titsias, Alexandre Galashov, Amal Rannen-Triki, Razvan Pascanu, Yee~Whye Teh, and Jorg Bornschein.
\newblock Kalman filter for online classification of non-stationary data.
\newblock In \emph{ICLR}, 2024.

\bibitem[Van~den Burg \& Williams(2020)Van~den Burg and Williams]{van2020evaluation}
Gerrit~JJ Van~den Burg and Christopher~KI Williams.
\newblock An evaluation of change point detection algorithms.
\newblock \emph{arXiv preprint arXiv:2003.06222}, 2020.

\bibitem[Van~Gael et~al.(2008)Van~Gael, Saatci, Teh, and Ghahramani]{van2008beam}
Jurgen Van~Gael, Yunus Saatci, Yee~Whye Teh, and Zoubin Ghahramani.
\newblock Beam sampling for the infinite hidden markov model.
\newblock In \emph{Proceedings of the 25th international conference on Machine learning}, pp.\  1088--1095, 2008.

\bibitem[West \& Harrison(1997)West and Harrison]{West97}
Mike West and Jeff Harrison.
\newblock \emph{Bayesian forecasting and dynamic models}.
\newblock Springer, 1997.

\bibitem[Wilson et~al.(2010)Wilson, Nassar, and Gold]{wilson2010-bocd-hazard-rate}
Robert~C Wilson, Matthew~R Nassar, and Joshua~I Gold.
\newblock Bayesian online learning of the hazard rate in change-point problems.
\newblock \emph{Neural computation}, 22\penalty0 (9):\penalty0 2452--2476, 2010.

\bibitem[Zhang(2023)]{zhang2023ltbook}
Tong Zhang.
\newblock \emph{Mathematical Analysis of Machine Learning Algorithms}.
\newblock Cambridge University Press, 2023.
\newblock \doi{10.1017/9781009093057}.

\end{thebibliography}
